\pdfoutput=1
\documentclass[10pt,twocolumn,letterpaper]{article}
\usepackage[pagenumbers]{cvpr}
\usepackage{caption}
\usepackage{booktabs}
\usepackage{multirow}
\usepackage{makecell}
\usepackage{graphicx}
\usepackage{xcolor}
\usepackage{colortbl}
\usepackage{amsmath,amssymb}
\usepackage{pifont}
\usepackage{xspace}
\usepackage{tabularx}
\usepackage{array}
\usepackage{wrapfig}
\usepackage{subcaption}
\usepackage[most]{tcolorbox}
\usepackage{microtype}
\usepackage{enumitem}

\newcommand{\bench}{VibeVideoBench\xspace}

\newcommand{\cmark}{\ding{51}}
\newcommand{\xmark}{\ding{55}}
\definecolor{pendgray}{gray}{0.55}

\definecolor{finding}{RGB}{42,120,214}
\newtcolorbox{findingbox}{colback=finding!6,colframe=finding!70,boxrule=0.5pt,arc=1.5pt,left=3pt,right=3pt,top=2pt,bottom=2pt}

\definecolor{best}{RGB}{255,232,204}

\usepackage{tikz}
\usetikzlibrary{positioning,arrows.meta,calc,fit,backgrounds}
\definecolor{vblue}{HTML}{2A78D6}
\definecolor{vorange}{HTML}{EB6834}
\definecolor{vgreen}{HTML}{1BAF7A}
\definecolor{vpurple}{HTML}{6250D6}
\definecolor{vyellow}{HTML}{D99A00}
\usepackage{fontawesome5}

\usepackage{placeins}  %
\newcommand{\ico}[1]{\raisebox{-0.17em}{\includegraphics[height=1.05em]{fig/icons/fluent/#1.pdf}}}
\newcommand{\runin}[2]{\noindent\ico{#1}\hspace{0.3em}\textbf{#2}}
\newcommand{\logo}[1]{\raisebox{-0.12em}{\includegraphics[height=0.85em,width=0.85em,keepaspectratio]{fig/icons/#1.pdf}}}
\definecolor{lvorange}{HTML}{EB6834}

\definecolor{catC01}{HTML}{CB3E60}\definecolor{catC02}{HTML}{EFC34B}\definecolor{catC03}{HTML}{BDCA50}
\definecolor{catC04}{HTML}{3371D4}\definecolor{catC05}{HTML}{79B852}\definecolor{catC06}{HTML}{8B55D1}
\definecolor{catC07}{HTML}{5A66D1}\definecolor{catC08}{HTML}{DC6BAA}\definecolor{catC09}{HTML}{F3A247}
\definecolor{catC10}{HTML}{F17E69}\definecolor{catC11}{HTML}{13ACBF}\definecolor{catC12}{HTML}{12B897}
\makeatletter
\@namedef{caticon@C01}{megaphone}\@namedef{caticon@C02}{mobile-phone}\@namedef{caticon@C03}{party-popper}\@namedef{caticon@C04}{light-bulb}
\@namedef{caticon@C05}{bar-chart}\@namedef{caticon@C06}{artist-palette}\@namedef{caticon@C07}{video-game}\@namedef{caticon@C08}{clapper-board}
\@namedef{caticon@C09}{building-construction}\@namedef{caticon@C10}{musical-notes}\@namedef{caticon@C11}{teacher}\@namedef{caticon@C12}{page-facing-up}
\newcommand{\caticon}[1]{\ico{\@nameuse{caticon@#1}}}
\makeatother
\newcommand{\catsq}[1]{\tikz[baseline=-0.6ex]\fill[cat#1,rounded corners=0.5pt] (0,0) rectangle (0.62em,0.62em);}
\newcommand{\cattag}[1]{\catsq{#1}\hspace{0.3em}\caticon{#1}\hspace{0.3em}}
\newtcolorbox{specbox}[3]{enhanced,colback=cat#1!4,colframe=cat#1!70,boxrule=0.6pt,arc=2.5pt,
  left=5pt,right=5pt,top=3pt,bottom=3pt,colbacktitle=cat#1!13,coltitle=black,toptitle=2pt,bottomtitle=2pt,
  title={\caticon{#1}\hspace{0.35em}\textcolor{cat#1!75!black}{\bfseries #2}\hfill{\footnotesize\color{black!60}#3}}}
\newtcolorbox{l2box}[3]{enhanced,colback=cat#1!4,colframe=cat#1!70,boxrule=0.6pt,arc=2.5pt,
  left=5pt,right=5pt,top=3pt,bottom=3pt,colbacktitle=cat#1!13,coltitle=black,toptitle=2pt,bottomtitle=2pt,
  title={\ico{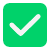}\hspace{0.35em}\textcolor{cat#1!75!black}{\bfseries #2}\hfill{\footnotesize\color{black!60}#3}}}
\newcommand{\circnum}[1]{\tikz[baseline=(c.base)]\node[draw,circle,line width=0.4pt,inner sep=0.5pt,minimum size=1.05em,font=\fontsize{6}{6}\selectfont](c){#1};}

\usepackage{placeins}
\usepackage{import}
\usepackage{float}
\definecolor{cvprblue}{rgb}{0.21,0.49,0.74}
\definecolor{titlelink}{rgb}{0.06,0.30,0.62}
\usepackage[pagebackref,breaklinks,colorlinks,allcolors=cvprblue]{hyperref}
\def\confName{CVPR}
\def\confYear{2027}

\newcommand\vvblogo[1]{\raisebox{-0.18\height}{\begin{tikzpicture}[x=#1/300,y=-#1/300]
  \fill[fill={rgb,255:red,48;green,44;blue,40}] (0,0) [rounded corners=#1*0.24] rectangle (300,300);
  \draw[draw={rgb,255:red,250;green,247;blue,242}, line width=#1*0.06, line cap=round, line join=round]
    (80.75,84) -- (130.25,216) -- (179.75,84) (214.25,216) -- (214.25,132);
  \fill[fill={rgb,255:red,200;green,92;blue,60}] (214.25,89.25) circle (14);
\end{tikzpicture}}}
\title{\vvblogo{1.15em}\ VibeVideoBench:\\ Evaluating Multimodal Agents on Code-Driven Video Generation}

\author{%
Muxi Diao$^{*}$\quad Wenbin Xu$^{*}$\quad Lele Yang$^{*}$\quad Xuyang Ma\quad Yuzhe Chen\quad Jiankang Chen$^{\dagger}$\\
Kaiyu Jiang\quad Changyi Liu\quad Haoyu Zhang\quad Hongyi Fu\quad Jianxiong Wang\quad Tianke Zhang\\
Haonan Fan\quad Yingxin Li\quad Bingyi Li\quad Xu Wang\quad Xuanyu Zheng$^{\dagger}$\quad Bin Wen$^{\dagger\ddagger}$\\
Tingting Gao\quad Wentao Zhang$^{\ddagger}$\quad Weiran Xu\quad Kongming Liang\quad Zhanyu Ma$^{\ddagger}$\\[0.7em]
{\normalsize Beijing University of Posts and Telecommunications\quad Zhongguancun Academy}\\
{\normalsize Peking University\quad Jilin University\quad Tianjin University}\\[0.7em]
{\small\raisebox{-0.15em}{\includegraphics[height=1.05em]{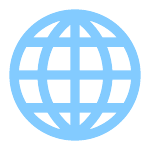}}\ \href{https://vibevideobench.github.io/}{\textcolor{titlelink}{vibevideobench.github.io}}\qquad
\raisebox{-0.15em}{\includegraphics[height=1.05em]{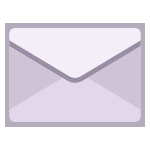}}\ \href{mailto:dmx@bupt.edu.cn}{\textcolor{titlelink}{dmx@bupt.edu.cn}}}
}

\makeatletter
\def\@maketitle{\null\kern 0.48in
  {\centering{\Large\bf \@title\par}\vspace*{20pt}
  {\large\lineskip .5em\begin{tabular}[t]{c}\@author\end{tabular}\par}\vspace*{2.9em}}
  {\centering
\vspace{-1.2em}
\definecolor{tcC01}{HTML}{CB3E60}\definecolor{tcC02}{HTML}{D9A62A}\definecolor{tcC04}{HTML}{3371D4}%
\definecolor{tcC06}{HTML}{8B55D1}\definecolor{tcC07}{HTML}{5A66D1}\definecolor{tcC09}{HTML}{F3A247}%
\definecolor{tcC10}{HTML}{F17E69}\definecolor{tcC12}{HTML}{12B897}%
\definecolor{tsInk}{HTML}{302C28}\definecolor{tsMuted}{HTML}{686158}%
\pgfmathsetlengthmacro{\tsGap}{5pt}%
\pgfmathsetlengthmacro{\tsCW}{(\textwidth-3*\tsGap)/4}%
\pgfmathsetlengthmacro{\tsFH}{\tsCW*9/16}%
\newcommand{\tscell}[5]{\makebox[\tsCW]{%
\begin{tikzpicture}[inner sep=0pt, outer sep=0pt, baseline={(0,-15pt)}]
  \useasboundingbox (0,-15pt) rectangle (\tsCW,\tsFH+1.6pt);
  \begin{scope}
    \clip[rounded corners=2.4pt] (0,0) rectangle (\tsCW,\tsFH);
    \node[anchor=south west] at (0,0) {\includegraphics[width=\tsCW,height=\tsFH]{fig/teaser/#1}};
  \end{scope}
  \draw[rounded corners=2.4pt, line width=0.35pt, black!14] (0,0) rectangle (\tsCW,\tsFH);
  \fill[#2] (1.2pt,\tsFH+0.6pt) rectangle (\tsCW-1.2pt,\tsFH+1.6pt);
  \node[anchor=north west, text=tsInk, text height=5.0pt, text depth=1.4pt, font=\sffamily\fontsize{6.4}{7}\selectfont\bfseries]
        at (0.5pt,-2.0pt) {#5};
  \node[anchor=north west, text=tsMuted, text height=4.4pt, text depth=1.2pt, font=\sffamily\fontsize{5.5}{6}\selectfont]
        at (0.5pt,-9.0pt) {\textcolor{#2}{\large\raisebox{-0.6pt}{$\bullet$}}\,#3\enspace\textcolor{black!30}{|}\enspace\faRobot\,#4};
\end{tikzpicture}}}%
\noindent\makebox[\textwidth]{\tscell{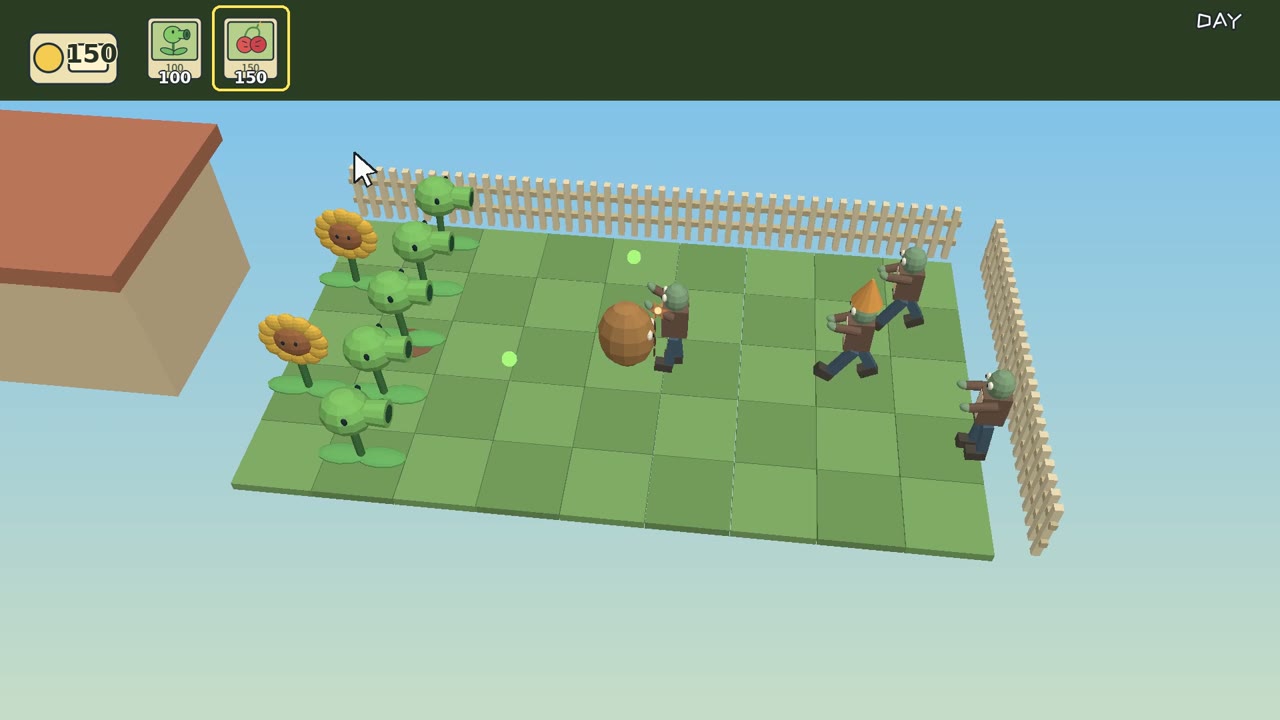}{tcC07}{Games \& sports}{Kimi K3}{Plants vs.\ Zombies, from 2D into 3D}\hspace{\tsGap}%
\tscell{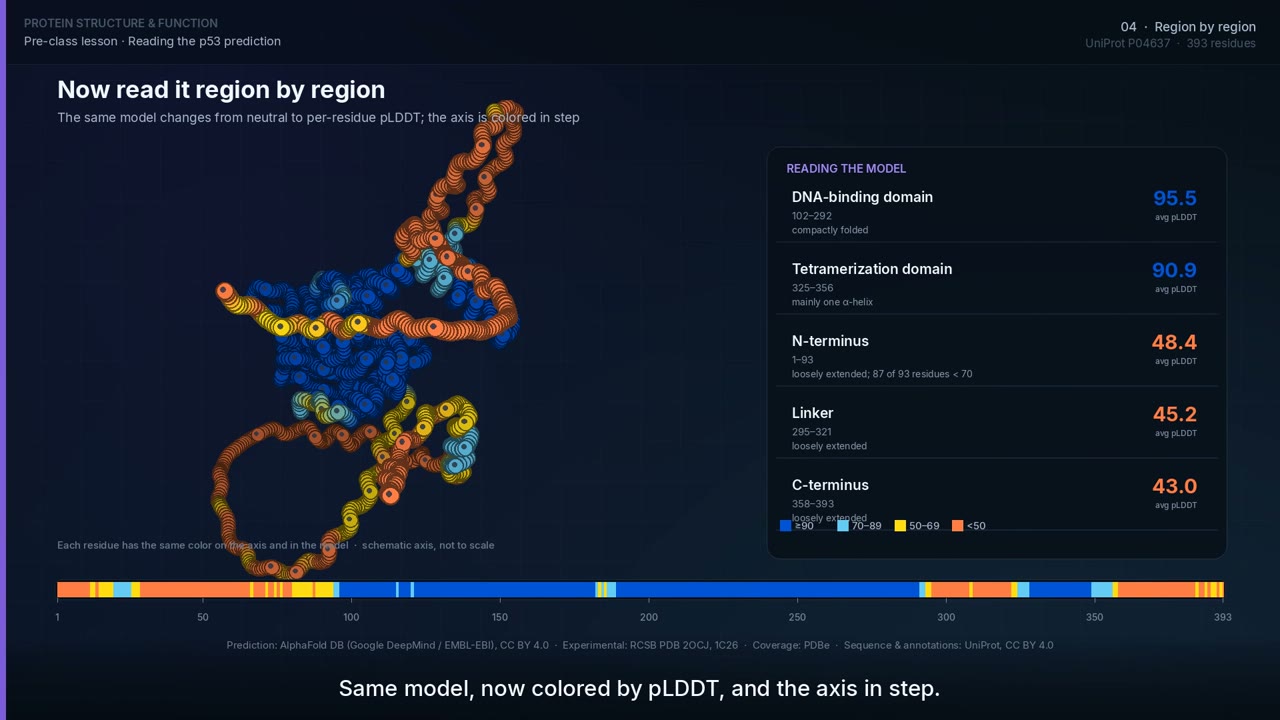}{tcC04}{Concept explainer}{DeepSeek-V4.1-Flash}{Reading an AlphaFold model of p53}\hspace{\tsGap}%
\tscell{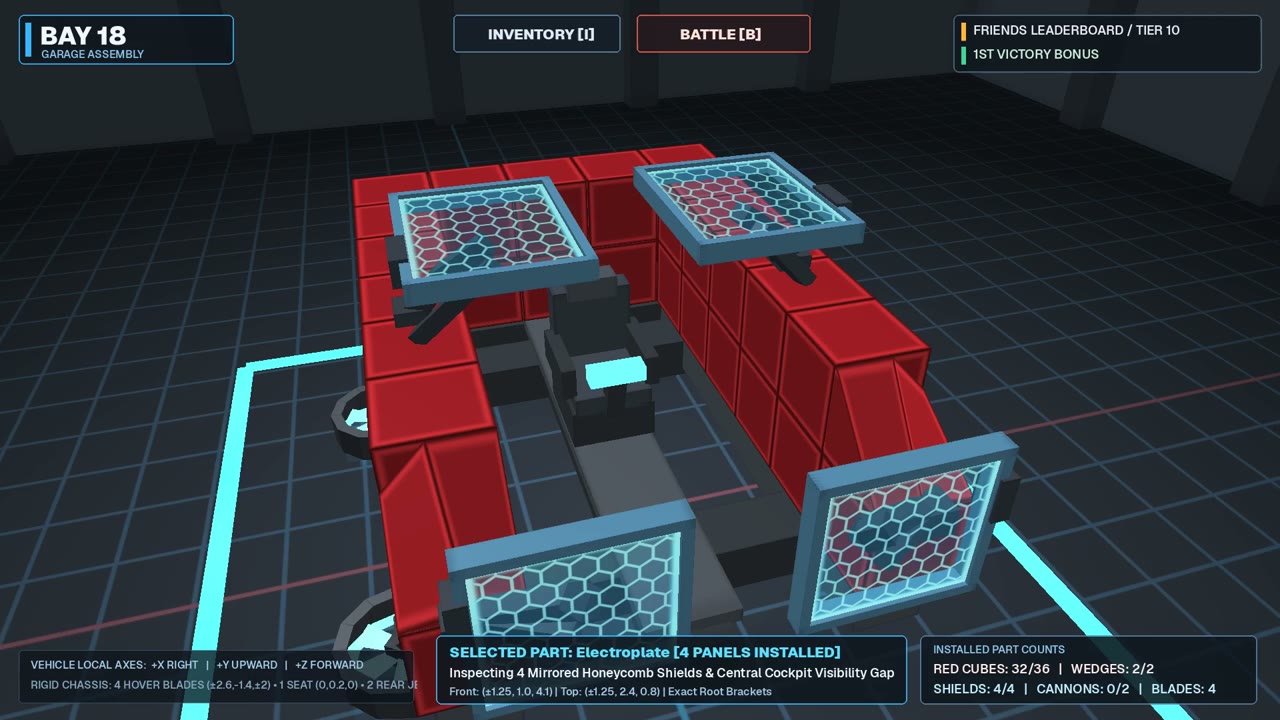}{tcC09}{Building \& process}{Gemini 4 Argon}{Hover-vehicle build and Mars test}\hspace{\tsGap}%
\tscell{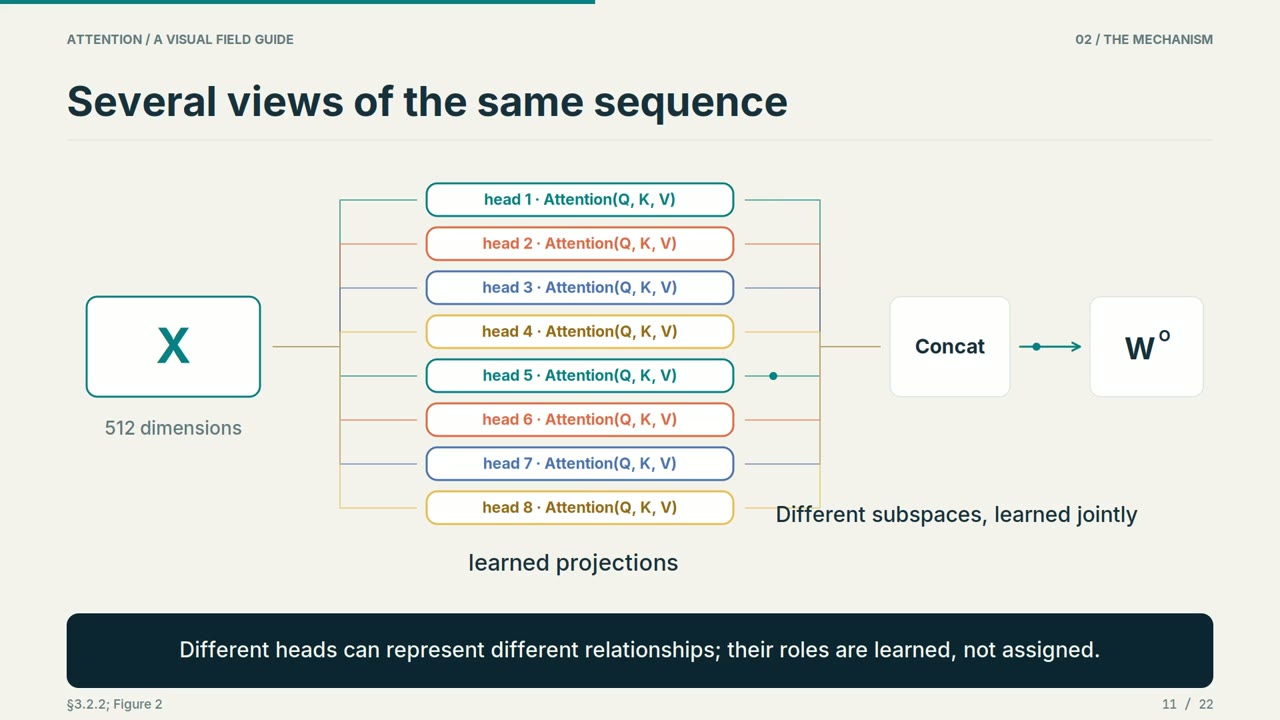}{tcC12}{Paper explainer}{GPT-6 Astra}{Attention Is All You Need, explained}}\\[4.5pt]
\makebox[\textwidth]{\tscell{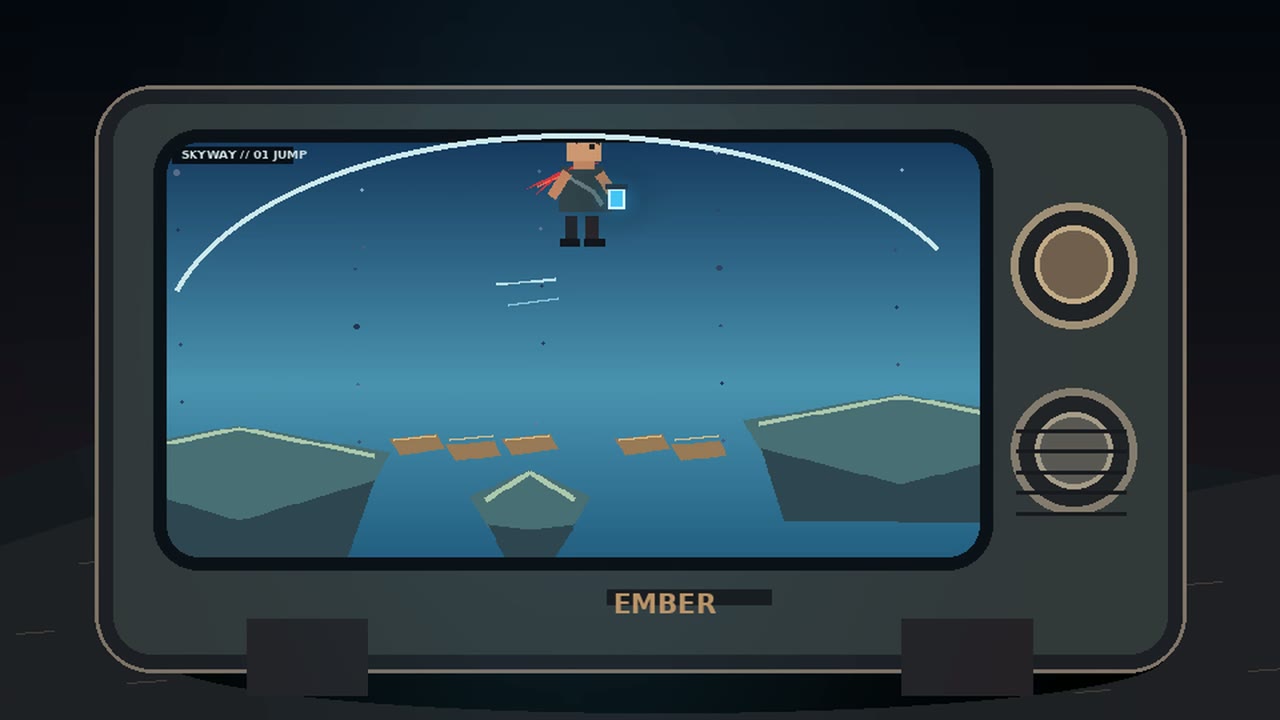}{tcC01}{Brand film}{GPT-5.6 Sol}{Ember Atlas: an old TV as a game portal}\hspace{\tsGap}%
\tscell{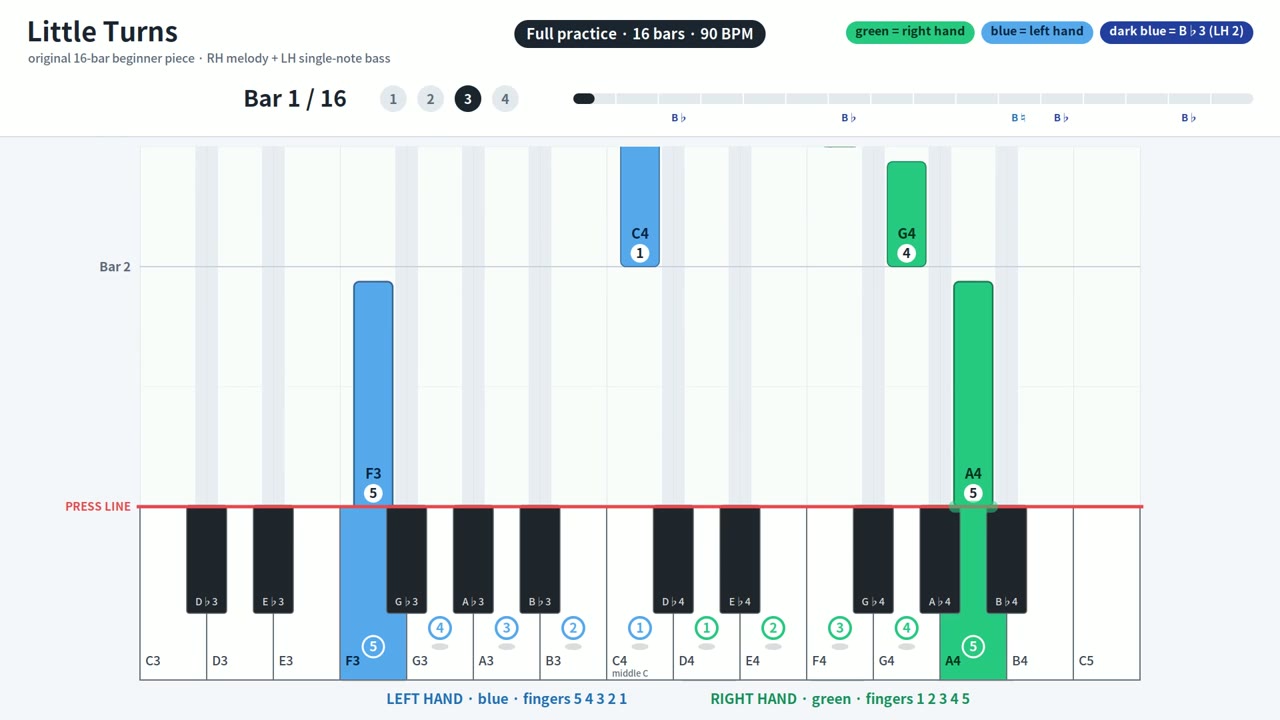}{tcC10}{Music \& sound}{Claude Opus 5.5}{Falling-note piano lesson}\hspace{\tsGap}%
\tscell{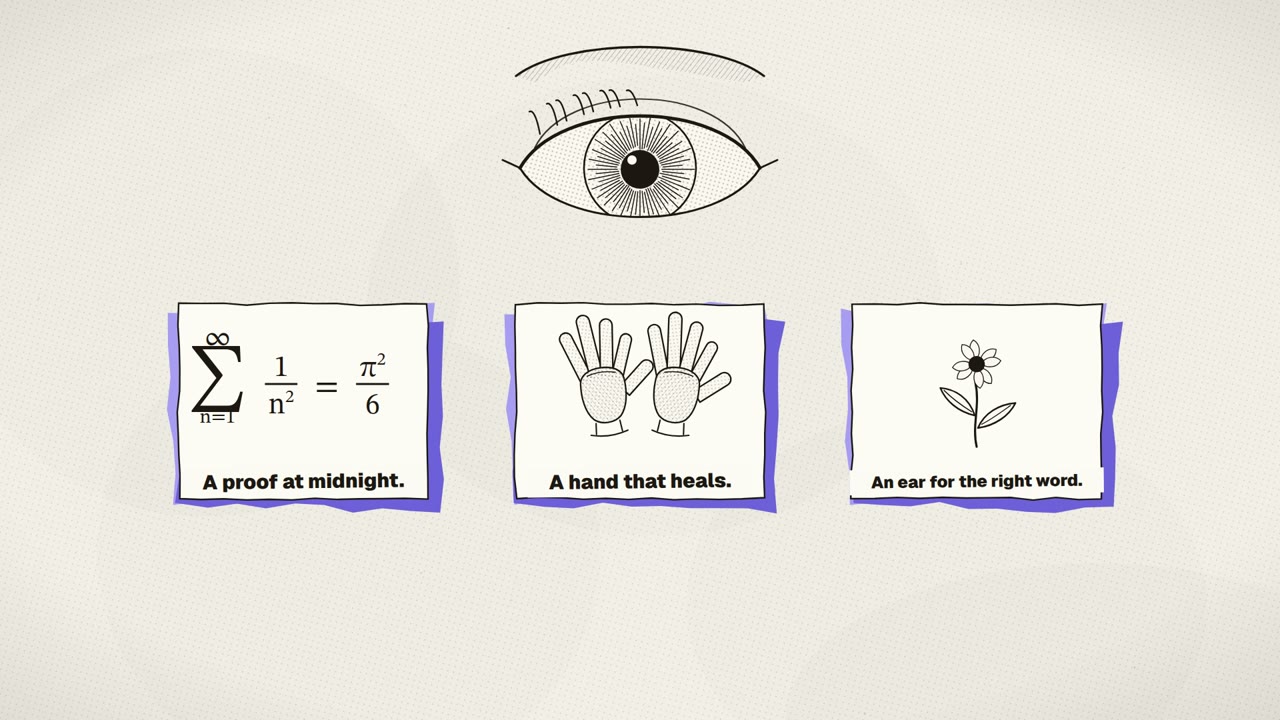}{tcC06}{Visual art \& form}{GLM-5.3-Flash}{From a question to a poster}\hspace{\tsGap}%
\tscell{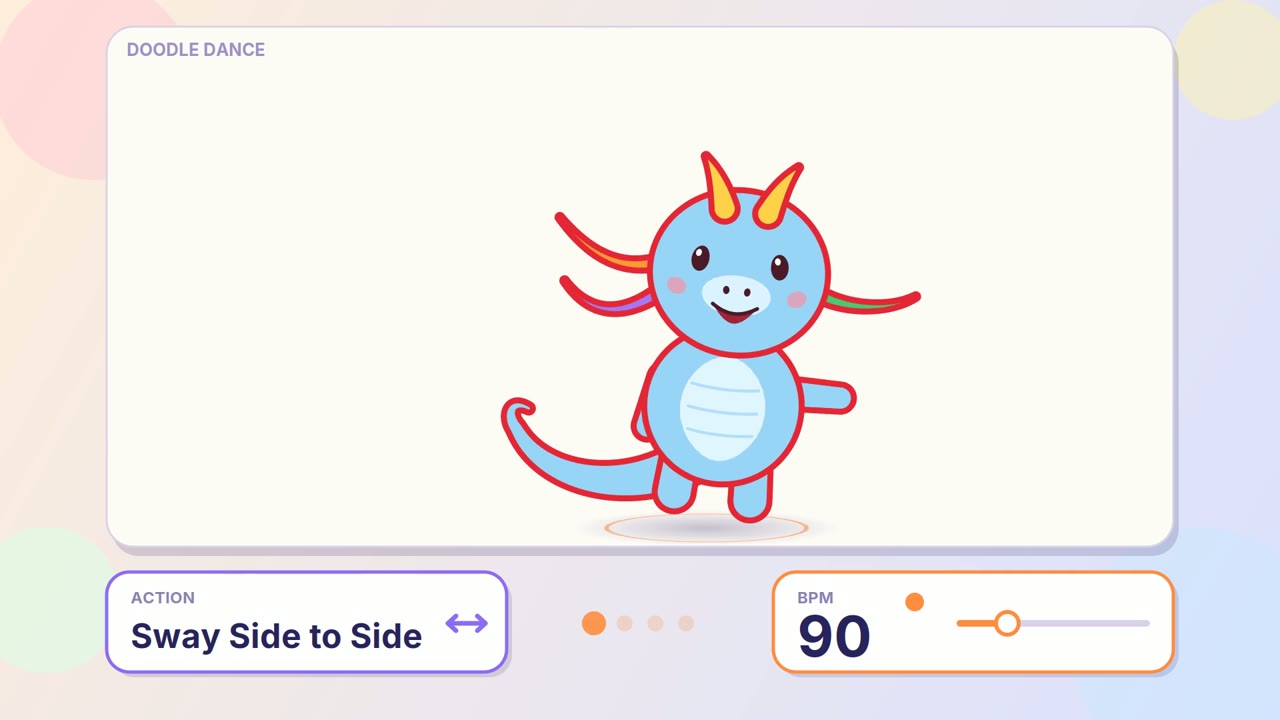}{tcC02}{Product demo}{Claude Sonnet 5.5}{A little dragon that dances to the beat}}\par
\vspace{0.4em}
\captionof{figure}{\textbf{Examples of agent-generated videos in \bench.}
Frames from unedited videos show 3D games, scientific explanations, product demos, and music-driven animations. Agents choose their tools and create each video autonomously through code within one hour, without image or video generation models. Labels identify the task, category, and agent.}
\label{fig:teaser}
\vspace{-0.6em}
}\vspace*{3.6em}}
\makeatother

\hypersetup{pdftitle={VibeVideoBench: Evaluating Multimodal Agents on Code-Driven Video Generation},pdfauthor={Muxi Diao and coauthors}}
\begin{document}
\addtocontents{toc}{\protect\setcounter{tocdepth}{-1}}
\maketitle
{\renewcommand\thefootnote{}\footnotetext{$^{*}$Equal contribution.\quad $^{\dagger}$Project lead.\quad $^{\ddagger}$Corresponding author.}}
\begin{abstract}
Multimodal agents can create videos by writing code and coordinating rendering and audio tools. Yet it remains unclear how reliably they can turn detailed specifications into complete videos. We introduce \bench, a benchmark of 239 self-contained tasks across 12 categories in English and Chinese. Agents operate in their native harnesses within a sandbox and must deliver a finished video within one hour, with audio as required. No image or video generation models are available. Evaluation combines programmatic delivery checks, approximately 13,000 task-specific rubric checks, and whole-video quality assessment. Across nine frontier agent systems and 2,151 rollouts, GPT-6 Astra and Claude Opus 5.5 lead both the judge-based quality ranking and the human preference ranking. However, the highest strict task-success rate is only 76.3\%. Agents typically build custom rendering pipelines, but their outputs still exhibit errors in required actions, state changes, and timing. Execution traces also reveal gaps between identifying a defect, revising the code, and delivering a corrected video. \bench provides a testbed for evaluating video production as an end-to-end agent task, encompassing requirement satisfaction, audiovisual quality, and delivery reliability.
\end{abstract}

\clearpage
\section{Introduction}
\label{sec:intro}

Video generation has advanced rapidly through diffusion transformers~\cite{brooks2024sora,veo3,kong2024hunyuanvideo,wan2025wan,yang2025cogvideox,polyak2024moviegen}. Yet many practical videos, from classroom lessons to product demonstrations, require precise text and numbers, consistent state changes, and sound aligned with on-screen events. Code offers explicit control over these elements and allows creators to revise the underlying content and timing. Tools such as Manim and Remotion already support this approach. Coding agents such as Claude Code and Codex~\cite{anthropic2025claudecode,openai2025codexcli} can now operate rendering and audio tools autonomously, allowing users to request complete videos from specifications and input assets. However, producing a finished video requires more than writing executable code. An agent must coordinate visuals and sound, manage rendering time, and inspect the output to ensure that it meets the user's requirements.

Existing benchmarks provide only a partial view of this capability. VBench~\cite{huang2024vbench,zheng2025vbench2} and EvalCrafter~\cite{liu2024evalcrafter} assess visual quality, temporal consistency, and prompt alignment in videos produced by generative models. Closer to our setting, Code2Video-Compass~\cite{chen2026code2videocompass} evaluates code-based video generation, but its evaluation protocol uses a single model call without execution feedback or repair. AgenticVBench~\cite{cao2026agenticvbench} evaluates autonomous agents on post-production tasks using existing footage. These settings do not directly assess whether a general-purpose agent can manage the full process of constructing a video with sound through code. Evaluating this capability requires measuring whether agents deliver a finished video, whether it meets the specification, and how well it communicates through visuals and sound.

We introduce \textbf{\bench} (Figs.~\ref{fig:teaser} and~\ref{fig:overview}), a benchmark for evaluating multimodal agents on code-driven video generation. Its 239 self-contained tasks span 12 categories in English and Chinese, from data stories and music-driven videos to classroom lessons and paper explainers. Given a specification and optional input assets, an agent must deliver a finished MP4 within one hour, with audio as specified by the task. Agents work autonomously in a sandbox and choose their own rendering tools; no image or video generation models are available. \mbox{Three design choices shape the benchmark.}

\textbf{Realistic yet checkable tasks.} We build tasks from real user requests, existing videos, asset collections, and educational or scientific materials. We turn these sources into self-contained specifications that define the required content, event sequence, timing, and audio. These specifications provide concrete criteria for evaluation while preserving creative freedom in presentation and implementation.

\textbf{Evaluation grounded in the video.} We assess each video at three levels. Program checks (L1) verify delivery requirements such as resolution, duration, and audible sound. Task-specific rubric checks (L2; about 13,000 in total) assess individual requirements, such as whether an event unfolds correctly, a state change is shown, or an audio cue occurs at the specified time. Quality scores (L3) evaluate the video as a whole across five dimensions: visual quality, aesthetics, content logic, timing, and audio.

\textbf{Evaluation of complete agent systems.} We evaluate each model together with its native harness, reflecting how users access these systems in practice. All systems receive the same tasks and run in identical containers with a common rendering toolchain and a one-hour budget. Each system runs autonomously at its highest supported reasoning effort, without further user input. Our evaluation covers nine frontier systems from six labs, yielding 2,151 scored rollouts.

\begingroup
\setlength{\parskip}{3pt}
\noindent\textbf{Findings.}
\begin{enumerate}[label=(\arabic*),leftmargin=1.8em,labelsep=0.4em,topsep=3pt,itemsep=4pt,parsep=0pt,partopsep=0pt]
\item {\raggedright\textbf{The benchmark is far from solved.}\par}
GPT-6 Astra~\cite{openai2026gpt6astra} and Claude Opus 5.5~\cite{anthropic2026opus55} lead both the judge-based quality ranking and the human preference ranking. However, the highest L2 strict success rate is only 76.3\%.

\item {\raggedright\textbf{Delivery reliability matters alongside video quality.}\par}
{\looseness=-1 Counting missing videos as zero changes system rankings. Of 224 runs without a playable video, 169 involved final renders that started too late, ran too slowly, or never began.\par}

\item {\raggedright\textbf{Agents fail more often on action and \mbox{timing requirements.}}\par}
In our sample, failure rates are roughly 10\% for action, timing, and continuity requirements, compared with 3.7\% for factual requirements.
\end{enumerate}
\endgroup

\noindent Our contributions are:
\begin{itemize}
\item \textbf{\bench.} We introduce a benchmark for general-purpose multimodal agents to produce videos through code. It provides 239 self-contained tasks across 12 categories in English and Chinese, covering diverse visual and audio requirements within a standardized execution environment (\S\ref{sec:bench}, Tab.~\ref{tab:compare}).
\item \textbf{A hierarchical evaluation protocol.} We combine programmatic delivery checks, approximately 13,000 task-specific rubric checks, and whole-video quality assessment. The protocol evaluates whether required content and events are realized, alongside the overall audiovisual quality. We assess judge reliability through blind human preference comparisons (\S\ref{sec:eval}).
\item \textbf{A large-scale empirical study.} We evaluate nine frontier agent systems across 2,151 rollouts. Analysis of generated videos and execution traces characterizes their rendering pipelines, inspection and revision behavior, delivery failures, and unmet task requirements (\S\ref{sec:exp}).
\end{itemize}

\section{Related Work}
\label{sec:related}

\noindent\textbf{Video generation: the pixel paradigm.}
Diffusion-transformer generators~\cite{brooks2024sora,veo3,kling,wan2025wan} synthesize pixels directly from text or images and are evaluated by dimension-wise suites~\cite{huang2024vbench,zheng2025vbench2,liu2024evalcrafter}, learned scorers~\cite{he2024videoscore}, physics probes~\cite{meng2024phygenbench,bansal2025videophy} and arenas~\cite{jiang2024genaiarena}; VideoArgus~\cite{zeng2026videoargus} builds output-blind, sample-specific rubrics for generation and editing. Story-video systems~\cite{lin2023videodirectorgpt,wu2025movieagent} use LLM planners to guide generative models. We study code-driven production, where an agent writes and executes programs that drive renderers, TTS and ffmpeg, enabling explicit control over content, timing and sound. These evaluations do not directly assess whether an agent can autonomously complete a production task and satisfy detailed visual and audio requirements.

\noindent\textbf{Code as a visual medium: the program paradigm.}
Programs are an editable, verifiable visual representation of charts, web pages, SVG and 3D scenes~\cite{shi2025chartmimic,si2025design2code,chen2025svgenius,gu2025blendergym}; ArtifactsBench~\cite{zhang2025artifactsbench} grades interactive artifacts from screenshots with checklist-guided MLLM judges. OpusAnimation~\cite{li2025opusanimation} introduces DCG-Bench for evaluating code-rendered animated charts from text or video inputs. For video, TheoremExplainAgent~\cite{ku2025theoremexplainagent}, Code2Video~\cite{chen2026code2video} and ManimAgent~\cite{jiang2026manimagent} generate educational Manim videos, and Paper2Video~\cite{zhu2025paper2video} turns research papers into narrated slide talks. These specialized systems organize planning, code generation, rendering and refinement through predefined workflows. VisualEDU~\cite{chen2025visualedu} benchmarks educational video generation, while an industry benchmark~\cite{heygen2026code2videobench} scores short silent clips written in one framework. Code2Video-Compass~\cite{chen2026code2videocompass} evaluates single-call code generation without execution feedback or repair, and does not evaluate audio. \bench instead gives a general-purpose agent a complete task and an execution environment: the agent chooses tools, writes and executes code, launches renders, inspects outputs, and decides how to revise and deliver the final video. No task-specific production workflow is prescribed.

\begin{table}[t]
\centering
\renewcommand{\cmark}{\textcolor{vgreen!90!black}{\ding{51}}}\renewcommand{\xmark}{\textcolor{black!28}{\ding{55}}}%
\newcommand{\pmark}{\textcolor{vyellow}{$\circ$}}%
\caption{\textbf{Positioning.} Among the benchmarks we know of, only \bench has general-purpose agents build complete audio-visual videos from scratch with code across open domains, graded with task-specific rubrics. \cmark: yes; \xmark: no; \pmark: partially.}
\label{tab:compare}
\setlength{\tabcolsep}{1.3pt}
\scriptsize
\resizebox{\columnwidth}{!}{%
\begin{tabular}{@{}llcccccc@{}}
\toprule
Benchmark & Output & \makecell{Pixels\\from} & \makecell{General\\agents} & \makecell{Open\\domain} & \makecell{Audio\\judged} & \makecell{Input\\assets} & \makecell{Task-spec.\\rubric} \\
\midrule
VBench~\cite{huang2024vbench} & T2V clip & diffusion & \xmark & \cmark & \xmark & \xmark & \xmark \\
\rowcolor{black!3.5}EvalCrafter~\cite{liu2024evalcrafter} & T2V clip & diffusion & \xmark & \cmark & \xmark & \xmark & \xmark \\
VideoArgus~\cite{zeng2026videoargus} & gen./edit. video & diffusion & \xmark & \cmark & \xmark & \xmark & \cmark \\
\rowcolor{black!3.5}ArtifactsBench~\cite{zhang2025artifactsbench} & web page & LLM code & \xmark & \cmark & \xmark & \xmark & \cmark \\
TheoremExplain~\cite{ku2025theoremexplainagent} & Manim video & pipeline & \xmark & \xmark & \pmark & \xmark & \xmark \\
\rowcolor{black!3.5}Paper2Video~\cite{zhu2025paper2video} & talk video & pipeline & \xmark & \xmark & \pmark & \cmark & \pmark \\
C2V-Compass~\cite{chen2026code2videocompass} & code video & LLM code & \xmark & \cmark & \xmark & \pmark & \cmark \\
\rowcolor{black!3.5}AgenticVBench~\cite{cao2026agenticvbench} & edited video & footage & \cmark & \cmark & \cmark & \cmark & \cmark \\
\midrule
\rowcolor{best!55}\textcolor{vorange}{\faStar}\,\textbf{\bench} & \textbf{code video} & \textbf{agent code} & \cmark & \cmark & \cmark & \cmark & \cmark \\
\bottomrule
\end{tabular}}
\vspace{-1.0em}
\end{table}

\noindent\textbf{Agent benchmarks.}
Software and computer-use agents are measured by SWE-bench~\cite{jimenez2024swebench}, Terminal-Bench~\cite{merrill2026terminalbench}, OSWorld~\cite{xie2024osworld} and PaperBench~\cite{starace2025paperbench}, and GDPval~\cite{patwardhan2026gdpval} grades occupational deliverables by expert comparison. Programmatic checks alone cannot establish whether a video communicates the required content clearly and coherently through visuals and sound. AgenticVBench~\cite{cao2026agenticvbench} also runs general agents in vendor harnesses, but on post-production of \emph{given} footage; in our setting, the agent must implement and execute the complete video-production process through code. Tab.~\ref{tab:compare} compares \bench\ with prior work on production, task requirements, deliverables and evaluation.

\noindent\textbf{Judges, rubrics and human preference.}
LLM and multimodal judges~\cite{zheng2023judging,kim2024prometheus,chen2024mllmjudge} exhibit position and self-preference biases~\cite{wang2024notfair} (we test self-preference in App.~\ref{app:x10}); rubric decomposition improves reliability in non-verifiable domains~\cite{arora2025healthbench,gunjal2025rubrics}, and pairwise human preference with Bradley--Terry aggregation~\cite{bradley1952rank,chiang2024chatbotarena} is a common validation approach. Omni models can see and hear~\cite{xu2025qwen3omni}, but audio-visual benchmarks show they remain weak at fine temporal alignment~\cite{gong2024avodyssey,hong2025worldsense}. We therefore route checks by modality, require timestamped evidence, use three independent panels, and measure where they fail.

\section{\bench}
\label{sec:bench}
\begin{figure*}[t]
\centering
\includegraphics[width=0.985\textwidth]{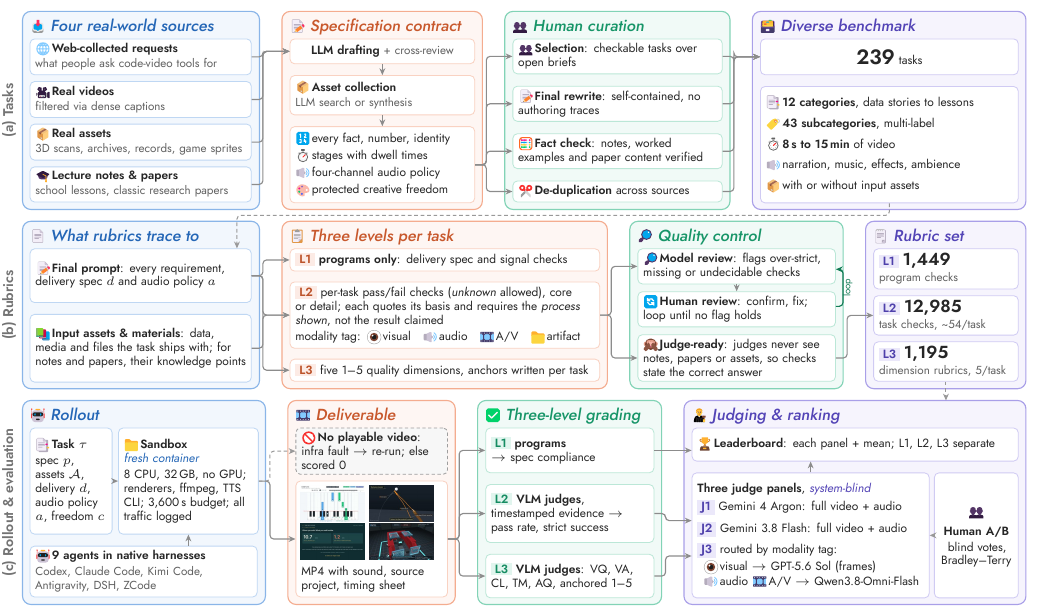}
\vspace{-8.68bp}%
\vspace{-0.4em}
\caption{\textbf{Overview of \bench.} \textbf{(a) Task construction.} LLM drafting, asset collection, and human curation turn four types of real-world sources into 239 self-contained tasks. \textbf{(b) Rubric construction.} Three-level rubrics are grounded in task specifications and input assets and refined through model and human review. \textbf{(c) Rollout and evaluation.} Agents operate in native harnesses within standardized sandboxes under a one-hour budget. Programmatic checks verify delivery requirements (L1), while three judge panels assess task-specific requirements (L2) and overall audiovisual quality (L3), with one panel routing evaluation by modality.}
\label{fig:overview}
\vspace{-0.6em}
\end{figure*}

\subsection{Task formulation}
We represent each task as $\tau=(p,\mathcal{A},d,a,c)$. The natural-language specification $p$ describes the required content, and the optional read-only asset package $\mathcal{A}$ supplies data, fonts, sprites, 3D scans or audio. The machine-readable delivery specification $d$ defines the resolution, frame rate, duration range $[t_{\min},t_{\max}]$ and, where required, exact frame count. The audio policy $a$ marks narration, music, sound effects and ambience as \emph{required}, \emph{optional} or \emph{not used}, while $c$ states the creative choices left to the agent.
An agent receives the task in a fresh container and must deliver an MP4 video, an editable source project and a timing sheet within one hour. Outputs are judged against the frozen task specification; no reference video is prescribed. Agents may choose any rendering stack. Image and video generation models are unavailable, while a TTS tool provides speech synthesis.

\subsection{Task construction}
\label{sec:construction}
Fig.~\ref{fig:overview}a summarizes task construction. \emph{Sources.} Tasks draw on four kinds of real-world sources: requests for code-based video tools, dense captions of real videos, asset packs such as 3D scans and game sprites, and lecture notes and research papers.
\emph{Self-contained specifications.} LLMs draft tasks from these sources and cross-review them under shared authoring guidelines. Each specification supplies the facts, numbers and object identities needed to complete the task, and makes the required event sequence, timing and audio explicit. It also states which creative choices remain open. Missing assets are retrieved or synthesized during task construction.
\emph{Human curation.} Human reviewers select tasks with checkable requirements, verify their facts and remove duplicates across sources. They ensure that prompts are self-contained and remove authoring notes and scoring hints. This process yields \textbf{239 tasks} (App.~\ref{app:data}).

\begin{figure*}[t]
\centering
\includegraphics[width=\linewidth]{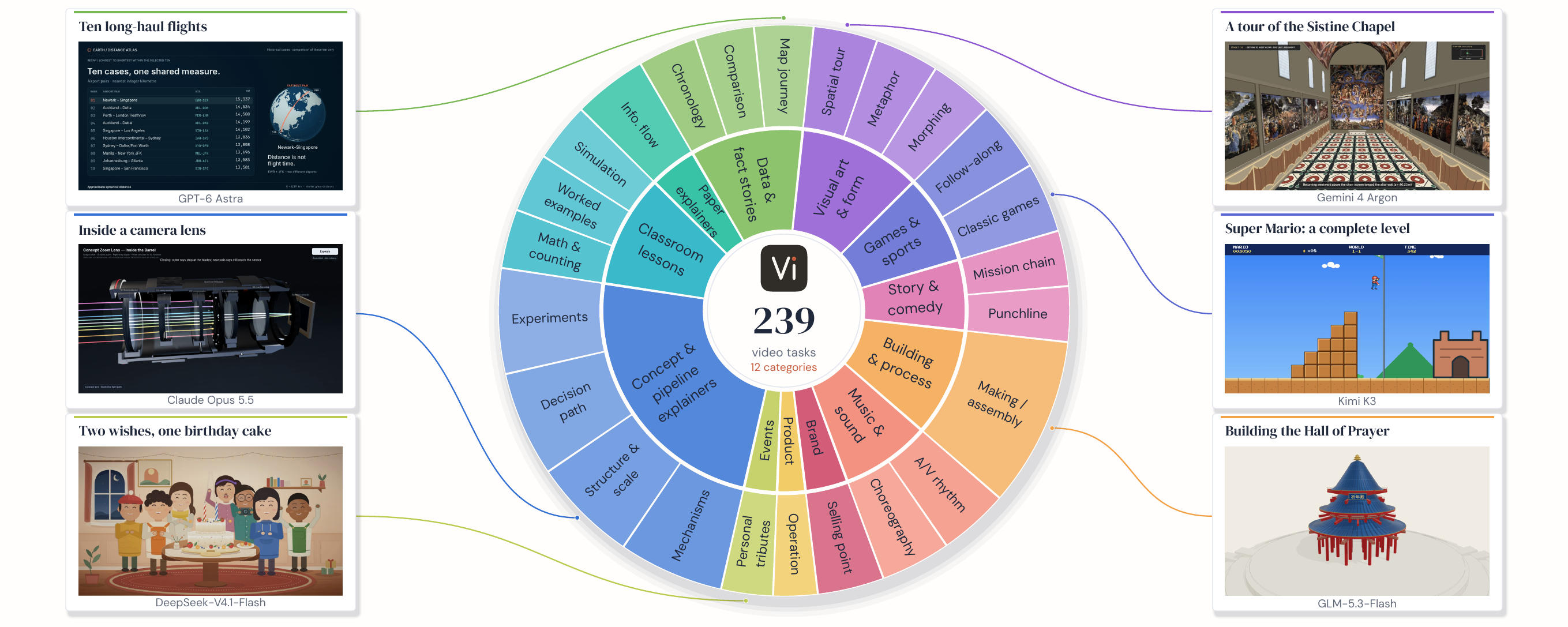}\\[0.2em]
\includegraphics[width=\linewidth]{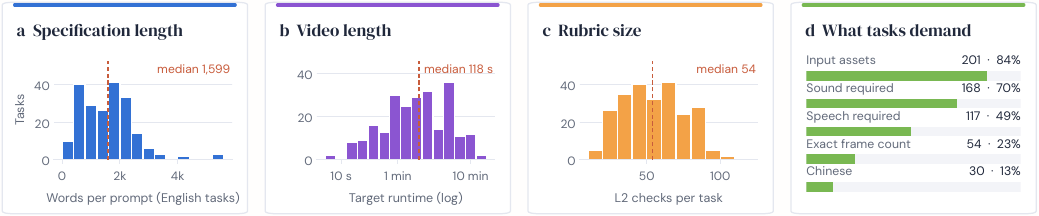}
\vspace{-1.5em}
\caption{\textbf{Benchmark composition (239 tasks).} \emph{Top:} the inner ring shows 12 categories, with spans proportional to task counts. The outer ring shows 24 presentation-form tags, each placed in a characteristic category; equal angles illustrate forms rather than their frequencies. Tasks can carry multiple tags (full taxonomy: App.~\ref{app:data}). The six cards are real deliveries, one per model family. \emph{Bottom:} (a) prompt length of the 209 English tasks (lessons and papers supply their content in notes or PDFs); (b) target runtime, 8\,s--15\,min; (c) L2 checks per task, 13--110; (d) share of tasks with each requirement.}
\label{fig:stats}
\vspace{0.15em}
\end{figure*}

\subsection{Statistics}
Fig.~\ref{fig:stats} summarizes the benchmark. Each task belongs to one of 12 categories based on content and purpose. Tasks also carry one or more of 43 cross-category tags describing presentation forms and knowledge domains. There are 209 English and 30 Chinese tasks, with requested video lengths ranging from 8\,s to 15\,min. Input assets accompany 201 tasks; 70\% require at least one audio channel and 49\% require speech. Most tasks specify $1920{\times}1080$ video at 30\,fps. Detailed statistics and the full taxonomy are in App.~\ref{app:data}.

\subsection{Rubrics}
\label{sec:rubric}
Rubrics are derived from the final task specification and input materials (Fig.~\ref{fig:overview}b) and remain hidden from agents during generation. \emph{L1 program checks} assess the delivery requirements in $d$ and $a$ using code.

\emph{L2 task checks} assess whether individual task requirements are satisfied. Each check quotes its supporting requirement, carries a modality tag and is labeled \emph{core} or \emph{detail} before any output is seen. Checks for required actions or processes demand observable evidence of their execution in the video; a caption claiming completion is insufficient.

\emph{L3 quality} evaluates the whole video along five shared dimensions: visual quality (VQ), visual aesthetics (VA), content logic (CL), timing and motion (TM), and audio quality and A/V sync (AQ). Each dimension uses a 1--5 scale with task-specific anchors at 1, 3 and 5.

Model review identifies gaps in requirement coverage, overly strict checks and criteria that cannot be judged reliably. Human reviewers verify these findings, and confirmed issues are corrected through iterative revision. Since judges do not receive the input assets, checks state any required facts from those assets explicitly. The final set contains 1,449 L1 checks, 12,985 L2 checks (about 54 per task) and 1,195 L3 dimension rubrics. See App.~\ref{app:rubric} for details and \S\ref{sec:eval} for scoring.

\section{Evaluation Protocol}
\label{sec:eval}

{\looseness=-1\noindent\textbf{Rollout.} Fig.~\ref{fig:overview}c summarizes the protocol. Every rollout runs in a fresh Docker sandbox with a common rendering toolchain (\eg, Python, Node, a headless browser, ffmpeg) and a TTS command-line tool backed by Qwen3-TTS-Flash~\cite{qwen3tts}, under a one-hour wall-clock budget. All model traffic is logged, so every trajectory, token and tool call is recorded, and each (system, task) has exactly one scored run; runs lost to an infrastructure fault are re-run. Details are in App.~\ref{app:sandbox}.\par}

\noindent\textbf{L1: delivery and program checks.} L1 is decided by programs alone; no model is involved. A run \emph{delivers} if it leaves an MP4 that decodes and can be judged; the \emph{delivery rate} is the share of tasks with such a video. Each delivery is then measured (ffprobe, a full decode, audio levels, VAD and Whisper language ID) and checked by code against the task's machine-readable delivery spec $d$ and audio policy $a$: every task checks container and decodability, resolution, frame rate ($\pm$0.01) and duration within $[t_{\min}{-}0.5, t_{\max}{+}0.5]$\,s. The 205 core tasks add a delivered source project, an exact frame count where $d$ requires it (54 tasks) and an audible audio stream (peak $>{-}50$\,dBFS) where $a$ labels a channel \emph{required}. Lessons and papers, which all require narration, add an audible audio stream, narration coverage (VAD), narration language (ASR) and the absence of long black or frozen stretches. This gives 1{,}449 checks per system (1{,}177 core, 272 lesson and paper). \emph{Spec compliance} is the share of tasks whose run passes all applicable checks; a non-delivery fails them all. (The rollout collector also runs a lighter container-level check at collection time, App.~\ref{app:sandbox}; we use it only in the behavior analysis of \S\ref{sec:behavior}.)

{\looseness=-1\noindent\textbf{L2: rubric checks.} For each check a judge returns pass, fail, unknown or n/a, with timestamped evidence. Judges score the 12{,}985 L2 checks (\S\ref{sec:rubric}), and we report two metrics. \emph{L2 pass} pools checks: it is $\#\text{pass}/(\#\text{pass}{+}\#\text{fail})$, with unknown and n/a excluded. \emph{L2 strict} (strict task success) is decided per run: a run succeeds if none of its core checks fails and at least one passes; unknown and n/a do not count against it, and a non-delivery fails. For example, the Super Mario level run (CG01) has 59 checks; a video in which Mario does not grow after touching the mushroom fails the core check ``growth is discernible after contact'', so the run fails L2 strict even if the other 58 pass (L2 pass 98\%); strict success is thus far more discriminative.\par}

{\looseness=-1\noindent\textbf{L3: quality.} The judge scores each of the five dimensions from 1 to 5 against the task's own anchors, with reasons, strengths and defects; a dimension the judge cannot assess is left \emph{unknown} and excluded. For CG01, timing and motion (TM) asks whether continuous motion, landings and state changes can be followed with enough time to see key dangers and successes; it scores 1 if key processes are hard to track, 3 if they can be tracked but transitions or holds are awkward, and 5 if key changes are clear and motion serves the task. The \emph{overall} score is the mean of the five dimensions, not a weighted total of L1--L3.\par}

\noindent\textbf{Delivered-only and no video $=0$.} Our primary display reports the delivery rate next to \emph{delivered-only} L2 and L3 scores, which isolate the quality of what was delivered. We also report every metric with \emph{no video $=0$}: a non-delivery scores zero on every dimension and fails every check, so reliability and quality combine into one number.

\begin{table*}[t]
\centering
\caption{\textbf{Main results on 239 tasks.} Judge-based scores (L2, L3) are means over the three judge panels (J1--J3, \S\ref{sec:eval}). \emph{L1}: playable videos (of 239) and spec compliance (\% of tasks passing every program check). \emph{Delivered videos}: L2 pass rate, L3 overall (1--5) and human Elo from blind pairwise preferences (\S\ref{sec:judge}), all on playable videos; rows follow L3 overall. \emph{No video $=0$}: a missing video fails; L2 strict is the share of tasks with no failed core check and at least one passed core check (\S\ref{sec:eval}); grey numbers are ranks. Full results: App.~\ref{app:panels}.}
\label{tab:main}
\definecolor{vvbscorewash}{HTML}{5A8BFC}\definecolor{vvbzerowash}{HTML}{E4AD4B}%
\definecolor{vvbink}{HTML}{232326}\definecolor{vvbsep}{HTML}{CBCCD4}%
\newcommand{\vvbbar}[3]{\begin{tikzpicture}[baseline=0.40ex,x=1cm,y=1cm]
  \fill[#1!10,rounded corners=0.8pt] (0,0) rectangle (0.95,0.29);
  \shade[left color=#1!47,right color=#1!18,rounded corners=0.8pt] (0,0) rectangle ({#3*0.95},0.29);
  \node[inner sep=0pt,text=vvbink] at (0.475,0.145) {\fontsize{8}{9}\selectfont #2};\end{tikzpicture}}%
\newcommand{\vvbbest}[3]{\begin{tikzpicture}[baseline=0.40ex,x=1cm,y=1cm]
  \fill[#1!10,rounded corners=0.8pt] (0,0) rectangle (0.95,0.29);
  \shade[left color=#1!59,right color=#1!26,rounded corners=0.8pt] (0,0) rectangle ({#3*0.95},0.29);
  \draw[#1!64,rounded corners=0.8pt,line width=0.4pt] (0,0) rectangle (0.95,0.29);
  \node[inner sep=0pt,text=vvbink] at (0.475,0.145) {\fontsize{8}{9}\selectfont\bfseries #2};\end{tikzpicture}}%
\newcommand{\vvbmodel}[2]{\makebox[0.34cm][c]{\raisebox{-0.15\height}{\includegraphics[width=0.26cm,height=0.26cm,keepaspectratio]{fig/icons/#1.pdf}}}\hspace{0.25em}#2}%
\footnotesize
\setlength{\tabcolsep}{4pt}
\renewcommand{\arraystretch}{1.1}
\begin{tabular}{@{}ll cc !{\color{vvbsep}\vrule width 0.35pt} ccc !{\color{vvbsep}\vrule width 0.35pt} cc@{}}
\toprule
\multirow{2}{*}{\textbf{Model}} & \multirow{2}{*}{\textbf{Harness}} & \multicolumn{2}{c!{\color{vvbsep}\vrule width 0.35pt}}{\textbf{L1 program checks}} & \multicolumn{3}{c!{\color{vvbsep}\vrule width 0.35pt}}{\textbf{Delivered videos}} & \multicolumn{2}{c}{\textbf{No video $=0$}} \\
\cmidrule(lr){3-4}\cmidrule(lr){5-7}\cmidrule(lr){8-9}
 & & Videos & Spec (\%) & L2 pass (\%) & L3 overall & Human Elo & L2 strict (\%) & L3 overall (rank) \\
\midrule
\csname @@input\endcsname tables/main_rows.tex
\bottomrule
\end{tabular}
\vspace{-0.6em}
\end{table*}

\noindent\textbf{Judge panels.} Each delivered video is judged independently by three panels, each given the exact task prompt and the rubric but never the system identity. Judges are told that on-screen or narrated self-declarations are not evidence, and that unsampled instants cannot be asserted.
\textbf{J1} (Gemini 4 Argon~\cite{google2026gemini4}) and \textbf{J2} (Gemini 3.8 Flash~\cite{google2026gemini38flash}) watch the full MP4 with sound (4\,fps up to 150\,s, adaptively lower for longer videos) and judge everything.
\textbf{J3} splits by modality following the check tags: GPT-5.6 Sol~\cite{openai2026gpt56} judges visual checks and VQ/VA/CL/TM from 24--300 sampled frames (up to 1280\,px), and Qwen3.8-Omni-Flash~\cite{qwen2026omni38flash} judges audio and audio-visual checks and AQ from a full-length proxy with the original audio. Audio observations are passed to the vision judge as reference.
All three panels contain evaluated models (J1, J2 and J3's vision judge); J3 shares no family with J1/J2. Each panel's metrics are computed separately and then averaged; scores are never mixed across panels within a metric ($3\times2{,}151$ run results). Tab.~\ref{tab:main} reports the three-panel mean (per panel: App.~\ref{app:panels}).

\noindent\textbf{Human preference.} To validate the judges we ran a blind pairwise study on all 239 tasks. Annotators saw the full prompt and two synchronized players with videos of two systems on the same task, both playable, and chose \emph{A}, \emph{B}, \emph{tie} or \emph{both bad} by overall quality and adherence. Human preference therefore measures the quality of delivered videos, not delivery. We fit Bradley--Terry strengths~\cite{bradley1952rank} on decisive votes (Elo scale, geometric mean 1000) with 2,000-sample bootstrap CIs and compare system ranks and pair outcomes with each panel.

\begingroup
\setlength{\parskip}{2pt}
\section{Experiments}
\label{sec:exp}

\noindent\textbf{Systems.} We evaluate nine agent systems from six labs, each in its native harness at the highest supported reasoning effort: GPT-6 Astra~\cite{openai2026gpt6astra} and GPT-5.6 Sol~\cite{openai2026gpt56} with Codex CLI~\cite{openai2025codexcli}; Claude Opus 5.5~\cite{anthropic2026opus55} and Sonnet 5.5~\cite{anthropic2026sonnet55} with Claude Code~\cite{anthropic2025claudecode}; Gemini 4 Argon~\cite{google2026gemini4} and Gemini 3.8 Flash~\cite{google2026gemini38flash} with Antigravity CLI~\cite{google2026antigravity}; DeepSeek-V4.1-Flash (1017)~\cite{deepseek2026v41flash} with DSH~\cite{deepseek2026dsh}; Kimi K3~\cite{moonshot2026kimik3} with Kimi Code~\cite{moonshot2026kimicode}; and GLM-5.3-Flash~\cite{zhipu2026glm53flash} with ZCode~\cite{zhipu2026zcode}. Each system completes the same 239 tasks autonomously, without follow-up user input, giving 2,151 rollouts.

\subsection{Main results}
\label{sec:main}

\noindent\textbf{Overall performance.} Astra and Opus lead in video quality and human preference, with L3 scores of 4.13 and 4.07, respectively (Table~\ref{tab:main}). However, high average performance does not ensure consistent task success. Opus delivers all 239 videos and achieves an L2 item-level pass rate of 97.3\%, yet its strict task-success rate is only 76.3\%, the highest among the nine systems. This gap persists even with complete delivery, showing that satisfying individual requirements remains insufficient for reliably completing an entire task. Strict success requires at least one passed core check and no failed core check; unknown checks are permitted.

\noindent\textbf{Delivery reliability.} Evaluating only completed videos can obscure substantial delivery failures. Gemini 4 Argon ranks fourth in delivered-video quality but produces playable videos for only 184 of 239 tasks. Counting non-deliveries as zero lowers its L3 score from 3.99 to 3.07 and its rank from fourth to seventh. Astra and Opus remain first and second. Video quality must therefore be interpreted alongside delivery reliability to assess overall task performance.

\subsection{Which requirements remain unmet?}
\label{sec:requirements}

Among 1,927 playable outputs, 613 (31.8\%) contain a core-check failure reported by at least two panels. We use this agreement criterion to identify errors in delivered videos; it differs from the panel-averaged strict success metric.

\begin{figure}[t]
\centering
\includegraphics[width=\columnwidth]{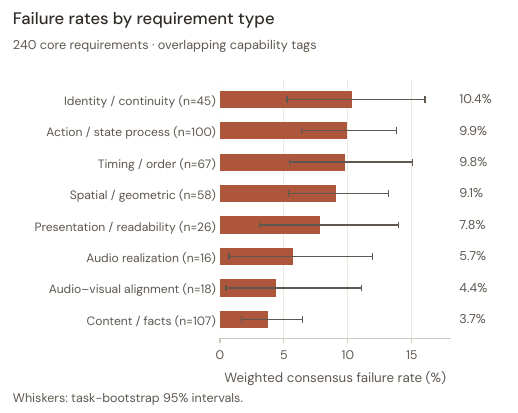}
\caption{\textbf{Failures by requirement type.} Weighted rates of failure reported by at least two panels for 240 sampled core requirements (1,913 eligible system--requirement instances). Tags overlap; unresolved verdicts are not counted as consensus failures.}
\label{fig:requirementtypes}
\end{figure}

{\looseness=-1\noindent\textbf{Processes, timing, and continuity.} To compare requirement types, we sample 20 core requirements from each task category (240 total) and assign overlapping capability tags before examining verdicts. After excluding majority n/a cases, 1,913 delivered system--requirement instances remain. Weighted failure rates are 9.9\% for actions and state processes, 9.8\% for timing, and 10.4\% for identity and continuity, compared with 3.7\% for content and facts (Fig.~\ref{fig:requirementtypes}). These descriptive comparisons use assistant-assigned tags and do not establish causal differences in difficulty.\par}

{\looseness=-1\noindent\textbf{Task complexity.} We analyze failures per core requirement, controlling for system and task category. More core requirements and longer target durations are associated with higher per-check failure rates.\par}

\subsection{Why videos are not delivered}
\label{sec:failures}

We read the records of all 224 runs that ended without a playable video and assign each one main cause (Fig.~\ref{fig:analysis}a). Three quarters of the failures (169) are final renders that started too late, ran too slowly or never started; the rest are stalled or interrupted renders (45), outputs the agent destroyed or lost (7), and other causes (3).

\begin{figure}[t]
\centering
\includegraphics[width=\columnwidth]{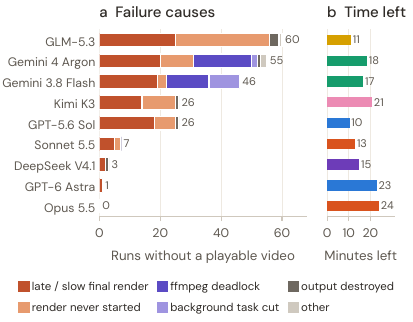}
\caption{\textbf{Why videos are not delivered.} (a) Main cause of the 224 runs without a playable video. (b) Median minutes left after the final write of successful runs on the 205 core tasks.}
\label{fig:analysis}
\end{figure}

{\looseness=-1\noindent\textbf{Time management.} Agents polish components and previews before launching the full render. In failed GPT-5.6 Sol runs, final renders start a median of nine minutes before the deadline, often using browser capture that takes 5--35 minutes. Across systems, 65 runs never start a final render, including 31 GLM runs. Another 33 renders stall in ffmpeg, and 12 are cut off after the agent ends its turn with a render still running. Output handling can also undo completed work: DeepSeek verifies a video twice, then re-encodes it in place one minute before the deadline and leaves a truncated file.\par}

\noindent\textbf{Successful runs leave margin.} Opus, the only system with no delivery failures, finishes its video a median of 24 minutes before the deadline, compared with 10--11 minutes for Sol and GLM (Fig.~\ref{fig:analysis}b). It first completes a video, then improves copies and verifies the final encode.

\begin{figure}[t]
\centering
\includegraphics[width=\columnwidth]{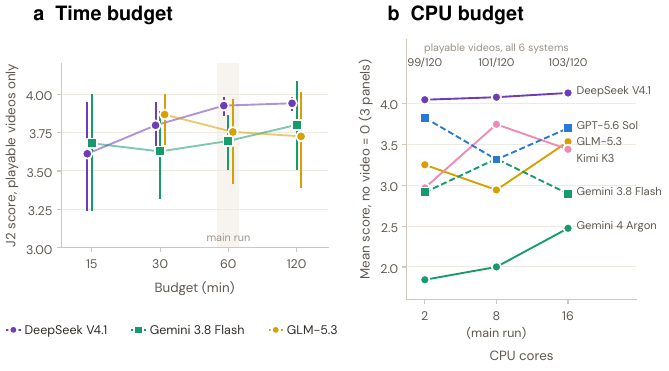}
\caption{\looseness=-1\textbf{Resource budgets.} (a) Time: J2 quality on playable videos, with 95\% CIs. (b) CPU: three-panel quality with non-deliveries scored zero. Main settings are 60 minutes and eight cores.}
\label{fig:budgets}
\end{figure}

\noindent\textbf{Time budget.} We test DeepSeek, Gemini 3.8 Flash and GLM on 20 tasks each at 15, 30, 60 and 120 minutes, once per setting. Within each system, delivered-video quality varies by at most 0.17 points, except for DeepSeek at 15 minutes (Fig.~\ref{fig:budgets}a). Means cover different delivered subsets.

\noindent\textbf{CPU budget.} Six systems run 20 render-heavy tasks with 2, 8 and 16 cores (360 runs). Pooled L3 scores, with non-deliveries scored zero, show no statistically significant gain from more cores (all paired-$t$ $p\geq0.20$; Fig.~\ref{fig:budgets}b).

\subsection{How agents make and revise videos}
\label{sec:behavior}

\noindent\textbf{Custom rendering pipelines.} Agents typically assemble their own pipelines from general-purpose graphics libraries. Across systems, ffmpeg appears in 96--100\% of runs, whereas Manim appears in only eight runs and MoviePy in 13. Claude agents and Gemini 4 Argon favor cairo; Sol, Kimi, and GLM often rasterize frames with PIL or numpy; Astra, DeepSeek, and Gemini 3.8 Flash frequently use browser canvases and three.js. In 57--96\% of each system's runs, music or effects are synthesized in code.

\begin{figure}[t]
\centering
\includegraphics[width=\columnwidth]{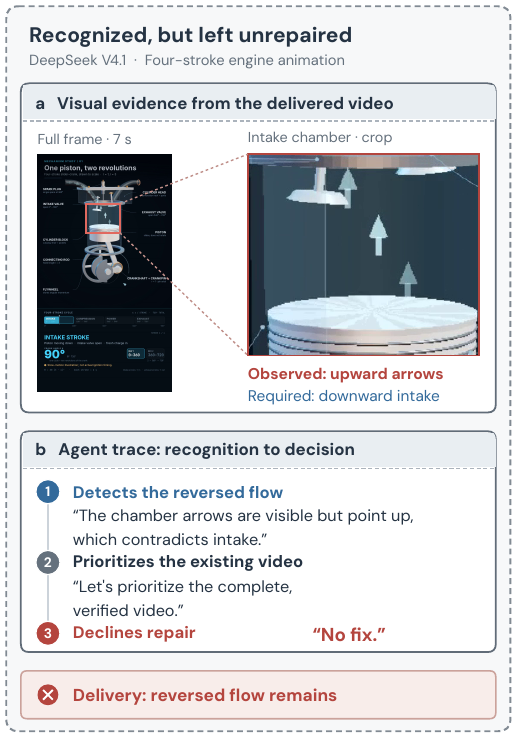}
\caption{\textbf{A detected defect can remain in the delivered video.} (a) DeepSeek's delivered frame at 7\,s and an unaltered crop show reversed intake arrows. (b) The agent recognizes the defect, prioritizes the existing complete video, and explicitly declines repair. The delivered file matches all three judge inputs. This selected case does not estimate a population repair rate.}
\label{fig:repaircase}
\end{figure}

\noindent\textbf{From inspection to a revised delivery.} We screen 54 selected failure traces, following one focal error per trace. In Fig.~\ref{fig:repaircase}, DeepSeek recognizes reversed intake arrows but prioritizes the existing complete video and declines repair. In another case, Kimi identifies dial occlusion, edits character placement, inspects an improved preview, and starts rerendering. The collected fallback video retains the occlusion; the trace does not establish completion of a replacement encode. These cases distinguish defect recognition, code editing, preview improvement, and final delivery. They do not estimate overall detection or repair rates, nor show that the second code edit was ineffective.

\subsection{Judges and humans}
\label{sec:judge}

\noindent\textbf{Agreement with human judgments.} We assess human agreement for both requirement judgments (L2) and video quality (L3). For L2, human review confirms 86.5\% of consensus visual failure judgments among resolved cases. For L3, the three-panel mean closely follows human system rankings ($\rho=0.98$) and agrees with human preferences on 73.3\% of decisive pairs after excluding judge ties (Tab.~\ref{tab:judge}). Agreement rises from 63\% for score gaps below 0.2 to 86\% for gaps of at least 0.5. Panels give similar mean quality scores (3.81) to pairs that humans label \emph{both bad} and to other pairs, so agreement in rankings does not certify the acceptability of each video.

\begin{table}[t]
\centering
\caption{\textbf{Judge agreement.} Top: pairwise panel agreement on 1,927 videos. Bottom: agreement with blind human preferences on delivered videos; system rank correlations use nine systems, and pair agreement excludes judge ties on human-decisive pairs.}
\label{tab:judge}
\setlength{\tabcolsep}{3pt}
\small
\renewcommand{\arraystretch}{1.12}
\begin{tabularx}{\columnwidth}{@{}>{\raggedright\arraybackslash}Xccc@{}}
\toprule
 & J1--J2 & J1--J3 & J2--J3 \\
\midrule
System Spearman $\rho$ (overall) & 0.950 & 0.967 & 0.983 \\
Check agreement / Cohen's $\kappa$ & 94.6\,/\,0.54 & 91.3\,/\,0.51 & 91.6\,/\,0.42 \\
Video-level Pearson $r$ (overall) & 0.44 & 0.49 & 0.50 \\
Same-task pair ordering \mbox{agreement (\%)} & 69.1 & 69.7 & 65.6 \\
\midrule
\end{tabularx}
\par\nointerlineskip
\begin{tabularx}{\columnwidth}{@{}>{\raggedright\arraybackslash}Xcccc@{}}
 & J1 & J2 & J3 & Mean \\
\midrule
Human vs.\ judge, system $\rho$ & 0.98 & 0.80 & 0.90 & \textbf{0.98} \\
Human vs.\ judge, pair agreement (\%) & 70.5 & 73.3 & 69.6 & \textbf{73.3} \\
Judge ties on human-decisive pairs (\%) & 18 & 46 & 14 & 6 \\
\bottomrule
\end{tabularx}
\end{table}

\noindent\textbf{Agreement between panels.} The three panels rank systems similarly (Spearman $\rho=0.95$--$0.98$), but agree less on individual videos: overall scores correlate at $r=0.44$--$0.50$, and same-task pair orderings agree 66--70\% of the time. Check-level agreement is 91--95\%, with Cohen's $\kappa=0.42$--$0.54$. These results support comparisons across many tasks while indicating greater uncertainty in individual judgments.

\endgroup

\section{Conclusion}
\label{sec:conclusion}

We introduced \bench, a benchmark for evaluating multimodal agents on video production through code. Its 239 tasks span 12 categories in English and Chinese, with evaluation covering delivery requirements, task-specific rubrics, and whole-video quality. Across nine frontier agent systems and 2,151 rollouts, GPT-6 Astra and Claude Opus 5.5 lead both the judge-based quality ranking and human preferences. However, the highest L2 strict success rate is only 76.3\%, and accounting for missing videos changes system rankings. Agents build custom rendering pipelines, yet their outputs still exhibit errors in required actions, timing, and continuity. Execution traces reveal further gaps in managing rendering time and turning identified defects into corrected final videos. These findings highlight the need to coordinate planning, execution, inspection, and revision throughout video production. \bench provides a testbed for improving agents' ability to turn detailed specifications into complete, coherent, and reliable \mbox{audiovisual deliverables}.
\label{endbody}

\section*{LLM Usage Statement}
The models specified in Sec.~\ref{sec:exp} and App.~\ref{app:judge} served as evaluated agents and judges, respectively, and OpenAI GPT assisted with language polishing, condensation, and rewriting of the main text and supplementary material.


\clearpage
\onecolumn
\appendix
\counterwithin{figure}{section}\counterwithin{table}{section}
\renewcommand{\theHfigure}{A\thesection.\arabic{figure}}
\renewcommand{\theHtable}{A\thesection.\arabic{table}}
\renewcommand{\floatpagefraction}{0.6}
\let\vvbsection\section\renewcommand{\section}{\FloatBarrier\vvbsection}
\addtocontents{toc}{\protect\setcounter{tocdepth}{2}}
\renewcommand{\contentsname}{Appendix}
\tableofcontents
\clearpage
\section{Dataset Construction and Statistics}
\label{app:data}

\subsection{Source streams and funnel}
Tab.~\ref{tab:funnel} traces the 239 tasks back to their sources.

\begin{table}[htbp]
\centering
\caption{\textbf{Construction funnel.} Counts at each stage; ``kept'' rates in parentheses.}
\label{tab:funnel}
\small
\resizebox{\linewidth}{!}{%
\begin{tabular}{@{}llr@{}}
\toprule
Stage & Description & Count \\
\midrule
\multicolumn{3}{@{}l}{\textit{Stream (i): web prompts}}\\
Collected & prompts written for code-based video tools (452 social posts, 25 Remotion gallery, 7 other) & 484 \\
Kept after strict review & every prompt read in full; nine removal codes & 152 (31\%) \\
\midrule
\multicolumn{3}{@{}l}{\textit{Stream (ii): dense captions of real videos}}\\
Scanned & caption records (HD-VILA, Molmo2, LLaVA-Video, ActivityNet sources) & 508,993 \\
Lexical recall & 26 content directions & 6,711 \\
LLM screen and close reading & graded screen, then full-caption review (keep 200, hold 109, reject 13) & 956 $\rightarrow$ 322 $\rightarrow$ 200 \\
Human decision & two annotators on the 200 kept; ``$\ge$1 keep and no discard'' & 84 (42\%) \\
\midrule
\multicolumn{3}{@{}l}{\textit{Authoring and selection}}\\
Drafted & drafts from the streams above, plus gap-filling drafts & 447 \\
Human selection & two selectors & 184 (41\%) \\
Final rewrite & removal of leaked authoring traces & 182 \\
\midrule
\multicolumn{3}{@{}l}{\textit{Extension}}\\
Added tasks & asset-grounded tasks (stream iii), classic-game tasks and tasks kept from a pilot & +29 \\
Removed on review & duplicates and an infeasible task & $-$6 \\
\textbf{Core tasks} & 199 English, 6 Chinese & \textbf{205} \\
\midrule
\multicolumn{3}{@{}l}{\textit{Stream (iv): lecture notes and papers}}\\
Lessons (ZH) & self-written lecture notes, seven subjects; 2 politics lessons removed & 24 \\
Papers (PA) & classic papers supplied as PDF (Transformer, ResNet, GAN, DDPM, NeRF, VAE, DQN, ViT, CLIP, LIGO) & 10 \\
\midrule
\textbf{Released and scored} & 209 English, 30 Chinese; every task judged by all three panels & \textbf{239} \\
\bottomrule
\end{tabular}}
\end{table}

\runin{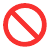}{What was removed.} Most web prompts failed because they presupposed material that was not provided (125 of 332 removals) or were content-free style templates (99); the rest were intros and logo stings (43), ambient loops without development (25), tasks that do not ask for a finished video (19) and duplicates (15); three rarer codes cover the remaining 6.

\runin{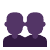}{Selection.} Human selectors preferred checkable tasks: they kept 88\% of the music and sound-driven drafts and about half of the explainers, but only 17\% of the story and comedy drafts and 15\% of the game drafts.

\runin{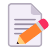}{Authoring.} LLM drafters with cross-review recorded each task's source, selected time range and all additions during authoring (facts, lines, durations, assets, structure). A final rewrite removed all authoring traces (reviewer notes, review status, scoring hints) from the prompts.

\subsection{Taxonomy}
Each task has exactly one of 12 \emph{categories} (content and purpose; Tab.~\ref{tab:taxonomy}) and one or more of 43 \emph{subcategories} that cut across categories (Tab.~\ref{tab:subcats}): 34 \emph{presentation forms}, the first of which is the task's primary form, and 9 \emph{knowledge domains}, tagged on tasks that convey subject knowledge (C04, C05, C11, C12). Tasks carry 2.0 subcategories on average; the per-task mapping is released with the benchmark.

\begin{table}[htbp]
\centering
\caption{\textbf{Categories} (239 tasks) with their most frequent subcategories.}
\label{tab:taxonomy}
\scriptsize
\resizebox{\linewidth}{!}{\begin{tabular}{@{}l r p{0.62\linewidth}@{}}
\toprule
Category & Tasks & Most frequent subcategories (tasks; multi-label)\\
\midrule
\cattag{C01}C01 Brand \& marketing & 9 & Contextual selling point \& call to action (7), Brand \& product showcase (3), Musical structure \& AV choreography (2), Visual metaphor (1)\\
\cattag{C02}C02 Product \& feature demos & 6 & Software \& device tutorials (4), Making, assembly \& modification (1), Musical structure \& AV choreography (1), System simulation \& causal modelling (1)\\
\cattag{C03}C03 Events \& personal expression & 7 & Event invitation (4), Personal expression \& tributes (3), Statistical charts \& comparison (1), Musical structure \& AV choreography (1)\\
\cattag{C04}C04 Concept \& pipeline explainers & 57 & Mechanism \& causal chain (26), Engineering \& machines (14), Physics \& astronomy (13), Mathematical derivation \& counting (12)\\
\cattag{C05}C05 Data \& fact stories & 24 & History \& society (12), Statistical charts \& comparison (11), History \& chronology (11), Geography \& earth science (8)\\
\cattag{C06}C06 Visual art \& form & 26 & Spatial tour \& scene experience (9), Character performance \& choreography (5), Visual metaphor (5), Abstract form \& scale (3)\\
\cattag{C07}C07 Games \& sports & 19 & Real-time follow-along guidance (7), Tactics \& strategy breakdown (6), Classic game reconstruction (4), Observe--guess--reveal (2)\\
\cattag{C08}C08 Story \& visual comedy & 15 & Escalating missions \& crises (8), Musical structure \& AV choreography (2), Character performance \& choreography (2), Dialogue \& punchline (2)\\
\cattag{C09}C09 Building \& process & 23 & Making, assembly \& modification (23), Mechanism \& causal chain (5), Spatial tour \& scene experience (4), Interactive model, structure \& scale exploration (1)\\
\cattag{C10}C10 Music \& sound-driven & 19 & Musical structure \& AV choreography (18), Character performance \& choreography (7), Real-time follow-along guidance (2), Abstract form \& scale (2)\\
\cattag{C11}C11 Classroom lessons & 24 & Mechanism \& causal chain (12), Mathematical derivation \& counting (8), Mathematics (8), Interactive model, structure \& scale exploration (6)\\
\cattag{C12}C12 Paper explainers & 10 & Artificial intelligence (9), System architecture \& information flow (5), Mechanism \& causal chain (4), System simulation \& causal modelling (2)\\
\midrule
Total & 239 & 205 core tasks + 24 classroom lessons (C11) + 10 paper explainers (C12)\\
\bottomrule
\end{tabular}

}
\end{table}

\begin{table}[htbp]
\centering
\caption{\textbf{Subcategories} (multi-label). $n$: tasks carrying the subcategory; Cat.: number of categories it appears in.}
\label{tab:subcats}
\scriptsize
\resizebox{\linewidth}{!}{\begin{tabular}{@{}l p{0.30\linewidth} r r @{\hspace{1.2em}} l p{0.30\linewidth} r r@{}}
\toprule
ID & Presentation form & $n$ & Cat. & ID & Presentation form & $n$ & Cat.\\
\midrule
S09 & Mechanism \& causal chain & 51 & 6 & S27 & Abstract form \& scale & 6 & 3\\
S29 & Musical structure \& AV choreography & 28 & 7 & S28 & Visual metaphor & 6 & 2\\
S07 & Making, assembly \& modification & 25 & 3 & S38 & Worked examples \& problem solving & 6 & 2\\
S08 & Mathematical derivation \& counting & 23 & 4 & S34 & Conceptual frameworks \& decision paths & 5 & 1\\
S14 & Statistical charts \& comparison & 18 & 5 & S01 & Software \& device tutorials & 4 & 1\\
S10 & System architecture \& information flow & 17 & 3 & S25 & Time-span, identity \& emotional change & 4 & 2\\
S16 & Interactive model, structure \& scale exploration & 17 & 4 & S32 & Event invitation & 4 & 1\\
E04 & Character performance \& choreography & 15 & 4 & CG & Classic game reconstruction & 4 & 1\\
E01 & Spatial tour \& scene experience & 14 & 3 & E05 & Poetry \& emotion & 3 & 2\\
S05 & System simulation \& causal modelling & 14 & 6 & S23 & Journey \& arrival & 3 & 3\\
S13 & History \& chronology & 13 & 2 & S26 & Kinetic typography \& channel packaging & 3 & 3\\
E02 & Real-time follow-along guidance & 11 & 3 & S31 & Brand \& product showcase & 3 & 1\\
S03 & Map journeys \& route tours & 10 & 4 & S37 & Personal expression \& tributes & 3 & 1\\
S22 & Escalating missions \& crises & 10 & 3 & E07 & Miniature \& fictional worlds & 2 & 1\\
S04 & Tactics \& strategy breakdown & 8 & 2 & S06 & Observe--guess--reveal & 2 & 1\\
S12 & Experiments \& evidence reading & 8 & 4 & S24 & Dialogue \& punchline & 2 & 1\\
S33 & Contextual selling point \& call to action & 7 & 1 & S18 & Picture-book reading \& early concepts & 1 & 1\\
\midrule
ID & Knowledge domain & $n$ & Cat. & ID & Knowledge domain & $n$ & Cat.\\
\midrule
D01 & Mathematics & 17 & 2 & D06 & History \& society & 18 & 3\\
D02 & Physics \& astronomy & 18 & 4 & D07 & Computing \& information & 12 & 2\\
D03 & Chemistry \& materials & 6 & 2 & D08 & Artificial intelligence & 13 & 3\\
D04 & Biology \& medicine & 11 & 3 & D09 & Engineering \& machines & 15 & 2\\
D05 & Geography \& earth science & 12 & 3 & & & &\\
\bottomrule
\end{tabular}

}
\end{table}

\subsection{Task format and statistics}
Agents receive only the task prompt and input files; source records and audit notes remain internal, and only judges receive the rubric.
\noindent A typical prompt gives the audience and purpose, a structure line, the content inputs (facts, numbers, real or fictional assets), the progression with minimum dwell times, the four-channel audio policy, the creative freedom and the delivery list (MP4, editable project, a timing sheet and production notes). Tab.~\ref{tab:taskstats} summarizes the tasks.

\begin{table}[htbp]
\centering
\caption{\textbf{Task statistics} (239 tasks; core and lesson/paper subsets in brackets).}
\label{tab:taskstats}
\small
\begin{tabular}{@{}lp{0.70\linewidth}@{}}
\toprule
Language & 209 English / 30 Chinese [core 199 / 6; lessons 24 Chinese; papers 10 English]\\
Prompt length & core English prompts: median 1,638 words, IQR 967--2,179, range 503--5,581; lesson and paper prompts are short (median 262 words/characters) and carry their content in the supplied note or PDF\\
Resolution / fps & 1920$\times$1080: 227; other: 12; 30\,fps: 228, 60\,fps: 11; exact frame count: 54\\
Target duration & midpoint median 118\,s, IQR 55--270\,s, range 8--894\,s; lessons 180--480\,s; 38 tasks exceed 5 minutes; 11.1 hours per pass\\
Audio policy & $\ge$1 required channel: 168 (70\%) [core 134 (65\%)]; speech required: 117 (49\%); every lesson and paper requires narration\\
Input assets & 201 tasks [core 167, all 34 lessons and papers]; 2.69\,GB\\
Rubric & 12,985 L2 checks [core 10,813; lessons and papers 2,172]; mean 54.3, median 54, range 13--110\\
\bottomrule
\end{tabular}

\end{table}

\section{Rubric Design}
\label{app:rubric}

\subsection{Structure of a rubric}
Each rubric groups the task checks (L2) under a few core goals, maps every requirement of the prompt to the checks that enforce it, and adds the task's quality rubric (L3), whose five dimensions are anchored at 1, 3 and 5 by descriptions written for that task. A check specifies pass/fail criteria, where to seek evidence in the video, the prompt sentence it enforces, and criticality and modality tags (Tab.~\ref{tab:rubricex}). Over the 239 tasks there are 12,985 checks (54.3 per task, 13--110); 85.8\% are \emph{core}, the rest \emph{detail} (\emph{supporting} for lessons and papers). Unknown is reserved for insufficient evidence: ``confirmed omissions or violations in the complete work are a fail and must not be concealed with unknown''.

\begin{table}[htbp]
\centering
\caption{\textbf{Example task checks} from CG01 (Super Mario; 7 core goals, 59 checks).}
\label{tab:rubricex}
\scriptsize
\resizebox{\linewidth}{!}{%
\begin{tabular}{@{}p{0.07\linewidth}p{0.12\linewidth}p{0.26\linewidth}p{0.26\linewidth}p{0.2\linewidth}@{}}
\toprule
ID / tag & Title & Pass & Fail & Observation scope \\
\midrule
V05 core, visual & Initial small form & Mario is small at departure and does not possess large-form abilities without cause during the initial stage. & Starts in the large form and skips the first growth, or has already grown before acquiring the mushroom. & From departure until before the first mushroom contact \\
V12 core, visual & Large form breaks a brick from below & After becoming large, Mario hits BRICK1 or BRICK2 from below and causes at least one to shatter. & Breaking the brick early in small form, without contact, or hitting only the question block. & After the transformation until the first brick breaks \\
V30 core, visual & Complete jumps & Required jumps have traceable takeoff, airborne movement and landing/contact. & Pose changes impersonate displacement; teleporting to the landing point. & All required jumps \\
V41 core, visual & Order of key events & Q1 mushroom $\rightarrow$ brick $\rightarrow$ G1 stomp $\rightarrow$ shell kicked into G2 $\rightarrow$ hit by G3 $\rightarrow$ lift over pit $\rightarrow$ flagpole and castle. & Reordering events, or montage that hides the actual sequence. & Whole video \\
V50 core, audio-visual & Jump SFX synchronization & A jump cue corresponding to the action is present at each required takeoff. & No cue, or the effect occurs during an unrelated action. & Every required takeoff and the audio track \\
\bottomrule
\end{tabular}}
\end{table}

\noindent\textbf{Quality rubric.} The five L3 dimensions share 20 observation points:
\par\vspace{3pt}
{\small\noindent\begin{tabular}{@{}l llll@{}}
\toprule
 & 1 & 2 & 3 & 4\\
\midrule
VQ & key-content legibility & stability & occlusion/cropping/layering & object consistency\\
VA & hierarchy & composition & unity of visual language & audience fit\\
CL & organization & key relations & information load & channel coordination\\
TM & action & state change and transitions & dwell on key content & overall rhythm\\
AQ & intelligibility and pronunciation & prosody & mixing & sound--event alignment\\
\bottomrule
\end{tabular}}
\par\vspace{3pt}
\noindent Each dimension has task-specific anchors at 1, 3 and 5 (\eg, VQ for CG01: 1 = ``key content remains unclear or is disrupted by visual defects'', 5 = ``key content clear, image stable, details support viewing''); a 2 or 4 must say which anchor it is closer to, and insufficient evidence yields \emph{unknown}, never a default 3. Intentional pixel-art style, grain or blur is not a defect.

\subsection{Authoring principles}
\begin{itemize}[leftmargin=1.2em,itemsep=1pt,topsep=2pt]
\item \textbf{Traceability.} Every check quotes the prompt sentence it enforces, and optional language (``may'', ``e.g.'') never becomes mandatory.
\item \textbf{Process over claims.} A correct final number or a caption reading ``synced'' cannot pass a check that requires a visible process: in a marble adder the sum must be produced by actual contacts.
\item \textbf{Criticality fixed in advance.} Core and detail are fixed before any output is seen, so cosmetic passes cannot outvote a failed core goal.
\item \textbf{Timing under sampling.} ``At least $N$ seconds'' passes only if the lower bound of the sampled interval is ${\ge}N$, fails if the upper bound is ${<}N$, and is otherwise unknown.
\item \textbf{No prescribed method.} 40--50 checks is a coverage aim, not a quota, and no check may prescribe a production method.
\end{itemize}

\subsection{Rubrics for lessons and papers}
\label{app:edurubric}
The 34 lessons and papers (C11, C12) use the same levels and judge interface, but their checks come from the supplied material. \textbf{L1} adds narration coverage (VAD speech share ${\ge}50\%$, no gap over 20\,s), narration language (ASR, ${\ge}90\%$ in the target language) and no long black or frozen stretch. \textbf{L2} has 48--83 checks per task (2,172 in total; 1,566 core): \emph{content checks} enumerate the knowledge points (each concept, derivation step, worked answer, common misconception and relation to be drawn; for papers the motivation, key idea, method components, key equations, results, evidence and limitations), and \emph{narration checks} require correct reading of terms, formulas and numbers, explanation rather than reading on-screen text aloud, and synchronization at 4--6 key moments. Because judges never see the note or the paper, 330 checks that referred to them (``consistent with the note'') were rewritten to state the required content, and six audit agents re-derived every number and formula against the source. Each task also has a grounded quiz (404 questions), reported separately (App.~\ref{app:x14}).

\subsection{Quality control}
Rubrics were refined iteratively: in each round, an independent model reviewed each rubric against its prompt, each finding was re-adjudicated in a fresh context and verified by a human reviewer, and confirmed issues were corrected. Review continued until no confirmed issue remained. The most frequent defect was over-specification (a rubric demanding more than the prompt), ahead of coverage gaps; the review also surfaced bugs in the tasks themselves, such as an infeasible inventory or contradictory timing tables, which were fixed in the prompts.

\subsection{Modality tags}
Each check is tagged with the evidence needed to decide it: visual 9,006 (69.4\%), audio-visual 3,012 (23.2\%) and audio 967 (7.4\%). The tags route checks in J3 (App.~\ref{app:judge}).

\section{Sandbox, Harnesses and Rollout Protocol}
\label{app:sandbox}

\begin{table}[htbp]
\centering
\caption{\textbf{Evaluated systems.} Each runs in its own native harness in unattended mode at the highest effort the provider accepted.}
\label{tab:systems}
\small
\begin{tabular}{@{}llll@{}}
\toprule
System & Lab & Harness & Effort \\
\midrule
GPT-6 Astra & OpenAI & Codex CLI 0.162 & max \\
GPT-5.6 Sol & OpenAI & Codex CLI 0.162 & max \\
Claude Opus 5.5 & Anthropic & Claude Code 2.1.301 & max \\
Claude Sonnet 5.5 & Anthropic & Claude Code 2.1.301 & max \\
Gemini 4 Argon & Google & Antigravity CLI & high \\
Gemini 3.8 Flash & Google & Antigravity CLI & high \\
DeepSeek-V4.1-Flash & DeepSeek & DSH 0.2.1 & max \\
Kimi K3 & Moonshot & Kimi Code 2.2.0 & max \\
GLM-5.3-Flash & Zhipu & ZCode 0.17.0 & max \\
\bottomrule
\end{tabular}
\end{table}

\runin{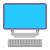}{Container.} Each run uses a fresh container from one image (Ubuntu 24.04) with 8 CPUs, 32\,GB of memory and no GPU; the task's input files are read-only.

\runin{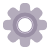}{Preinstalled tooling.} Python and Node with common imaging and browser-automation libraries, headless Chromium with software WebGL, ffmpeg, ImageMagick, SoX, three.js, GSAP and a large CJK and Latin font set. Internet access goes through a proxy, and package installation is allowed. A text-to-speech tool (Qwen3-TTS-Flash) is provided; no other generative APIs are available.

\runin{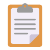}{Instruction.} The harness prompt is fixed: ``Read ENVIRONMENT.md and prompt.txt. Complete only that task autonomously using your native harness before the deadline\ldots'' \texttt{ENVIRONMENT.md} states the 3,600\,s budget, which includes model waits, tools, TTS, previews and the final render. It imposes no token, tool-call or preview limits and lists the deliverables (final MP4, editable project, timing sheet and production notes). It asks the agent to ``finish with honest verification results''. At the deadline the container is killed.

\section{Judge Protocol and Judge Analysis}
\label{app:judge}

\subsection{Protocol}
Each judge receives the same prompt as the agent, task checks and quality rubric, but never the system identity, source code or agent notes. For each check, it returns pass, fail, unknown or n/a with the supporting time range and channel (visual or audio), and scores the five quality dimensions against the task's anchors with justifications. On-screen or narrated claims (``\checkmark\ requirement met'') are not evidence, properties of the delivered file are left to the program checks, a judge that sees sampled frames may not assert what happens between them, and insufficient evidence yields unknown.

J1 (Gemini 4 Argon) and J2 (Gemini 3.8 Flash) evaluate all checks and quality dimensions using the full video with audio. J3 splits the work by modality tag: GPT-5.6 Sol judges the visual checks and VQ, VA, CL and TM from sampled frames, and Qwen3.8-Omni-Flash judges the audio and audio-visual checks and AQ from the full video with its original sound. Each of the 1,927 playable videos was judged by all three panels. A run without a playable video scores zero on every dimension and fails every check (\S\ref{sec:eval}).

\runin{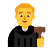}{Example judge output} (J1, CG01, Claude Opus 5.5):
\begin{tcolorbox}[colback=gray!4,colframe=gray!40,boxrule=0.4pt,left=3pt,right=3pt,top=2pt,bottom=2pt]
\scriptsize
\textbf{CG01-V41 pass.} ``The events occur in exact order: Q1 hit \& mushroom power-up (9.0--15.75\,s) $\rightarrow$ brick break (21.0\,s) $\rightarrow$ G1 stomp (22.75\,s) $\rightarrow$ K1 stomp, kick \& G2 defeat (30.25--33.0\,s) $\rightarrow$ G3 side hit (41.0\,s) $\rightarrow$ MOVE1 crossing (47.5--54.0\,s) $\rightarrow$ flagpole \& castle (61.75--68.5\,s).''\\[2pt]
\textbf{CG01-V50 pass} (audio). ``A rising pitch jump sound effect plays synchronously at each of Mario's takeoffs (\eg, 6.0\,s, 8.75\,s, 17.5\,s, 20.5\,s, \ldots, 61.75\,s).''\\[2pt]
\textbf{VQ = 4.} ``Closer to 5 than 3 because all characters, states, HUD counters, and contact points are crisp and stable, with only minor visual overlap during the side-hit protection pass.''
\end{tcolorbox}

\subsection{Leniency and scale use}
\begin{table}[htbp]
\centering
\caption{\textbf{Judge scale use} (239 tasks). L3, L2 and unknown: averages over the nine systems (delivered runs). Leniency: mean of the panel's five-dimension score minus the mean of the other two panels on the same run (1,927 runs). Spread: best minus worst system overall. Right: distribution of all dimension scores (\%; unk.: unknown or n/a).}
\label{tab:leniency}
\small
\begin{tabular}{@{}l cccc cc@{\hspace{1.5em}} ccccc c@{}}
\toprule
Panel & L3 (deliv.) & L2 pass & Unknown & Leniency & Spread nv$=$0 & Spread deliv. & 1 & 2 & 3 & 4 & 5 & unk.\\
\midrule
J1 & 3.61 & 90.9\% & 0.9\% & $-$0.41 & 1.47 & 0.89 & 0.8 & 6.0 & 28.6 & 57.4 & 6.0 & 1.3 \\
J2 & 4.01 & 95.7\% & 0.4\% & $+$0.19 & 1.19 & 0.46 & 0.2 & 2.5 & 9.2 & \textbf{69.9} & 17.3 & 0.8 \\
J3 & 4.03 & 88.2\% & 6.3\% & $+$0.22 & 1.39 & 0.65 & 2.1 & 2.2 & 16.3 & 46.8 & 31.4 & 1.2 \\
\bottomrule
\end{tabular}

\end{table}
\noindent\textbf{Panels differ in strictness, especially on weak systems.} J1 is the harshest on L3 ($-$0.41 against the other two panels) and the most discriminative (spread 0.89 on delivered videos). J2 is lenient and compressed: 70\% of its dimension scores are a 4 and its delivered-only spread is 0.46. J3 is the harshest on checks (L2 88.2\%; it also leaves 6.3\% unknown) but the most generous on VA and AQ ($+$0.46 and $+$0.61 against the other two). Leniency is not uniform: J1 is harsher on the weaker systems (GPT-5.6 Sol $-$0.67, Gemini 3.8 Flash $-$0.59, against $-$0.24 for Astra and Opus), and J2 and J3 are more lenient on them (J2: Sol $+$0.35, Flash $+$0.31, Astra $-$0.02). This is why panels are never mixed within a metric and only the three-panel mean is compared across systems.

\subsection{Self-preference}
\label{app:x10}
The panels use models also evaluated as agents: J1 is Gemini 4 Argon, J2 is Gemini 3.8 Flash, and J3's vision judge is GPT-5.6 Sol. Tab.~\ref{tab:selfpref} compares each panel's ratings of a system with those of the other two panels using the \emph{relative score} defined in the caption.

\begin{table}[htbp]
\centering
\caption{\textbf{Self-preference} on the 1,886 videos that all panels scored on the four visual dimensions (VQ, VA, CL, TM; for J3 these come from GPT-5.6 Sol alone). Relative score: the panel's score, centered on its mean over all videos, minus the same centered score averaged over the other two panels; positive = the panel rates the system higher than the other two panels do. Shaded: the panel's own model. 95\% CIs from a 2,000-sample bootstrap over videos.}
\label{tab:selfpref}
\small
\begin{tabular}{@{}l ccc@{}}
\toprule
System & J1 (Gemini 4 Argon) & J2 (Gemini 3.8 Flash) & J3 vision (GPT-5.6 Sol) \\
\midrule
GPT-6 Astra & $+0.17$ [0.11, 0.23] & $-0.23$ [$-$0.29, $-$0.17] & $+0.06$ [$-$0.02, 0.13] \\
Claude Opus 5.5 & $+0.21$ [0.15, 0.27] & $-0.11$ [$-$0.17, $-$0.04] & $-0.10$ [$-$0.18, $-$0.03] \\
DeepSeek-V4.1-Flash & $-0.03$ [$-$0.10, 0.05] & $+0.05$ [$-$0.02, 0.13] & $-0.03$ [$-$0.09, 0.04] \\
Claude Sonnet 5.5 & $+0.18$ [0.11, 0.25] & $-0.08$ [$-$0.15, $-$0.01] & $-0.10$ [$-$0.17, $-$0.03] \\
Kimi K3 & $+0.02$ [$-$0.04, 0.08] & $+0.03$ [$-$0.04, 0.10] & $-0.05$ [$-$0.12, 0.03] \\
GPT-5.6 Sol & $-0.32$ [$-$0.40, $-$0.24] & $+0.12$ [0.05, 0.20] & \colorbox{best}{$+0.20$ [0.12, 0.27]} \\
Gemini 4 Argon & \colorbox{best}{$+0.03$ [$-$0.05, 0.10]} & $+0.01$ [$-$0.08, 0.09] & $-0.03$ [$-$0.11, 0.04] \\
Gemini 3.8 Flash & $-0.22$ [$-$0.28, $-$0.15] & \colorbox{best}{$+0.16$ [0.09, 0.23]} & $+0.05$ [$-$0.02, 0.12] \\
GLM-5.3-Flash & $-0.16$ [$-$0.25, $-$0.07] & $+0.13$ [0.05, 0.22] & $+0.03$ [$-$0.05, 0.11] \\
\midrule
Own model vs.\ others & $+0.03$ [$-$0.05, 0.11] & $+0.18$ [0.10, 0.26] & $+0.22$ [0.14, 0.29] \\
Own family vs.\ others & $-0.12$ [$-$0.18, $-$0.06] & $+0.11$ [0.05, 0.17] & $+0.16$ [0.10, 0.22] \\
\bottomrule
\end{tabular}

\end{table}
\noindent\textbf{Two of three judges favor their own model, by about a sixth of a point.} J2 rates the Gemini 3.8 Flash agent $+$0.16 and J3's vision judge rates the GPT-5.6 Sol agent $+$0.20 higher than the other panels do; both intervals exclude zero. J1 shows no preference for the Gemini 4 Argon agent ($+$0.03) and rates the Gemini family \emph{lower} than the other panels ($-$0.12). Panels also show preferences of similar magnitude for other models: J1 favors the Claude agents and Astra ($+$0.17 to $+$0.21) and disfavors Sol ($-$0.32), and J2 disfavors Astra ($-$0.23) while being lenient on the weaker systems (GLM $+$0.13, Sol $+$0.12), so part of its $+$0.16 for Flash is its general leniency toward weaker systems.

\noindent\textbf{The ranking does not depend on it.} Scoring each system without the panel that is its own model (J1 for Argon, J2 for Flash, J3 for Sol) changes only the last two places: Flash and GLM, 0.008 apart on the pooled leaderboard (Tab.~\ref{tab:full}) and statistically tied, swap (Kendall $\tau{=}0.94$ with the headline ranking). Scoring each system only with panels outside its family (the Gemini agents by J3 alone, the GPT agents by J1 and J2), after shifting every panel by its leniency offset on the commonly judged videos ($-$0.27, $+$0.13 and $+$0.14 against the three-panel mean; the leniency of Tab.~\ref{tab:leniency}, which is measured against the mean of the other two panels, rescaled by $2/3$), leaves the order unchanged ($\tau{=}1$): the Astra--Opus gap narrows from 0.034 to 0.018 (paired bootstrap $P(\text{Astra}{>}\text{Opus}){=}0.74$), and Sol stays above Argon ($P{=}0.95$). Without the shift Opus overtakes Astra, but only because the panel dropped for the GPT agents (J3) is a lenient one.

\subsection{Historical retest and aggregation}
\label{app:judgeretest}
\label{app:judgeaggregation}
We reassess 30 videos spanning nine systems with each of the three panels, retaining the historical audio judgments in the GPT-based panel. After filtering for input and protocol consistency, mean L3 changes are $+0.510$ for Flash ($n{=}29$), $-0.043$ for Argon ($n{=}28$), and $-0.036$ for the GPT-based hybrid panel ($n{=}28$). On the 27 videos eligible for all three panels, the three-panel mean changes by $+0.146$ (95\% paired-bootstrap CI: $[+0.047,+0.247]$; 10,000 resamples), with a per-video mean absolute change of 0.249. Panel-averaged strict success changes on five videos and decreases by 1.23 percentage points overall. Averaging therefore does not fully remove the observed score shift. Date and provider changes are confounded, and this subset does not establish the stability of the full system ranking.

\section{Full Results}
\label{app:panels}
Results cover 239 tasks; metrics are computed separately for each of the three panels, then averaged unless a panel is specified.

\subsection{Full leaderboard}
Tab.~\ref{tab:full} reports the complete main table: all five L3 dimensions, spec compliance, the L2 pass rate with no video $=0$, human Bradley--Terry ratings and API cost.

\begin{table}[htbp]
\centering
\caption{\textbf{Full leaderboard on 239 tasks} (mean of three judge panels). L1: playable-delivery rate and spec compliance (all program checks pass: the checks of Tab.~\ref{tab:l1} on the 205 core tasks and the rule-based checks on lessons and papers; a non-delivery fails). L2: check pass rate; strict success = share of tasks whose core checks have no fail and at least one pass. L3: per-dimension and overall quality (1--5). Rows are ordered by delivered-only overall; the grey number is the rank with no video $=0$. Human: Bradley--Terry Elo from blind pairwise preferences between playable videos (App.~\ref{app:human}). \$/task: mean list-price API cost over the 239 tasks (App.~\ref{app:cost}). Best in \colorbox{best}{\textbf{bold}}, second \underline{underlined} (ties on the shown digits share the mark).}
\label{tab:full}
\definecolor{labOAI}{HTML}{2A78D6}\definecolor{labANT}{HTML}{D9531E}\definecolor{labGOO}{HTML}{169C6C}%
\definecolor{labDS}{HTML}{6D3CB8}\definecolor{labMS}{HTML}{EC8BB4}\definecolor{labZP}{HTML}{D6A000}%
\def\logolabOAI{openai}\def\logolabANT{claude}\def\logolabGOO{gemini}\def\logolabDS{deepseek}\def\logolabMS{kimi}\def\logolabZP{zai}%
\newcommand{\lab}[1]{\logo{\csname logo#1\endcsname}\hspace{0.32em}}%
\setlength{\tabcolsep}{3.1pt}
\footnotesize
\resizebox{\textwidth}{!}{%
\begin{tabular}{@{}lll r cc c ccccc c ccc c@{}}
\toprule
& & & & \multicolumn{2}{c}{L1 (\%)} & \multicolumn{7}{c}{\textbf{Delivered videos only}} & \multicolumn{3}{c}{No video $=0$} & \\
\cmidrule(lr){5-6}\cmidrule(lr){7-13}\cmidrule(lr){14-16}
& & & & & & L2 (\%) & \multicolumn{6}{c}{L3 quality (1--5)} & L2 (\%) & L2 (\%) & L3 & \\
\cmidrule(lr){7-7}\cmidrule(lr){8-13}\cmidrule(lr){14-14}\cmidrule(lr){15-15}\cmidrule(lr){16-16}
\# & System & Harness & \$/task & Deliv. & Spec & Pass & VQ & VA & CL & TM & AQ & \textbf{Overall} & Pass & Strict & Overall\,(rank) & Human Elo \\
\midrule
\csname @@input\endcsname appendix/tables/full_rows_239.tex
\bottomrule
\end{tabular}}
\vspace{-0.6em}
\end{table}

\noindent\textbf{Strict metrics and core vs.\ detail.} Core checks are harder than detail checks for every system (Astra 97.0 vs.\ 97.8\%; Gemini 3.8 Flash 79.4 vs.\ 86.7\%). The share of tasks with no failed check of any kind (no video counts as failure) ranges from 73.6\% (Opus) and 72.1\% (Astra) to 23.8\% (Gemini 3.8 Flash), close to strict success (76.3, 74.5 and 26.6\%): most tasks that fail strict success fail a core check, not only detail checks. Causes of strict-success failure vary by system: for Opus, Astra, Sonnet and DeepSeek it is almost always a failed core check, whereas 19--25\% of the runs of Gemini 4 Argon, GLM and Gemini 3.8 Flash fail it by delivering no playable video.

\subsection{Per-panel leaderboards}
Tab.~\ref{tab:panels} reports each panel separately. The panels differ in leniency but largely agree on the order: the top four are the same in every panel (Astra, Opus, DeepSeek, Sonnet), followed by Kimi and GPT-5.6 Sol. Their disagreements are between near-ties: J1 and J3 put Astra first (3.946 vs.\ 3.915; 4.285 vs.\ 4.172) and J2 puts Opus first (4.137 vs.\ 4.095); at the bottom J1 ranks GLM above Gemini 3.8 Flash (2.529 vs.\ 2.478), while J2 and J3 rank Flash above GLM. J3, whose vision judge sees sampled frames, leaves far more checks unknown (5.4--7.0\% of delivered checks, against 0.6--1.7\% for J1 and 0.2--0.6\% for J2), which is why the ``all core'' rate of Tab.~\ref{tab:panels}, where an unknown counts against the task, is much lower under J3 and why strict success counts only failures. Even so, the panels differ widely in strictness (Opus: 83.7, 93.3 and 51.9\% strict success under J1--J3), so only the three-panel mean is comparable across systems (App.~\ref{app:judge}).

\begin{table}[htbp]
\centering
\caption{\textbf{Per-panel results} (239 tasks; rows in the panel's no-video-$=0$ order). nv$=$0: no video $=0$. Task success: \% of tasks whose core checks all pass or are n/a, with at least one pass (``all core'', unknown counts against), and strict success (no core check fails, at least one passes); no video counts as failure. Unk.: unknown share of delivered checks.}
\label{tab:panels}
\resizebox{\linewidth}{!}{\begin{tabular}{@{}rl rr rr ccccc cc r@{}}
\toprule
\# & System & \multicolumn{2}{c}{L2 pass (\%)} & \multicolumn{2}{c}{Task success (\%)} & \multicolumn{5}{c}{L3 dims (delivered only)} & \multicolumn{2}{c}{Overall} & Unk.\\
\cmidrule(lr){3-4}\cmidrule(lr){5-6}\cmidrule(lr){7-11}\cmidrule(lr){12-13}
 & & deliv. & nv$=$0 & all core & strict & VQ & VA & CL & TM & AQ & deliv. & nv$=$0 & (\%)\\
\midrule
\multicolumn{14}{@{}l}{\textit{J1: Gemini 4 Argon (full video + audio)}}\\
1 & GPT-6 Astra & 98.4 & 98.2 & 76.6 & 83.7 & 4.00 & 4.06 & 4.07 & 3.87 & 3.80 & 3.963 & 3.946 & 0.6 \\
2 & Claude Opus 5.5 & 98.4 & 98.4 & 76.6 & 83.7 & 3.92 & 4.01 & 4.04 & 3.92 & 3.68 & 3.915 & 3.915 & 0.7 \\
3 & Claude Sonnet 5.5 & 95.2 & 92.5 & 67.8 & 73.6 & 3.79 & 3.92 & 3.97 & 3.83 & 3.62 & 3.826 & 3.712 & 0.7 \\
4 & DeepSeek-V4.1-Flash & 91.6 & 90.5 & 49.0 & 54.4 & 3.64 & 3.82 & 3.78 & 3.50 & 3.60 & 3.667 & 3.620 & 0.9 \\
5 & Kimi K3 & 91.4 & 79.8 & 40.2 & 43.1 & 3.44 & 3.67 & 3.78 & 3.54 & 3.57 & 3.600 & 3.205 & 0.9 \\
6 & GPT-5.6 Sol & 85.3 & 74.4 & 24.3 & 28.9 & 3.25 & 3.42 & 3.50 & 2.97 & 3.28 & 3.285 & 2.913 & 0.9 \\
7 & Gemini 4 Argon & 96.8 & 74.0 & 50.2 & 54.8 & 3.55 & 3.76 & 3.96 & 3.68 & 3.72 & 3.734 & 2.874 & 1.0 \\
8 & GLM-5.3-Flash & 83.8 & 60.9 & 25.1 & 29.7 & 3.26 & 3.46 & 3.50 & 3.30 & 3.41 & 3.385 & 2.529 & 1.7 \\
9 & Gemini 3.8 Flash & 77.0 & 58.8 & 14.6 & 17.2 & 2.87 & 3.09 & 3.32 & 2.79 & 3.29 & 3.072 & 2.478 & 1.0 \\
\midrule
\multicolumn{14}{@{}l}{\textit{J2: Gemini 3.8 Flash (full video + audio)}}\\
1 & Claude Opus 5.5 & 99.4 & 99.4 & 90.4 & 93.3 & 4.16 & 4.15 & 4.14 & 4.14 & 4.10 & 4.137 & 4.137 & 0.2 \\
2 & GPT-6 Astra & 99.0 & 98.8 & 91.2 & 92.5 & 4.14 & 4.14 & 4.11 & 4.11 & 4.07 & 4.112 & 4.095 & 0.4 \\
3 & DeepSeek-V4.1-Flash & 97.6 & 96.5 & 77.4 & 80.3 & 4.16 & 4.14 & 4.10 & 4.10 & 4.08 & 4.115 & 4.063 & 0.4 \\
4 & Claude Sonnet 5.5 & 96.2 & 93.5 & 82.0 & 84.5 & 4.12 & 4.09 & 4.07 & 4.07 & 4.07 & 4.084 & 3.963 & 0.4 \\
5 & Kimi K3 & 95.1 & 83.1 & 64.4 & 66.1 & 4.05 & 4.01 & 3.98 & 3.96 & 4.00 & 3.999 & 3.562 & 0.4 \\
6 & GPT-5.6 Sol & 96.2 & 84.1 & 58.6 & 60.7 & 4.04 & 4.01 & 3.94 & 3.87 & 3.94 & 3.963 & 3.523 & 0.4 \\
7 & Gemini 4 Argon & 99.3 & 76.1 & 69.9 & 71.1 & 4.12 & 4.08 & 4.15 & 4.10 & 4.07 & 4.107 & 3.160 & 0.3 \\
8 & Gemini 3.8 Flash & 87.9 & 67.3 & 45.2 & 46.0 & 3.75 & 3.70 & 3.70 & 3.46 & 3.80 & 3.681 & 2.968 & 0.6 \\
9 & GLM-5.3-Flash & 90.2 & 65.8 & 47.7 & 49.0 & 4.02 & 3.97 & 3.87 & 3.92 & 3.90 & 3.935 & 2.943 & 0.5 \\
\midrule
\multicolumn{14}{@{}l}{\textit{J3: GPT-5.6 Sol (frames) + Qwen3.8-Omni-Flash (audio)}}\\
1 & GPT-6 Astra & 93.9 & 93.7 & 27.6 & 47.3 & 4.08 & 4.71 & 4.43 & 3.88 & 4.42 & 4.303 & 4.285 & 5.9 \\
2 & Claude Opus 5.5 & 94.2 & 94.2 & 35.1 & 51.9 & 3.88 & 4.41 & 4.36 & 3.79 & 4.42 & 4.172 & 4.172 & 5.4 \\
3 & DeepSeek-V4.1-Flash & 90.1 & 89.0 & 22.6 & 38.1 & 3.79 & 4.43 & 4.23 & 3.65 & 4.40 & 4.101 & 4.049 & 6.8 \\
4 & Claude Sonnet 5.5 & 90.4 & 87.7 & 26.4 & 42.7 & 3.87 & 4.35 & 4.27 & 3.68 & 4.26 & 4.087 & 3.965 & 6.7 \\
5 & Kimi K3 & 86.6 & 75.1 & 20.1 & 29.7 & 3.70 & 4.25 & 4.13 & 3.59 & 4.29 & 3.992 & 3.554 & 5.6 \\
6 & GPT-5.6 Sol & 87.1 & 75.3 & 18.8 & 27.6 & 3.85 & 4.38 & 4.14 & 3.51 & 3.90 & 3.956 & 3.514 & 6.9 \\
7 & Gemini 4 Argon & 93.4 & 70.3 & 22.2 & 39.7 & 3.73 & 4.29 & 4.33 & 3.80 & 4.47 & 4.125 & 3.171 & 6.6 \\
8 & Gemini 3.8 Flash & 76.5 & 57.7 & 8.8 & 16.7 & 3.38 & 3.92 & 3.75 & 3.12 & 4.12 & 3.658 & 2.948 & 5.8 \\
9 & GLM-5.3-Flash & 81.8 & 58.4 & 13.0 & 21.8 & 3.70 & 4.19 & 3.95 & 3.50 & 4.11 & 3.889 & 2.898 & 7.0 \\
\bottomrule
\end{tabular}
}
\end{table}

\subsection{Uncertainty}
\begin{figure}[htbp]
\centering
\includegraphics[width=\linewidth]{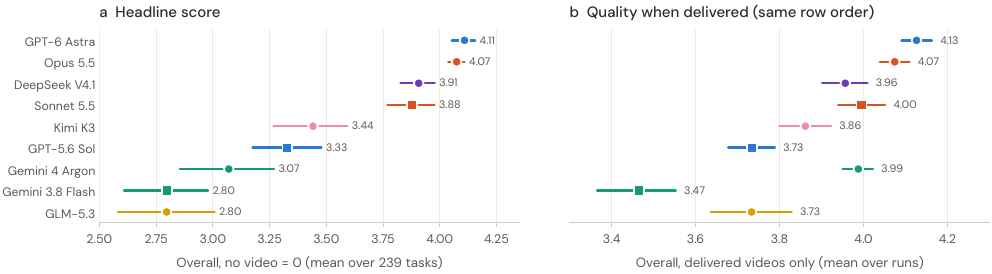}
\vspace{-1.2em}
\caption{\textbf{Per-system means with 95\% bootstrap CIs} (239 tasks, three-panel mean; $B{=}2000$). (a)~Overall with no video $=0$, as in Tab.~\ref{tab:ci}; (b)~overall over delivered videos only, in the same row order. Circles: each lab's flagship; squares: its second model.}
\label{fig:ciforest}
\vspace{-0.6em}
\end{figure}

\begin{table}[htbp]
\centering
\caption{\textbf{Uncertainty.} Left: 95\% bootstrap CI over tasks ($B{=}2000$) of the per-task three-panel overall (no video $=0$); the per-task mean differs from the pooled leaderboard value by $\le$0.011. Right: paired bootstrap probability that rank $k$ beats rank $k{+}1$, with per-task head-to-head wins/losses/ties.}
\label{tab:ci}
\small
\begin{tabular}{@{}lcc@{\hspace{2em}}lcc@{}}
\toprule
System & Mean & 95\% CI & Comparison & $P$ & W/L/T \\
\midrule
GPT-6 Astra & 4.109 & [4.052, 4.156] & Astra $>$ Opus & 0.911 & 127/89/23 \\
Claude Opus 5.5 & 4.074 & [4.037, 4.109] & Opus $>$ DeepSeek & 1.000 & 133/74/32 \\
DeepSeek-V4.1-Flash & 3.907 & [3.826, 3.978] & DeepSeek $>$ Sonnet & 0.697 & 89/121/29 \\
Claude Sonnet 5.5 & 3.878 & [3.769, 3.977] & Sonnet $>$ Kimi & 1.000 & 152/68/19 \\
Kimi K3 & 3.441 & [3.267, 3.593] & Kimi $>$ GPT-5.6 Sol & 0.860 & 143/86/10 \\
GPT-5.6 Sol & 3.328 & [3.174, 3.479] & GPT-5.6 Sol $>$ Argon & 0.980 & 82/142/15 \\
Gemini 4 Argon & 3.070 & [2.854, 3.269] & Argon $>$ Flash & 0.979 & 159/55/25 \\
Gemini 3.8 Flash & 2.799 & [2.609, 2.979] & Flash $>$ GLM & 0.498 & 79/139/21 \\
GLM-5.3-Flash & 2.796 & [2.582, 3.007] & & & \\
\bottomrule
\end{tabular}

\end{table}
\noindent Tab.~\ref{tab:ci} reports bootstrap intervals for the headline score and comparisons between adjacent ranks, showing how delivery and quality interact. Astra leads Opus ($P{=}0.91$; 127 vs.\ 89 head-to-head wins), and both are clearly ahead of DeepSeek ($P{=}1.00$). DeepSeek is ahead of Sonnet on the mean only through delivery (Sonnet wins 121 of the head-to-heads, DeepSeek 89; $P{=}0.70$), and Gemini 4 Argon wins 142 of 239 head-to-heads against GPT-5.6 Sol yet ranks below it, because its 23\% non-delivery contributes zeros. Gemini 3.8 Flash and GLM are statistically tied ($P{=}0.50$). Fig.~\ref{fig:ciforest} compares headline and delivered-only score intervals in the same row order: excluding missing videos moves DeepSeek below Sonnet and Gemini 4 Argon up to fourth.

\subsection{Quality profiles by dimension and category}
\begin{figure}[htbp]
\centering
\includegraphics[width=\linewidth]{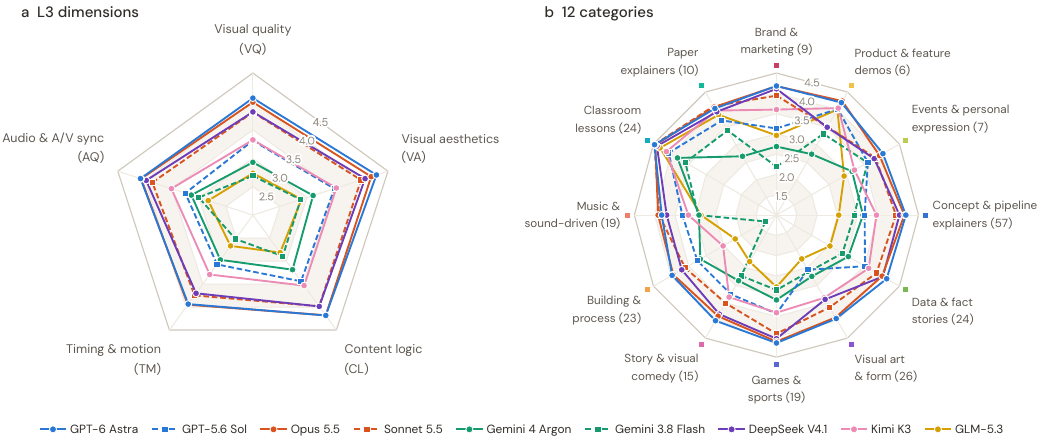}
\vspace{-1.2em}
\caption{\textbf{Quality profile per system} (239 tasks; three-panel mean, no video $=0$). (a)~The five L3 dimensions; (b)~overall score per category (mean of per-task scores; $n$ tasks in parentheses). Radial axes start at 2.0 in (a) and 1.0 in (b), not at 0; rings every 0.5. Circles: each lab's flagship; squares and dashed lines: its second model. Category colors as in Fig.~\ref{fig:stats}.}
\label{fig:radar}
\vspace{-0.6em}
\end{figure}
\noindent Fig.~\ref{fig:radar}(a) shows that the systems keep their tiers on every dimension; only near-tied neighbors swap (\eg, Opus edges out Astra on TM). VA is the highest dimension for most systems and TM the lowest for all but Gemini 4 Argon. By category (Fig.~\ref{fig:radar}b; mean over the nine systems, no video $=0$), \emph{Building \& process} is hardest (3.07) and spreads the systems most (best minus worst 2.66), because it demands long, stateful assembly; \emph{Visual art \& form} is next (3.15). \emph{Classroom lessons} score highest (4.19) and spread least (0.88); every system averages at least 3.59 on them. Astra or Opus is best in 10 of the 12 categories, Sonnet on lessons and papers.
\subsection{Program checks (L1)}
\begin{table}[htbp]
\centering
\caption{\textbf{Program checks} on the 205 core tasks (\% of runs passing; non-delivery fails all checks; rows in the order of Tab.~\ref{tab:full}). $^\dagger$54 tasks require an exact frame count. $^\ddagger$98 tasks whose audio policy \emph{explicitly} marks at least one channel required; pass = audio stream with peak above $-50$\,dBFS. The 134 core tasks of Tab.~\ref{tab:taskstats} also count requirements stated in prose, which only the judges check.}
\label{tab:l1}
\small
\begin{tabular}{@{}l ccccccc rr@{}}
\toprule
System & Playable & Resolution & FPS & Duration & Frame count$^\dagger$ & Audio$^\ddagger$ & Project & \multicolumn{2}{c}{All pass}\\
\midrule
GPT-6 Astra & 99.5 & 98.5 & 96.6 & 98.0 & 98.1 & 100.0 & 96.1 & 190/205 & 92.7 \\
Claude Opus 5.5 & 100.0 & 100.0 & 99.5 & 100.0 & 100.0 & 100.0 & 100.0 & 204/205 & 99.5 \\
Claude Sonnet 5.5 & 96.6 & 96.6 & 96.6 & 94.1 & 92.6 & 98.0 & 95.1 & 192/205 & 93.7 \\
Gemini 4 Argon & 75.6 & 75.6 & 75.6 & 75.6 & 79.6 & 74.5 & 75.6 & 155/205 & 75.6 \\
DeepSeek-V4.1-Flash & 98.5 & 98.5 & 97.6 & 97.1 & 96.3 & 99.0 & 96.6 & 192/205 & 93.7 \\
Kimi K3 & 87.8 & 87.8 & 87.8 & 84.9 & 87.0 & 86.7 & 87.3 & 172/205 & 83.9 \\
GLM-5.3-Flash & 70.7 & 70.7 & 70.2 & 64.9 & 64.8 & 67.3 & 69.3 & 127/205 & 62.0 \\
GPT-5.6 Sol & 87.3 & 85.4 & 84.9 & 86.8 & 88.9 & 88.8 & 86.8 & 169/205 & 82.4 \\
Gemini 3.8 Flash & 78.0 & 78.0 & 78.0 & 76.1 & 85.2 & 80.6 & 77.6 & 156/205 & 76.1 \\
\bottomrule
\end{tabular}

\end{table}
\noindent Among the 1,628 playable deliveries on the core tasks the most frequent program-check failure is duration (34), followed by a missing source project (20), frame rate (15; \eg, 24 instead of 30\,fps, or drift such as 30.01), exact frame count (10), silent or absent audio where audio is required (10) and resolution (6). On the 34 lessons and papers, the rule-based checks (decodable MP4, resolution, frame rate, duration, audible audio, narration coverage by VAD, narration language by Whisper language ID, black or frozen stretches) fail in only 14 of 306 runs, 6 of them Gemini 4 Argon's; 7 of the 14 are the runs without a playable video, the rest delivered videos failing a format or narration check. Program checks thus certify the container reliably.

\section{Agent Behavior}
\label{app:behavior}
This section covers all 2,151 runs (239 tasks $\times$ 9 systems).

\subsection{Outcomes and effort}
\begin{table}[htbp]
\centering
\caption{\textbf{Run outcomes and effort} (239 tasks; rows in the headline order, no video $=0$, here and in the other tables of this section). Video: an MP4 was collected; Playable: the judges could decode and score it. Median s: elapsed budget time. Calls, tool calls and output tokens: per-run medians. Reas.: reasoning share of output tokens.}
\label{tab:outcomes}
\resizebox{\linewidth}{!}{\begin{tabular}{@{}l rrr rr r rrr r@{}}
\toprule
System & Self-ended & Timed out & Error & Video & Playable & Median s & Calls & Tool calls & Out.\ tok (k) & Reas.\\
\midrule
GPT-6 Astra & 229 & 10 & 0 & 238 & 238 & 2{,}464 & 51 & 49 & 60.9 & 49\% \\
Claude Opus 5.5 & 238 & 1 & 0 & 239 & 239 & 2{,}283 & 51 & 49 & 130.7 & 56\% \\
DeepSeek-V4.1-Flash & 217 & 22 & 0 & 239 & 236 & 3{,}070 & 117 & 130 & 166.8 & 59\% \\
Claude Sonnet 5.5 & 218 & 20 & 1 & 235 & 232 & 3{,}065 & 56 & 63 & 254.7 & 75\% \\
Kimi K3 & 180 & 58 & 1 & 218 & 213 & 2{,}874 & 63 & 80 & 75.6 & 51\% \\
GPT-5.6 Sol & 147 & 90 & 2 & 224 & 213 & 3{,}370 & 109 & 101 & 129.7 & 57\% \\
Gemini 4 Argon & 177 & 60 & 2 & 209 & 184 & 2{,}994 & 46 & 68 & 125.5 & 28\% \\
Gemini 3.8 Flash & 184 & 55 & 0 & 224 & 193 & 2{,}935 & 122 & 118 & 106.9 & 41\% \\
GLM-5.3-Flash & 160 & 79 & 0 & 193 & 179 & 3{,}234 & 64 & 83 & 103.3 & 56\% \\
\bottomrule
\end{tabular}

}
\end{table}
\begin{figure}[htbp]
\centering
\includegraphics[width=\linewidth]{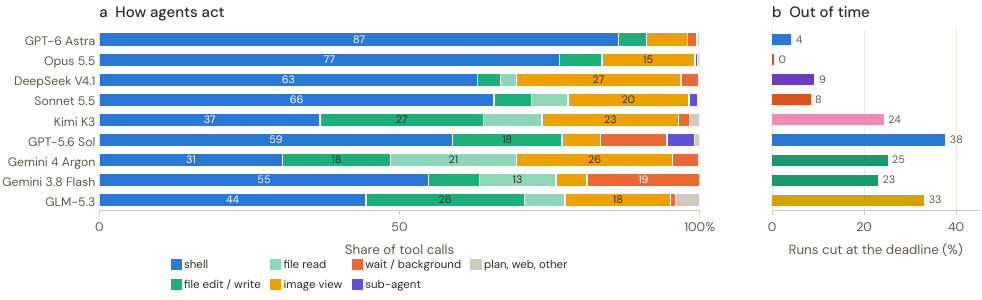}
\vspace{-1.2em}
\caption{\textbf{How agents act} (2,151 runs). (a)~Share of each system's tool calls by kind (numbers: shares of at least 12\%); (b)~share of runs cut at the one-hour deadline.}
\label{fig:toolmix}
\vspace{-0.6em}
\end{figure}
\noindent\textbf{Strong systems finish with time to spare.} Opus and Astra usually end well before the hour (Fig.~\ref{fig:toolmix}b), while GPT-5.6 Sol and GLM often run into the deadline; Sol has usually saved a playable video by then, GLM often has not. Within every system longer runs score lower. The harnesses shape how agents work (Fig.~\ref{fig:toolmix}a): Claude Code and Codex agents act almost entirely through the shell, Gemini, Kimi and GLM rely more on dedicated file tools, and Gemini 3.8 Flash spends much of its time waiting on background tasks. Sub-agents and web search are rarely used; installing packages is common.

\subsection{Rendering stacks}
\begin{table}[htbp]
\centering
\caption{\textbf{Primary renderer} per system (\% of runs) and, below, outcomes pooled by primary renderer. Other: Manim, MoviePy or unidentified. Synth.\ audio: runs that synthesize music or sound effects in code.}
\label{tab:stack}
\resizebox{\linewidth}{!}{\begin{tabular}{@{}l cccccc cc@{}}
\toprule
 & \multicolumn{6}{c}{Primary renderer (\% of runs)} & \multicolumn{2}{c}{Also used (\% of runs)}\\
\cmidrule(lr){2-7}\cmidrule(lr){8-9}
System & cairo & PIL/numpy & Browser 2D & three.js & Blender & Other & ffmpeg & Synth.\ audio\\
\midrule
GPT-6 Astra & 25 & 16 & 23 & 29 & 7 & 0 & 100 & 93 \\
Claude Opus 5.5 & 54 & 5 & 15 & 20 & 4 & 2 & 100 & 96 \\
DeepSeek-V4.1-Flash & 24 & 12 & 36 & 24 & 4 & 0 & 100 & 95 \\
Claude Sonnet 5.5 & 46 & 10 & 22 & 19 & 2 & 1 & 100 & 89 \\
Kimi K3 & 3 & 45 & 26 & 23 & 3 & 0 & 97 & 85 \\
GPT-5.6 Sol & 4 & 55 & 13 & 13 & 14 & 2 & 99 & 57 \\
Gemini 4 Argon & 42 & 20 & 17 & 18 & 3 & 1 & 97 & 93 \\
Gemini 3.8 Flash & 31 & 13 & 21 & 32 & 2 & 1 & 100 & 84 \\
GLM-5.3-Flash & 11 & 43 & 21 & 21 & 3 & 1 & 96 & 86 \\
\midrule
\multicolumn{9}{@{}l}{\textit{Pooled over the nine systems, by primary renderer}}\\
Runs & 576 & 524 & 463 & 473 & 95 & 20 &  &  \\
Playable (\%) & 94 & 90 & 94 & 82 & 83 & 70 &  &  \\
Timed out (\%) & 9 & 22 & 14 & 28 & 31 & 15 &  &  \\
Overall, no video $=$ 0 & 3.79 & 3.49 & 3.73 & 3.00 & 3.12 & 2.88 &  &  \\
Overall, delivered & 4.02 & 3.88 & 3.98 & 3.67 & 3.75 & 4.11 &  &  \\
\bottomrule
\end{tabular}

}
\end{table}
\noindent The pipeline is always ``code $\rightarrow$ frames $\rightarrow$ ffmpeg'', and each system has a characteristic rendering style (Tab.~\ref{tab:stack}): the Claude agents and Gemini 4 Argon draw vectors with cairo, GPT-5.6 Sol, Kimi and GLM rasterize frames with PIL/numpy, and Astra, DeepSeek and Gemini 3.8 Flash capture browser canvases and three.js scenes. Off-the-shelf video frameworks such as Manim or MoviePy are almost never chosen. Vector and 2D-canvas pipelines deliver most often and score highest; 3D scenes, rendered with software WebGL, are the slowest and score lowest. Since renderer and system are coupled, this is an association.

\subsection{Self-inspection}
\begin{table}[htbp]
\centering
\caption{\textbf{Self-inspection.} An image view is a tool call that returns an image to the model. Extract/run: shell commands pulling stills or contact sheets from a rendered video. Last 25\%: share of views in the final quarter of the budget. Right: mean delivered score of playable runs by number of image views (``--'': no runs; parentheses: fewer than 5 runs).}
\label{tab:selfverif}
\resizebox{\linewidth}{!}{\begin{tabular}{@{}l cccccccc ccccc@{}}
\toprule
 & \multicolumn{8}{c}{Self-inspection statistics} & \multicolumn{5}{c}{Mean delivered score of playable runs by \# images viewed}\\
\cmidrule(lr){2-9}\cmidrule(lr){10-14}
System & $\ge$1 view & Median & Mean & p90 & Extract/run & ffprobe/run & Runs ffprobe & Last 25\% & 0 & 1--4 & 5--14 & 15--39 & 40+\\
\midrule
GPT-6 Astra & 229/239 & 3 & 3.4 & 6 & 3.1 & 2.1 & 202 & 22\% & 4.27 & 4.13 & 4.08 & -- & -- \\
Claude Opus 5.5 & 238/239 & 7 & 7.9 & 12 & 4.6 & 3.9 & 237 & 11\% & (3.47) & 3.97 & 4.09 & 4.24 & (3.80) \\
DeepSeek-V4.1-Flash & 239/239 & 35 & 36.0 & 56 & 6.9 & 8.0 & 238 & 10\% & -- & (4.27) & 3.53 & 3.91 & 4.07 \\
Claude Sonnet 5.5 & 229/239 & 10 & 14.3 & 28 & 4.7 & 4.7 & 232 & 24\% & 3.27 & 3.77 & 3.94 & 4.20 & 4.24 \\
Kimi K3 & 239/239 & 17 & 18.8 & 31 & 1.9 & 2.7 & 214 & 38\% & -- & 3.29 & 3.84 & 3.90 & 4.05 \\
GPT-5.6 Sol & 231/239 & 6 & 6.8 & 13 & 2.8 & 4.6 & 222 & 20\% & 3.63 & 3.77 & 3.72 & 3.76 & -- \\
Gemini 4 Argon & 235/239 & 18 & 18.3 & 29 & 0.3 & 1.3 & 182 & 27\% & -- & (4.04) & 3.97 & 4.00 & (3.85) \\
Gemini 3.8 Flash & 182/239 & 6 & 6.5 & 14 & 0.4 & 3.4 & 218 & 19\% & 3.23 & 3.42 & 3.57 & 3.57 & -- \\
GLM-5.3-Flash & 230/239 & 14 & 14.4 & 26 & 1.5 & 2.4 & 196 & 15\% & (2.17) & 3.07 & 3.69 & 3.89 & -- \\
\bottomrule
\end{tabular}

}
\end{table}

\begin{figure}[htbp]
\centering
\includegraphics[width=0.43\linewidth]{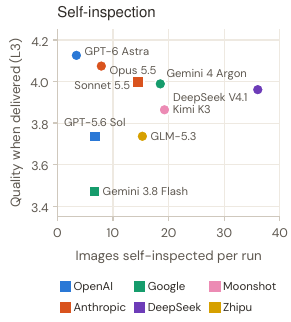}
\caption{\textbf{Self-inspection.} Images self-inspected per run vs.\ delivered quality (nine systems).}
\label{fig:imageview}
\end{figure}
\noindent Most runs (95\%) view their own rendered frames, yet the amount of looking does not predict delivered quality ($\rho{=}{-}0.07$, $n{=}9$; Fig.~\ref{fig:imageview}): Astra looks least (median 3 images per run) and scores best, while DeepSeek looks most (median 35). Reports often claim verifications that do not hold: among playable videos that fail spec compliance, the final report still claims verification (``verified'', ``ffprobe'', ``confirmed'') in 10 of 14 for Astra and 3 of 11 for Sol.

\noindent Within systems, looking more goes with better videos for some (Sonnet, Kimi, GLM, Gemini 3.8 Flash) and not for others.

\subsection{Failure taxonomy}
\label{app:failures}
\begin{table}[htbp]
\centering
\caption{\textbf{Why videos are not delivered.} Top: main cause of each of the 224 runs without a playable video, assigned by reading each run's records. Bottom: state at the end of these runs.}
\label{tab:failures}
\resizebox{\linewidth}{!}{\begin{tabular}{@{}l ccccccccc r@{}}
\toprule
Cause or failure mode (runs) & GPT-6 Astra & Opus 5.5 & DeepSeek & Sonnet 5.5 & Kimi K3 & GPT-5.6 Sol & Gem.~4 Argon & Gem.~3.8 Flash & GLM-5.3-Flash & Total\\
\midrule
\textit{Time management} & 1 & 0 & 2 & 7 & 25 & 25 & 31 & 22 & 56 & 169 \\
\quad Final render started late or ran too slowly & 1 & 0 & 2 & 5 & 14 & 18 & 20 & 19 & 25 & 104 \\
\quad Final render never started & 0 & 0 & 0 & 2 & 11 & 7 & 11 & 3 & 31 & 65 \\
\textit{Stalled or cut-off render} & 0 & 0 & 0 & 0 & 0 & 0 & 21 & 24 & 0 & 45 \\
\quad ffmpeg pipe deadlock (stderr never drained) & 0 & 0 & 0 & 0 & 0 & 0 & 19 & 14 & 0 & 33 \\
\quad Turn ended, background render cut off & 0 & 0 & 0 & 0 & 0 & 0 & 2 & 10 & 0 & 12 \\
\textit{Own output destroyed or lost} & 0 & 0 & 1 & 0 & 1 & 1 & 1 & 0 & 3 & 7 \\
\textit{Other} & 0 & 0 & 0 & 0 & 0 & 0 & 2 & 0 & 1 & 3 \\
\quad Provider content filter & 0 & 0 & 0 & 0 & 0 & 0 & 0 & 0 & 1 & 1 \\
\quad Request over the 20\,MB size limit & 0 & 0 & 0 & 0 & 0 & 0 & 2 & 0 & 0 & 2 \\
\midrule
\textbf{No playable video} & 1 & 0 & 3 & 7 & 26 & 26 & 55 & 46 & 60 & 224 \\
\midrule
\multicolumn{11}{@{}l}{\textit{State at the end of these runs}}\\
\quad Truncated MP4 (no \texttt{moov} atom) & 0 & 0 & 3 & 3 & 5 & 11 & 25 & 30 & 13 & 90 \\
\quad MP4 without a video stream & 0 & 0 & 0 & 0 & 0 & 0 & 0 & 1 & 1 & 2 \\
\quad No MP4, timed out & 1 & 0 & 0 & 3 & 21 & 15 & 26 & 9 & 38 & 113 \\
\quad No MP4, agent ended its turn & 0 & 0 & 0 & 1 & 0 & 0 & 2 & 6 & 8 & 17 \\
\quad No MP4, run ended on an error & 0 & 0 & 0 & 0 & 0 & 0 & 2 & 0 & 0 & 2 \\
\bottomrule
\end{tabular}

}
\end{table}
\noindent Tab.~\ref{tab:failures} extends \S\ref{sec:failures}. Writing renders to a temporary file would not have saved most truncated MP4s, since they mostly come from stalled or unfinished renders rather than from overwriting a finished video.

\section{Cost}
\label{app:cost}
\noindent The mean list-price API cost per task ranges from \$0.32 (GLM-5.3-Flash) to \$15.38 (Claude Opus 5.5), a $49\times$ spread. Across the nine systems its rank correlation with the headline score is $\rho{=}0.55$ (permutation $p{=}0.13$; 0.43 with the delivered-only score). The cost--score Pareto frontier is GLM $\rightarrow$ DeepSeek $\rightarrow$ Astra. Mean output tokens per task range from 61k (Astra) to 289k (Sonnet); Tab.~\ref{tab:outcomes} reports the lower per-run medians.
\begin{table}[htbp]
\centering
\caption{\textbf{API cost per task} (USD, list-price equivalents of the logged usage of every run; 239 tasks, 163,197 API calls; rows in the headline order). Calls/task: requests with usage records per task. Overall: headline three-panel score (no video $=0$).}
\label{tab:cost}
\small
\resizebox{\linewidth}{!}{\begin{tabular}{@{}l rrrr r r r r r r@{}}
\toprule
System & Mean & Median & p90 & Max & Total (239) & Calls/task & Cache hit (\%) & \$/playable video & Overall & Overall per \$\\
\midrule
GPT-6 Astra & 8.19 & 7.88 & 11.88 & 15.61 & 1{,}957 & 46 & 97 & 8.22 & 4.109 & 0.50 \\
Claude Opus 5.5 & 15.38 & 10.19 & 31.22 & 47.08 & 3{,}676 & 51 & 65 & 15.38 & 4.074 & 0.26 \\
DeepSeek-V4.1-Flash & 0.47 & 0.45 & 0.67 & 1.30 & 113 & 117 & 97 & 0.48 & 3.911 & 8.25 \\
Claude Sonnet 5.5 & 8.26 & 7.09 & 13.07 & 34.99 & 1{,}974 & 63 & 81 & 8.51 & 3.880 & 0.47 \\
Kimi K3 & 3.53 & 3.53 & 4.94 & 7.25 & 843 & 61 & 97 & 3.96 & 3.440 & 0.98 \\
GPT-5.6 Sol & 11.84 & 11.49 & 19.59 & 28.39 & 2{,}831 & 106 & 81 & 13.29 & 3.317 & 0.28 \\
Gemini 4 Argon & 2.94 & 2.84 & 4.32 & 7.13 & 703 & 48 & 89 & 3.82 & 3.069 & 1.04 \\
Gemini 3.8 Flash & 2.44 & 2.09 & 4.16 & 6.66 & 584 & 126 & 91 & 3.03 & 2.798 & 1.14 \\
GLM-5.3-Flash & 0.32 & 0.30 & 0.45 & 0.68 & 76 & 65 & 93 & 0.42 & 2.790 & 8.83 \\
\bottomrule
\end{tabular}

}
\end{table}
\noindent The total is about \$12.8k. 76\% of Opus's cost is cache writes (2.3M tokens written per task at a 65\% cache-hit rate), and 55\% of Sol's is uncached input. Some prices are promotional: at the announced regular prices Gemini 4 Argon would cost about \$5.9 and Gemini 3.8 Flash about \$4.9 per task.

\section{Human Preferences and Visual L2 Review}
\label{app:human}

\subsection{Pairwise video preferences}

\runin{busts-in-silhouette}{Protocol.} Annotators compare two playable videos from different systems on the same task, both evaluated by all three panels. The study thus measures delivered-video quality, not delivery. System pairs are balanced across tasks, and presentation sides are randomized; annotators are blind to system identity and judge scores. The interface shows the full prompt and two synchronized players with a choice of audio track, and offers \emph{A better}, \emph{B better}, \emph{tie} and \emph{both bad}; the instruction is ``which better matches the prompt and is better overall''. We fit Bradley--Terry strengths on decisive votes (MM algorithm with a weak prior; Elo scale with geometric mean 1000), with 2,000-sample bootstrap CIs over decisive pairs and Wilson intervals for win rates.

\begin{table}[htbp]
\centering
\caption{\textbf{Human Bradley--Terry ratings} (decisive votes; Elo scale; 95\% bootstrap CI and rank CI). Right: the system's delivered-only overall and its rank (Tab.~\ref{tab:full}).}
\label{tab:bt}
\small
\begin{tabular}{@{}lcccc@{}}
\toprule
System & Win rate [Wilson 95\%] & BT-Elo [95\% CI] & Rank CI & Overall, delivered (rank) \\
\midrule
GPT-6 Astra & 0.783 [0.644, 0.877] & 1242 [1134, 1387] & 1--3 & 4.126 (1) \\
Claude Opus 5.5 & 0.741 [0.616, 0.837] & 1203 [1111, 1327] & 1--3 & 4.074 (2) \\
Claude Sonnet 5.5 & 0.676 [0.508, 0.809] & 1121 [1016, 1257] & 1--5 & 3.999 (3) \\
Gemini 4 Argon & 0.607 [0.476, 0.724] & 1081 [1001, 1176] & 2--6 & 3.989 (4) \\
DeepSeek-V4.1-Flash & 0.511 [0.370, 0.650] & 1022 [900, 1154] & 3--7 & 3.961 (5) \\
Kimi K3 & 0.449 [0.319, 0.587] & 997 [873, 1107] & 4--7 & 3.864 (6) \\
GLM-5.3-Flash & 0.375 [0.252, 0.516] & 905 [792, 1012] & 5--7 & 3.736 (7) \\
Gemini 3.8 Flash & 0.170 [0.089, 0.301] & 723 [548, 833] & 8--9 & 3.470 (9) \\
GPT-5.6 Sol & 0.156 [0.077, 0.288] & 705 [516, 814] & 8--9 & 3.735 (8) \\
\bottomrule
\end{tabular}
\end{table}

\runin{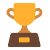}{Three tiers.} Astra and Opus lead and cannot be separated (+39 Elo, $P{=}0.69$); both are significantly above DeepSeek, Kimi and GLM, and Astra also above Argon. Sonnet, Argon, DeepSeek, Kimi and GLM form a middle group whose neighbors cannot be separated. Gemini 3.8 Flash and GPT-5.6 Sol share the last place (+18 Elo, $P{=}0.58$) and are both significantly below the other seven.

\begin{figure}[htbp]
\centering
\includegraphics[width=\linewidth]{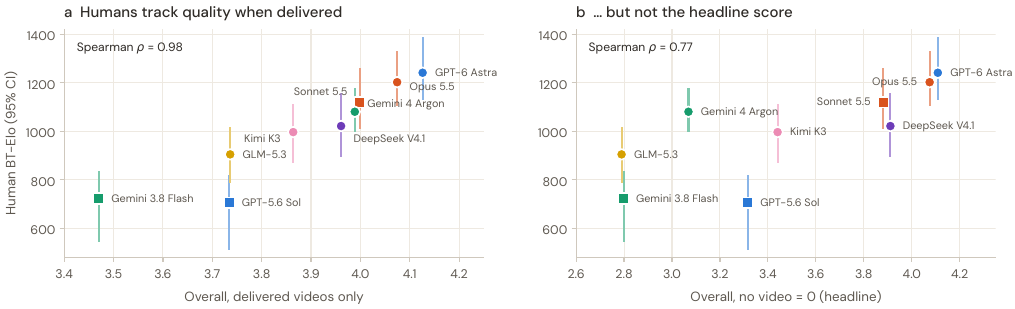}
\vspace{-1.2em}
\caption{\textbf{Human BT-Elo against the three-panel overall} (nine systems; vertical bars: 95\% bootstrap CI of the Elo, Tab.~\ref{tab:bt}). (a)~Overall over delivered videos only; (b)~headline overall with no video $=0$. Circles: each lab's flagship; squares: its second model.}
\label{fig:humanjudge}
\vspace{-0.6em}
\end{figure}

\runin{check-mark-button}{Humans validate quality, not delivery.} The human ranking is almost the delivered-only judge ranking (Spearman $\rho{=}0.98$, Kendall $\tau{=}0.94$; only Flash and Sol swap the last two places), and it agrees less with metrics that include delivery (Fig.~\ref{fig:humanjudge}, Tab.~\ref{tab:humanmetric}). With no video $=0$, Argon falls from 4th to 7th and Sol rises from 9th to 6th, because humans never see a missing video. The study thus validates the score's quality component; delivery rate is reported separately.

\begin{table}[htbp]
\centering
\caption{\textbf{Human Elo vs.\ leaderboard metrics} (system level, $n{=}9$): rank correlation with each three-panel metric.}
\label{tab:humanmetric}
\small
\begin{tabular}{@{}lcc@{}}
\toprule
Metric & Spearman & Kendall \\
\midrule
Overall, delivered only & \textbf{0.98} & \textbf{0.94} \\
L2 pass rate (delivered) & 0.92 & 0.78 \\
L2 strict success & 0.92 & 0.78 \\
Overall, no video $=0$ & 0.77 & 0.61 \\
Playable-delivery rate & 0.66 & 0.47 \\
\bottomrule
\end{tabular}
\end{table}

\runin{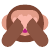}{Judges do not flag ``both bad''.} On pairs that humans labeled \emph{both bad}, the panels' mean score (3.81) equals that on decisive pairs (3.83), and only 4\% of such pairs have both videos below 3.5. Judge scores order videos but do not certify them as acceptable.

\subsection{Visual L2 review}
\label{app:visualreview}
We conduct a targeted review of visual L2 requirements marked fail by at least two judge panels, covering nine systems, 284 videos and 115 tasks. Requirements are selected within generator systems. Each human reviewer receives the task requirements, the check criterion and timestamped frames from the collected video; generator identity, prior verdicts and scores are withheld. Audio and audio-visual checks are excluded because the review evidence contains no audible audio. Checks requiring continuous motion or exact timing intervals that cannot be verified from static frames alone are marked unknown. Each item is reviewed in a single blind pass; incomplete or unusable form responses remain missing.

\begin{table}[htbp]
\centering
\caption{\textbf{Visual L2 review outcomes.} Shares use inverse inclusion weights across generator systems and include unresolved and missing outcomes. Fail supports the original failure judgment; pass indicates that the requirement is satisfied. Rounding may prevent percentages from summing to 100.}
\label{tab:visualreview}
\small
\begin{tabular}{@{}lr@{}}
\toprule
Review outcome & Weighted share (\%) \\
\midrule
Fail: original judgment supported & 60.8 \\
Pass: requirement satisfied & 8.7 \\
Unknown: insufficient static-frame evidence & 19.5 \\
Missing: incomplete response & 11.1 \\
\bottomrule
\end{tabular}
\end{table}

\noindent\textbf{Confirmation of flagged errors.} Among resolved cases (pass or fail), 86.5\% support the original failure judgment; inverse-inclusion weighting gives 87.5\%. Table~\ref{tab:visualreview} separately reports unknown and missing outcomes.

\section{Ablations}
\label{app:ablations}
Four ablations, run once per cell after the main evaluation, test score reliability and whether the setting limited the results:
\begin{itemize}[leftmargin=1.9em,labelsep=0.5em,itemsep=1pt,topsep=2pt]
\item[\ico{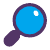}] Can the judges see defects? (\S\ref{app:x11})
\item[\ico{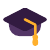}] Do the scores reflect whether a video actually teaches? (\S\ref{app:x14})
\item[\ico{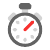}] Was one hour enough? (\S\ref{app:x13})
\item[\ico{desktop-computer}] Did the agents need more CPU cores? (\S\ref{app:cpu})
\end{itemize}
We additionally study feedback during production and revision after delivery (\S\ref{app:feedbackstudies}).

\subsection{Judge stress test}
\label{app:x11}
We took 30 delivered videos, re-encoded each one cleanly and built ten defective versions with ffmpeg: five picture defects in the window from 35\% to 65\% of the runtime (black frames, a grey box over the center, heavy blur, a frozen picture) or over the whole video (left--right mirror), two sound defects (a silent window, another task's soundtrack), audio delayed by 0.5 or 1.5\,s, and the two halves swapped. Apart from the black window, every defective version passes all program checks, so only a judge can notice it. J1 judged each video in its clean version and in five of the defects (15 videos per defect, 180 judgments) with the unchanged prompt and rubric, and was not told that anything was injected.

\begin{figure}[htbp]
\centering
\includegraphics[width=0.56\linewidth]{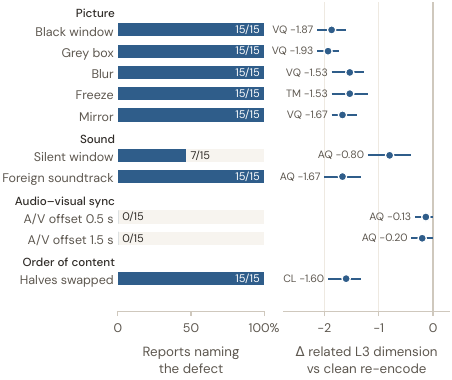}
\vspace{-0.4em}
\caption{\textbf{Judge stress test} (J1; 15 videos per defect). Left: share of reports that name the defect (checked by hand). Right: change of the most related L3 dimension against the clean re-encode of the same video (95\% bootstrap CI).}
\label{fig:abl_judge}
\vspace{-0.6em}
\end{figure}
\noindent\textbf{Judges see broken pictures but not late sound} (Fig.~\ref{fig:abl_judge}). Every picture defect, the foreign soundtrack and the swapped halves are named in all 15 reports and cost 1.5--1.9 points on the related dimension; a silent window is named in 7 of 15 ($-$0.80 on AQ). Neither audio delay is mentioned in any report, and reports often praise the synchronization; the five-dimension mean drops only slightly ($-$0.11 and $-$0.16; Wilcoxon $p{=}0.07$ and 0.05), spread over several dimensions rather than AQ. Detection does not always reach the checks: mirrored and reordered videos are named in every report, yet most of their target checks still pass (flip rate $+$4.5 and $+$14.9 points over clean), and the penalty lands on VQ and CL instead. Finally, the clean re-encode scores $+$0.12 [$+$0.03, $+$0.22] above the original judgment of the same video, so gaps of 0.1--0.2 between single videos are within judge noise. Audio-visual timing is thus a blind spot of both the judges and the program checks, so our rubrics set no strict offset limit: a synchronization check asks only that the right sound accompanies the right event (App.~\ref{app:rubric}), not that audio and picture align within a fixed tolerance.

\subsection{Do the videos teach? A grounded quiz}
\label{app:x14}
Each of the 34 lessons and papers has a multiple-choice quiz written from its source material (404 questions). A standard multiple-choice quiz is insufficient: a strong model can answer many questions from prior knowledge without viewing the video. Each answer must therefore cite a supporting moment in the video. A grader (Gemini 4 Argon) watches only the finished video and answers A--D or E (``the video does not cover it''). Every A--D answer must give a timestamp and the video's own words; narration quotes are matched against a Whisper transcript and on-screen quotes against the frames, and an answer whose evidence fails verification counts as E. A task's score is its share of correct answers; a run without a video scores 0. The quiz does not enter the leaderboard.

\begin{figure}[htbp]
\centering
\includegraphics[width=0.56\linewidth]{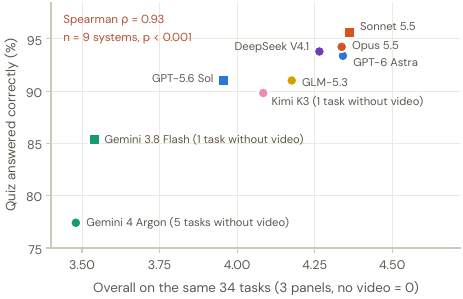}
\vspace{-0.4em}
\caption{\textbf{Grounded quiz} on the 34 lessons and papers: share of questions answered correctly (mean over tasks) vs.\ the three-panel overall on the same tasks (no video $=0$).}
\label{fig:abl_quiz}
\vspace{-0.6em}
\end{figure}
\noindent\textbf{The quiz agrees with the judges.} Although it uses neither the rubric nor the quality anchors, its system ranking correlates with the three-panel overall at Spearman $\rho{=}0.93$ (Fig.~\ref{fig:abl_quiz}; also 0.93 on playable videos only). The top four systems convey 93--96\% of the tested points; Gemini 4 Argon's 77.4\% comes from its five tasks without a playable video (90.8\% on the other 29). Almost no point is lost to a wrong answer (2 of 3,425 A--D answers); the losses are questions the video does not cover or whose evidence fails verification, so the quiz measures coverage of the material more than correctness.

\subsection{Time budget}
\label{app:x13}
We reran DeepSeek-V4.1-Flash, Gemini 3.8 Flash and GLM-5.3-Flash on 20 tasks with budgets of 15, 30, 60 and 120 minutes (one run per cell, otherwise as in the main evaluation). Runs are judged by J2 with no video $=0$ and compared with the same system's 60-minute arm by a paired Wilcoxon test.

\begin{figure}[htbp]
\centering
\includegraphics[width=0.6\linewidth]{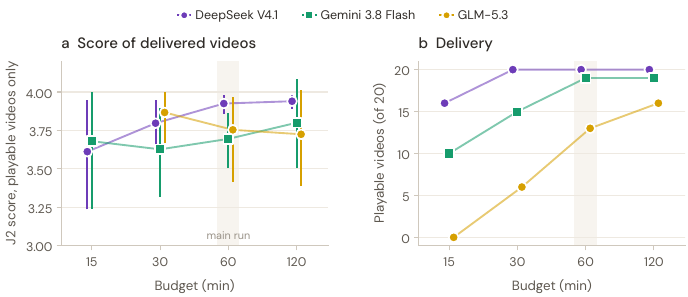}
\vspace{-0.4em}
\caption{\textbf{Time budget} (3 systems, 20 tasks each, J2). (a)~Score over playable videos only, 95\% CI (GLM has none at 15 minutes); (b)~playable videos of 20. Shaded: the main run's 60 minutes.}
\label{fig:abl_time}
\vspace{-0.6em}
\end{figure}

\begin{table}[htbp]
\centering
\footnotesize
\caption{\textbf{Time budget.} J2 score with no video $=0$ (playable videos of 20 in parentheses); $\Delta$ and paired Wilcoxon $p$ against the same system at 60 minutes.}
\label{tab:abl_time}
\setlength{\tabcolsep}{3.5pt}
\begin{tabular}{ll cc}
\toprule
System & Budget & Score & $\Delta$ ($p$) \\
\midrule
\multirow{4}{*}{DeepSeek-V4.1-Flash} & 15 min  & 2.89 (16) & $-$1.03 (0.006) \\
                                     & 30 min  & 3.80 (20) & $-$0.13 (0.11) \\
                                     & 60 min  & 3.92 (20) & -- \\
                                     & 120 min & 3.94 (20) & $+$0.01 (0.78) \\
\midrule
\multirow{4}{*}{Gemini 3.8 Flash}    & 15 min  & 1.84 (10) & $-$1.67 (0.001) \\
                                     & 30 min  & 2.72 (15) & $-$0.79 (0.13) \\
                                     & 60 min  & 3.51 (19) & -- \\
                                     & 120 min & 3.61 (19) & $+$0.10 (0.32) \\
\midrule
\multirow{4}{*}{GLM-5.3-Flash}       & 15 min  & 0.00 (0)  & $-$2.44 ($<$0.001) \\
                                     & 30 min  & 1.16 (6)  & $-$1.28 (0.017) \\
                                     & 60 min  & 2.44 (13) & -- \\
                                     & 120 min & 2.98 (16) & $+$0.54 (0.25) \\
\bottomrule
\end{tabular}
\end{table}

\noindent\textbf{The budget decides whether a video is delivered, not how good it is} (Fig.~\ref{fig:abl_time}, Tab.~\ref{tab:abl_time}). From 15 to 120 minutes, playable videos rise from 16, 10 and 0 of 20 (DeepSeek, Flash, GLM) to 20, 19 and 16. Except for DeepSeek at 15 minutes (3.61 against 3.92 at 60), the mean over playable videos moves by at most 0.17 within a system (DeepSeek 3.80--3.94, Flash 3.63--3.80, GLM 3.73--3.87), with overlapping CIs.

\noindent\textbf{How much time is enough differs between systems.} DeepSeek and Flash gain little once the budget reaches that of the main run, while GLM delivers more videos at every larger budget (0, 6, 13 and 16 of 20; Tab.~\ref{tab:abl_time}). Flash tends to end its runs on its own well before the deadline, whereas DeepSeek and GLM use the extra time they are given; some GLM runs are still cut at the deadline even with the longest budget, and a single GLM call can spend a long time reasoning before issuing any tool call.

\subsection{CPU budget}
\label{app:cpu}
We reran six systems (Gemini 4 Argon, Gemini 3.8 Flash, DeepSeek-V4.1-Flash, GLM-5.3-Flash, Kimi K3, GPT-5.6 Sol) on the same 20 render-heavy tasks with 2, 8 and 16 CPU cores; the main evaluation uses 8. Everything else is as in the main run; scores are the three-panel mean with no video $=0$ (360 runs), compared over the 120 task--system pairs.

\begin{figure}[htbp]
\centering
\includegraphics[width=0.56\linewidth]{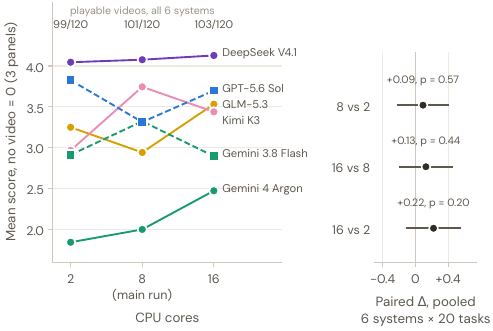}
\vspace{-0.4em}
\caption{\textbf{CPU budget} (6 systems $\times$ 20 tasks; three-panel mean, no video $=0$). Left: mean score per system and setting, with playable videos of all six systems on top. Right: pooled paired difference with 95\% CI and paired-$t$ $p$.}
\label{fig:abl_cpu}
\vspace{-0.6em}
\end{figure}
\noindent\textbf{More cores do not measurably help} (Fig.~\ref{fig:abl_cpu}). None of the pooled differences is significant ($+$0.09 for 8 vs.\ 2 cores, $+$0.13 for 16 vs.\ 8, $+$0.22 for 16 vs.\ 2; $p{=}0.57$, 0.44, 0.20; a bootstrap over tasks gives the same conclusion), and playable videos go from 99 to 101 and 103 of 120 (sign tests $p{\ge}0.56$). The systems disagree in direction (Kimi and Flash do best with eight cores, Sol with two, DeepSeek is flat), and delivered quality hardly moves (3.81, 3.84, 3.92). Argon's rise from 9 to 12 videos comes mostly from fewer ffmpeg pipe deadlocks in the 16-core arm (2 of its 8 failures there, against 9 of 21 in the other two), which barely use the CPU; render-heavy tasks still run out of time with sixteen cores. The CPU budget therefore does not measurably limit the main results.

\subsection{Feedback studies}
\label{app:feedbackstudies}
We examine feedback at two stages: optional Gemini feedback during production and revision of an existing video after delivery.

\subsubsection{Gemini feedback during video production}
\label{app:feedbackstudy}
\noindent\textbf{Design.} We compare DeepSeek-V4.1-Flash with native image viewing against the same setup augmented with an optional Gemini 3.8 Flash feedback tool. Both conditions retain native image viewing and use a one-hour budget, including tool waits. The study covers 12 tasks with previously observed visual core-check failures and native image-viewing activity. Each task has two independent executions per condition, with condition order randomized within each task--repeat pair, giving 48 assigned runs. Tool availability defines the condition: runs that do not call the tool remain in the feedback group. Gemini 4 Argon evaluates the full videos with audio using the original rubrics, blinded to condition and feedback.

\noindent\textbf{Measures.} The primary measure is the fraction of fixed core checks explicitly marked pass; unknown and n/a do not count as pass. This differs from the leaderboard's strict-success criterion. Runs without an on-time playable delivery score zero; delivered videos without complete judgments remain missing. We average the two executions within each task and condition before computing paired differences, with 95\% intervals from 10,000 task-bootstrap resamples.

\begin{table}[htbp]
\centering
\caption{\textbf{Optional Gemini feedback.} Counts are runs. Complete judgments refer to delivered videos; missing judgments are retained as missing. All assigned runs remain in their original condition.}
\label{tab:feedbackstudy}
\small
\begin{tabular}{@{}lrr@{}}
\toprule
 & Native tools & Native tools + Gemini \\
\midrule
Assigned runs & 24 & 24 \\
On-time playable deliveries & 21 & 24 \\
Deliveries with complete judgments & 16 & 15 \\
Deliveries missing complete judgments & 5 & 9 \\
Runs invoking feedback & -- & 18 \\
Runs receiving valid feedback & -- & 17 \\
\bottomrule
\end{tabular}
\end{table}

\noindent\textbf{Results.} The feedback group has a 12.5-percentage-point higher delivery rate (95\% task-bootstrap CI: 0.0--25.0 points). Only three tasks have complete judgments for all four executions, so complete-case quality comparisons do not represent the full task set. Across all assigned runs, the confirmed-core-pass fraction lies within 57.4--78.3\% for native tools and 55.6--93.1\% with feedback; the paired difference lies within $-22.6$ to $+35.7$ percentage points. These are bounds under missing judgments, not confidence intervals, and do not determine the direction of the quality effect. The selected task subset, missing judgments and shared model family of the feedback tool and evaluator limit generalization beyond this study.

\subsubsection{Post-delivery revision with error feedback}
\label{app:revisionstudy}
\noindent\textbf{Design.} We start from 40 frozen original projects, 20 each from DeepSeek-V4.1-Flash and GLM-5.3-Flash. Each project is independently resumed in three new sessions, one per condition, with an additional 30 minutes including inspection, editing, rendering and delivery (120 assigned runs). All conditions receive the original task, assets, editable project and video. \textbf{R0} requests ordinary continued improvement; \textbf{R1} asks the agent to derive a checklist, inspect the rendered video, repair discrepancies and recheck affected portions; \textbf{R2} supplies specific problems selected from multi-panel consensus failures, which the agent must verify against the original requirements before repairing. Original and revised videos are evaluated with the same Gemini 3.8 Flash protocol, blinded to condition and feedback; historical consensus judgments are used to select feedback, not as the before-revision scores.

\noindent\textbf{Coverage.} Table~\ref{tab:revisioncoverage} reports assignments, on-time deliveries and available scores. Of the assigned runs, 114 deliver on time, five produce no playable delivery and one delivers late. Scores are available for 103 revised videos and 38 original videos, yielding 99 original--revision pairs. This descriptive analysis includes all on-time artifacts; late, missing and unscored outputs remain separate, and unknown or n/a verdicts are not counted as passes or repairs.

\begin{table}[htbp]
\centering
\caption{\textbf{Post-delivery revision coverage.} Each condition starts from the same 20 original projects per system. Scored: revised videos with valid judgments. Paired: revisions with both original and revised judgments. R0/R1/R2 outputs from the same project are not independent original tasks.}
\label{tab:revisioncoverage}
\small
\begin{tabular}{@{}llrrrr@{}}
\toprule
System & Condition & Assigned & On time & Scored & Paired \\
\midrule
DeepSeek-V4.1-Flash & R0 & 20 & 19 & 16 & 16 \\
DeepSeek-V4.1-Flash & R1 & 20 & 19 & 18 & 18 \\
DeepSeek-V4.1-Flash & R2 & 20 & 19 & 16 & 15 \\
GLM-5.3-Flash & R0 & 20 & 20 & 17 & 16 \\
GLM-5.3-Flash & R1 & 20 & 18 & 17 & 16 \\
GLM-5.3-Flash & R2 & 20 & 19 & 19 & 18 \\
\bottomrule
\end{tabular}
\end{table}

\noindent\textbf{Condition comparisons.} Across the 99 observed pairs, mean explicit L2 failures decrease from 3.65 to 2.25. To assess the additional benefit of structured self-checking or supplied errors, we compare revised-video failure counts between conditions on the same original projects, using only projects scored in both conditions. Table~\ref{tab:revisioncontrasts} reports these paired differences with 95\% percentile intervals from 10,000 bootstrap resamples over original projects. All six intervals include zero: the comparisons do not establish an advantage of R1 or R2 over ordinary continuation. Missing scores, selected original projects and one execution per condition limit the interpretation of the observed improvements.

\begin{table}[htbp]
\centering
\caption{\textbf{Revision-condition comparisons.} Difference in revised-video explicit L2 failure counts, first condition minus second; negative values favor the first condition. $n$: original projects with scores for both conditions. Intervals resample paired original projects, not individual checks or pooled output rows.}
\label{tab:revisioncontrasts}
\small
\begin{tabular}{@{}llrrc@{}}
\toprule
System & Contrast & $n$ & Difference & 95\% CI \\
\midrule
DeepSeek-V4.1-Flash & R1$-$R0 & 15 & $-0.87$ & $[-2.27, +0.13]$ \\
DeepSeek-V4.1-Flash & R2$-$R0 & 12 & $-0.83$ & $[-1.92, +0.34]$ \\
DeepSeek-V4.1-Flash & R2$-$R1 & 15 & $+0.20$ & $[-1.00, +1.67]$ \\
GLM-5.3-Flash & R1$-$R0 & 13 & $-0.77$ & $[-2.46, +0.31]$ \\
GLM-5.3-Flash & R2$-$R0 & 15 & $-0.40$ & $[-1.00, +0.20]$ \\
GLM-5.3-Flash & R2$-$R1 & 16 & $-0.12$ & $[-1.19, +1.06]$ \\
\bottomrule
\end{tabular}
\end{table}

\section{Limitations and Broader Impact}
\label{app:ethics}

\noindent\textbf{Limitations.}
\begin{itemize}[leftmargin=1.9em,labelsep=0.5em,itemsep=2pt]
\item[\ico{stopwatch}] \textbf{Fixed budget.} The main evaluation uses a one-hour budget. The time-budget ablation (\S\ref{app:x13}) covers three systems and 20 tasks with budgets up to 120 minutes; broader evaluation remains future work.
\item[\ico{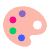}] \textbf{Adherence over creativity.} The rubrics mainly measure whether a video follows its specification. Creativity and originality are partly assessed through aesthetics (VA), but not independently.
\item[\ico{man-judge}] \textbf{Judge dependence.} All panels are proprietary models, and the judges are insensitive to audio-visual timing (App.~\ref{app:x11}). Rankings are stable across the three panels, but absolute values are panel-specific and are never mixed.
\item[\ico{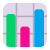}] \textbf{Coverage.} The task set is mostly English (209 of 239) and limited to 1080p-class outputs. Tasks with very long runtimes are few.
\end{itemize}

\runin{globe-with-meridians}{Broader impact.} Code-driven video makes precise, editable and accessible content (lessons, data stories, instructions) cheaper to produce, and our findings point to concrete reliability improvements. Risks include persuasive marketing content and misinformation rendered with an appearance of precision. Tasks with factual content ship their facts as inputs and forbid fabricated claims (\eg, ``do not claim a real interception probability''). No task depicts real private individuals. Classic-game tasks use newly designed levels and assets in a recognizable style for research evaluation; we will not redistribute original game assets.

\clearpage
\addtocontents{toc}{\protect\setcounter{tocdepth}{1}}  %
\section{Selected Cases}
\label{app:cases}
\noindent One case per category, each on one page: a frame from one delivered video (the system is named), the opening of the task specification, the supplied assets, the core goals with the number of L2 checks under each, one L2 check, and the five L3 dimensions with the anchor for a score of 5. Fail conditions are shown where the rubric names specific failures; the L3 scale and the shared observation points are in App.~\ref{app:rubric}.
\vspace{4pt}
\subsection{Brand \& marketing (C01)}
\label{app:case:D0032}
\noindent\caticon{C01}\hspace{0.4em}{\bfseries Ember Atlas: An Old TV Gateway to Three Adventure Worlds}\par\vspace{4pt}
\noindent\begin{minipage}[t]{0.42\textwidth}\vspace{0pt}
{\color{catC01}\rule{\linewidth}{1.2pt}}\par\vspace{1.5pt}
\includegraphics[width=\linewidth]{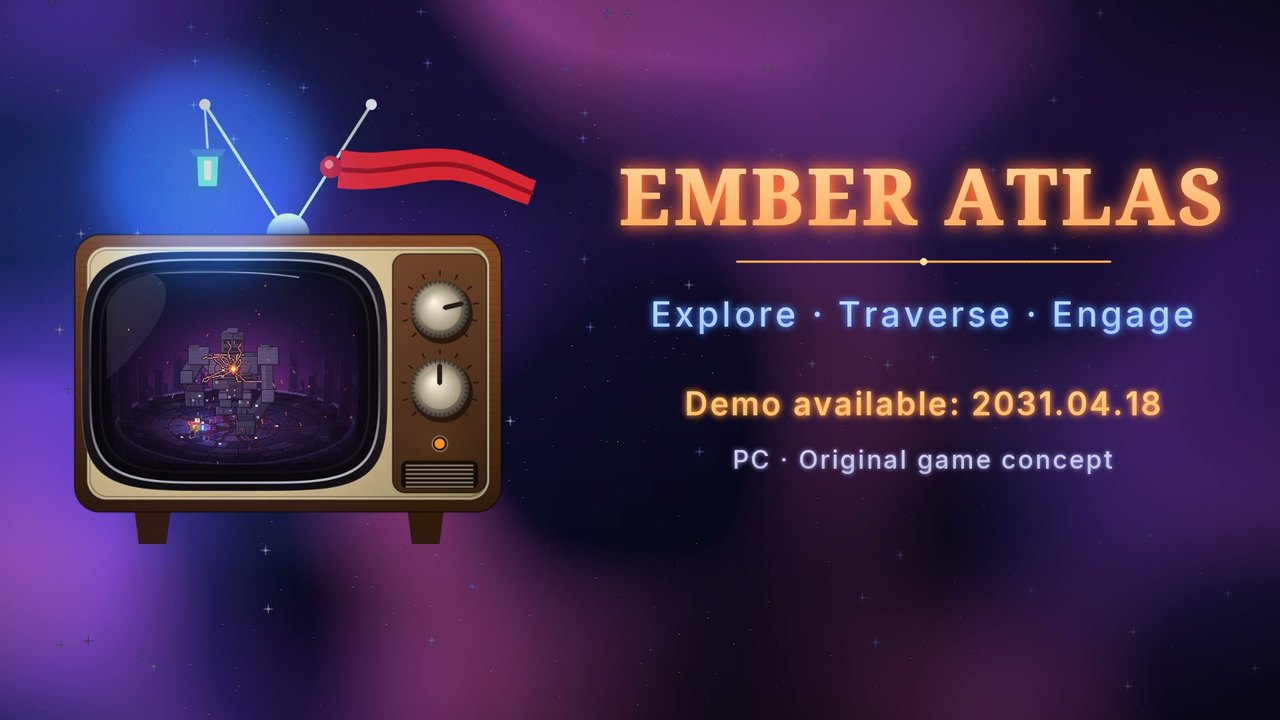}\par\vspace{2pt}
{\scriptsize\color{black!70}Frame at 28.0\,s of the video delivered by \logo{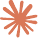}\,Claude Sonnet 5.5.}
\end{minipage}\hfill
\begin{minipage}[t]{0.555\textwidth}\vspace{0pt}
{\footnotesize\setlength{\tabcolsep}{3pt}\renewcommand{\arraystretch}{1.05}
\begin{tabular}{@{}l>{\raggedright\arraybackslash}p{0.68\linewidth}rr@{}}
\toprule Goal & Core goal & L2 & core \\ \midrule
G1 & Old TV interior/exterior relationship and opening & 3 & 3 \\
G2 & Actions and outcomes of the three gameplay modes & 8 & 5 \\
G3 & Television array and its initial takeoff & 4 & 4 \\
G4 & Fixed release card and original (fictional) identity & 4 & 4 \\
G5 & Music and action sounds & 8 & 6 \\
G6 & Complete delivery & 0 & 0 \\
\midrule & \textit{All L2 checks} & 27 & 22 \\ \bottomrule
\end{tabular}}
\end{minipage}\par\vspace{6pt}
\begin{specbox}{C01}{Task specification}{opening of 612 words}
\footnotesize\setlength{\parindent}{0pt}\setlength{\parskip}{2pt}
Ember Atlas: An Old TV Gateway to Three Adventure Worlds\par
For viewers who enjoy fantasy adventure games, create a fictional game trailer in which the old television visual motif connects three distinct gameplay types, culminating in a complete product launch.\\
Task type: structural-constraint task; structure: single-screen suspense --- three gameplay previews --- television array --- cosmic product reveal.\par
Content inputs\\
The game ``EMBER ATLAS'' is a fictional work for this task; do not use characters or battle footage from the original film ``Tales of Destiny'', or the PlayStation logo. The protagonist is a short-haired explorer wearing a red scarf and always carries a blue lantern. All footage must be self-drawn; pixel-art environments with simple 3D television shells may be used.\\
Three preview worlds and required actions: \circnum{1} Twilight Ruins: the protagonist uses the lantern to illuminate three symbols on a door, and the stone door opens; \circnum{2} Floating Islands: the protagonist jumps across a broken bridge, and the blue lantern lights up at the landing point; {[}\ldots{]}
\end{specbox}
\runin{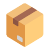}{Supplied assets.} None; everything is made from the specification.
\par\vspace{4pt}
\begin{l2box}{C01}{Example L2 check}{G2 $\cdot$ core $\cdot$ one of 27 checks}
\footnotesize\renewcommand{\arraystretch}{1.08}\setlength{\tabcolsep}{0pt}
\begin{tabular}{@{}>{\bfseries}p{5.3em}>{\raggedright\arraybackslash}p{\dimexpr\linewidth-5.3em\relax}@{}}
Pass & In Dark Arena, a circular red attack warning first appears and shrinks; the protagonist dodges outside the circle and then hits the glowing point on the statue's chest. \\
Prompt & \textit{``\circnum{3} Dark Arena: a circular red attack warning contracts, the protagonist dodges outside the circle, then strikes the bright point on the stone statue's chest.''} \\
\end{tabular}
\end{l2box}
\par\vspace{5pt}\noindent\ico{trophy}\hspace{0.3em}\textbf{L3 quality rubric:} task-specific anchors for a score of 5 in all five dimensions:
\par\vspace{2pt}{\footnotesize\setlength{\tabcolsep}{4pt}\renewcommand{\arraystretch}{1.08}
\noindent\begin{tabular}{@{}>{\raggedright\arraybackslash}p{0.17\textwidth}>{\raggedright\arraybackslash}p{\dimexpr0.83\textwidth-8pt\relax}@{}}
\toprule
\textbf{VQ} Visual quality & The small-screen gameplay/main character/publication text is clear and consistently effective for this task's purpose. \\
\textbf{VA} Visual aesthetics & The composition/visual language of the three worlds, old television, and starry-sky release is coordinated, with clear emphasis and suitability for a fantasy-adventure trailer; {[}\ldots{]} \\
\textbf{CL} Content logic & The action results of the three gameplay modes and the fictional release consistently and effectively support the purpose of this task. \\
\textbf{TM} Timing \& motion & The door light/jump and fall/circle-evading hit, together with the same TV lift, consistently and effectively support the purpose of this task. \\
\textbf{AQ} Audio \& A/V sync & The sound actually used is clear, layered, and naturally timed, with continuous positive evidence for the original adventure music and four event sounds; merely adding more sound layers earns no credit. \\
\bottomrule
\end{tabular}}
\clearpage
\subsection{Product \& feature demos (C02)}
\label{app:case:W0037}
\noindent\caticon{C02}\hspace{0.4em}{\bfseries Branchboard: See One Color Change Clearly}\par\vspace{4pt}
\noindent\begin{minipage}[t]{0.42\textwidth}\vspace{0pt}
{\color{catC02}\rule{\linewidth}{1.2pt}}\par\vspace{1.5pt}
\includegraphics[width=\linewidth]{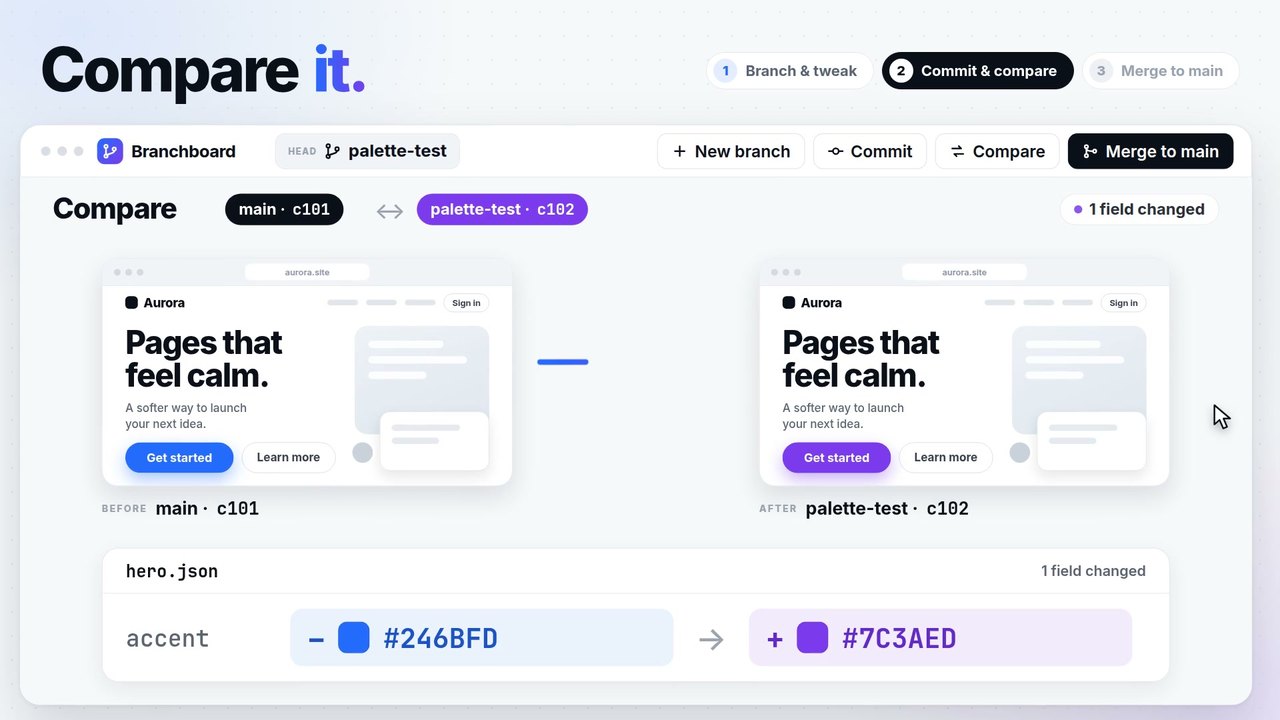}\par\vspace{2pt}
{\scriptsize\color{black!70}Frame at 8.1\,s of the video delivered by \logo{claude}\,Claude Sonnet 5.5.}
\end{minipage}\hfill
\begin{minipage}[t]{0.555\textwidth}\vspace{0pt}
{\footnotesize\setlength{\tabcolsep}{3pt}\renewcommand{\arraystretch}{1.05}
\begin{tabular}{@{}l>{\raggedright\arraybackslash}p{0.68\linewidth}rr@{}}
\toprule Goal & Core goal & L2 & core \\ \midrule
G1 & Initial state: main points to c101 & 12 & 11 \\
G2 & c102 is produced after clicking Commit & 15 & 14 \\
G3 & The opening displays ``Design it.\ Version it.'' & 9 & 5 \\
\midrule & \textit{All L2 checks} & 36 & 30 \\ \bottomrule
\end{tabular}}
\end{minipage}\par\vspace{6pt}
\begin{specbox}{C02}{Task specification}{opening of 563 words}
\footnotesize\setlength{\parindent}{0pt}\setlength{\parskip}{2pt}
Create an exactly 15-second promotional video in an app/SaaS style for the fictional app Branchboard, a visual tool that helps designers understand Git version changes. Use a light visual style with a \#F7F8FA interface, black and gray text, blue and purple accents, dramatic cuts, kinetic typography, and a dynamic Apple-style product presentation. All interface elements must be self-drawn.\par
The audience is designers who want to understand Git version changes. Use one clear design change to show a verifiable sequence of branch creation, modification, commit, comparison, and merge. The source data consists only of the accent field in the supplied file and one webpage Hero preview whose title remains unchanged.\par
Begin with the dynamic text ``Design it. Version it.'' and a blue preview. The initial state is main pointing to c101, accent=\#246BFD, a blue preview button, and a clean worktree. Show the user clicking the ``New branch'' control, creating a branch named palette-test from c101, and selecting it. In the design panel, change accent to \#7C3AED; the button becomes purple and the status becomes ``1 change''. Show the user clicking the ``Commit'' control to commit the change, producing c102 with the description ``Update accent''; palette-test points to c102 while main still points to c101.\par
Show a Compare panel containing only the accent change from blue to purple and the corresponding preview comparison; do not add unrelated files. Then show the user clicking the ``Merge to main'' control to merge palette-test to main. Because main has no other changes, the merge must fast-forward to c102; do not create a third merge commit. {[}\ldots{]}
\end{specbox}
\runin{package}{Supplied assets.} None; everything is made from the specification.
\par\vspace{4pt}
\begin{l2box}{C02}{Example L2 check}{G1 $\cdot$ core $\cdot$ one of 36 checks}
\footnotesize\renewcommand{\arraystretch}{1.08}\setlength{\tabcolsep}{0pt}
\begin{tabular}{@{}>{\bfseries}p{5.3em}>{\raggedright\arraybackslash}p{\dimexpr\linewidth-5.3em\relax}@{}}
Pass & The same button or the same Hero card serves as the persistent visual anchor throughout the video; the preview button is blue in its initial state and purple after the color change, corresponding to accent in the field and Compare; no particular button instance is mandatory \\
Prompt & \textit{``Use the same button or the same Hero card as a visual anchor throughout the video so that the color change can be found in the preview, the field, and the comparison.''} \\
\end{tabular}
\end{l2box}
\par\vspace{5pt}\noindent\ico{trophy}\hspace{0.3em}\textbf{L3 quality rubric:} task-specific anchors for a score of 5 in all five dimensions:
\par\vspace{2pt}{\footnotesize\setlength{\tabcolsep}{4pt}\renewcommand{\arraystretch}{1.08}
\noindent\begin{tabular}{@{}>{\raggedright\arraybackslash}p{0.17\textwidth}>{\raggedright\arraybackslash}p{\dimexpr0.83\textwidth-8pt\relax}@{}}
\toprule
\textbf{VQ} Visual quality & The visuals are clear and stable; composition and handling of occlusion/exposure consistently support viewing, and key details are in place. \\
\textbf{VA} Visual aesthetics & Visual style, color, material, and composition remain consistently coordinated and effectively serve viewing of this task's content. \\
\textbf{CL} Content logic & Information is clearly organized, key points are prominent, and the burden is moderate, enabling viewers to consistently understand the main relationships in this task. \\
\textbf{TM} Timing \& motion & Action trajectories and state changes are consistently coherent and easy to track; repeated sections are compressed naturally, and speed and holds leave appropriate viewing time for key actions and results. \\
\textbf{AQ} Audio \& A/V sync & The sound actually selected remains clear, natural, and appropriately layered, with positive evidence supporting the viewing purpose of ``using a unique accent change to let designers understand the verifiable states of branch creation, commit comparison {[}\ldots{]} \\
\bottomrule
\end{tabular}}
\clearpage
\subsection{Events \& personal expression (C03)}
\label{app:case:W0246}
\noindent\caticon{C03}\hspace{0.4em}{\bfseries ARI LIN: A 15-Second Motion Resume That Proves Each Item Before Checking It}\par\vspace{4pt}
\noindent\begin{minipage}[t]{0.42\textwidth}\vspace{0pt}
{\color{catC03}\rule{\linewidth}{1.2pt}}\par\vspace{1.5pt}
\includegraphics[width=\linewidth]{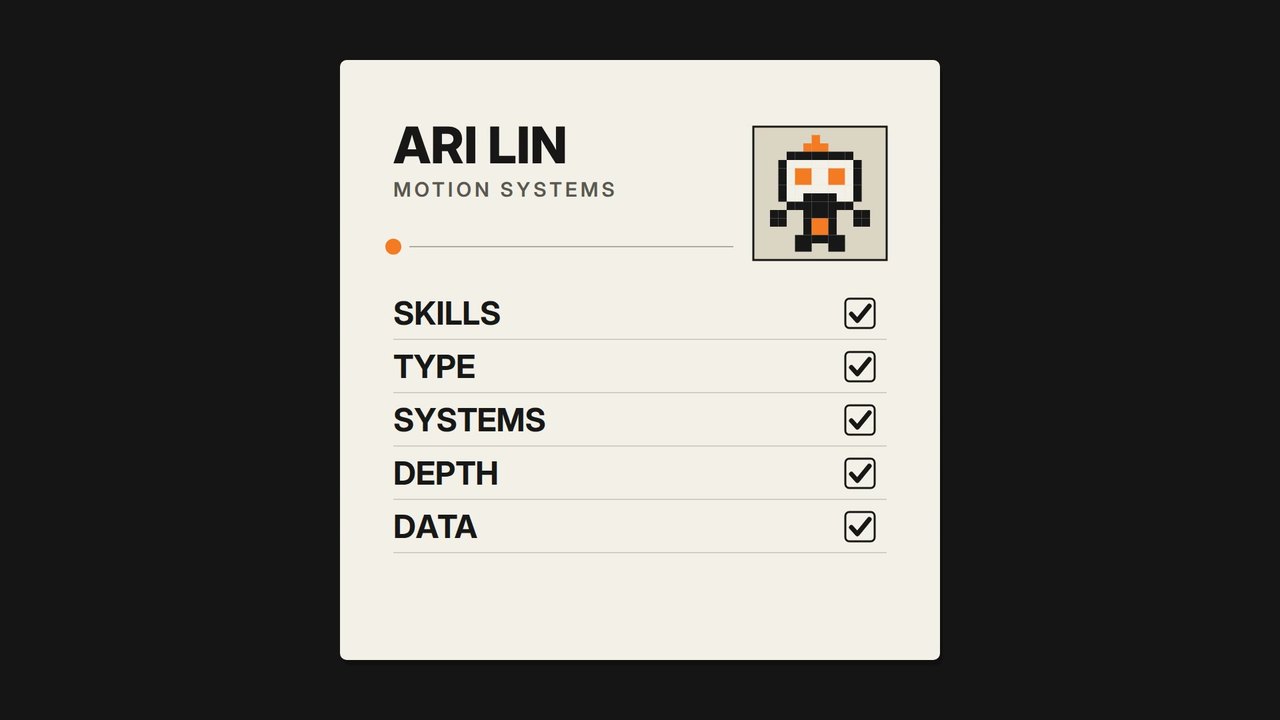}\par\vspace{2pt}
{\scriptsize\color{black!70}Frame at 14.2\,s of the video delivered by \logo{claude}\,Claude Opus 5.5.}
\end{minipage}\hfill
\begin{minipage}[t]{0.555\textwidth}\vspace{0pt}
{\footnotesize\setlength{\tabcolsep}{3pt}\renewcommand{\arraystretch}{1.05}
\begin{tabular}{@{}l>{\raggedright\arraybackslash}p{0.57\linewidth}rr@{}}
\toprule Goal & Core goal & L2 & core \\ \midrule
G\_STORY & Complete return to the same resume & 6 & 6 \\
G\_PROOF & Check off after five real proofs & 26 & 26 \\
G\_TIME & 128BPM eight-bar subdivision clock & 4 & 4 \\
G\_STYLE & Orange handoff and spring style & 3 & 3 \\
G\_AUDIO & Complete original synthesized audio track & 4 & 4 \\
G\_DELIVERY & Actual 900 frames/source code and proof record & 3 & 3 \\
\midrule & \textit{All L2 checks} & 46 & 46 \\ \bottomrule
\end{tabular}}
\end{minipage}\par\vspace{6pt}
\begin{specbox}{C03}{Task specification}{opening of 1,774 words}
\footnotesize\setlength{\parindent}{0pt}\setlength{\parskip}{2pt}
Create a 15-second motion-resume video for viewers browsing a code/motion-design portfolio. Use one resume to show actual visual proof of five capabilities, checking each item only after its proof is complete, and finally return to the complete identity page with the portrait and all checks. The resume identity is the original fictional character ARI LIN, with the role line MOTION SYSTEMS; do not add fictional employers, awards, or career achievements. The five rows, in order, are SKILLS, TYPE, SYSTEMS, DEPTH, and DATA. Use the original Bit Finch pixel character in the final photo frame; create and provide its 16x16 matrix and reference assets with the editable project, and use the same character for the later grid and photo frame. Build the whole piece from a dark page, a paper resume, and orange accent elements passed through the transitions; do not use corner labels or fictional HUD numbers.\par
{[}\ldots{]}\par
Clock and overall rhythm\\
The entire film is 128 BPM, 4/4 time, 8 bars, 32 quarter beats, and 15 seconds long. Each quarter beat is 0.46875 seconds, and each bar is 1.875 seconds. {[}\ldots{]}
\end{specbox}
\runin{package}{Supplied assets} (43 files, 2.0\,MB), read-only in the sandbox:
\par\vspace{2pt}\noindent\begin{minipage}[t]{0.6\textwidth}\vspace{0pt}
{\footnotesize\setlength{\tabcolsep}{4pt}\renewcommand{\arraystretch}{1.12}
\begin{tabularx}{\linewidth}{@{}c l r >{\raggedright\arraybackslash}X@{}}
\ico{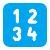} & Data & 14 & resume content, the mascot as a pixel matrix, a timing clock and cue grid, and structure, depth, spring and sound-design inputs \\
\ico{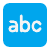} & Fonts & 24 & six font families with their licences and metadata \\
\ico{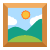} & Image & 2 & the pixel-art mascot reference (PNG and SVG) \\
\ico{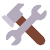} & Scripts & 2 & optional checkers for the media file and the observations \\
\ico{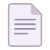} & Notes & 1 & a README \\
\end{tabularx}}
\end{minipage}\hfill\begin{minipage}[t]{0.37\textwidth}\vspace{0pt}\centering
{\setlength{\fboxsep}{0pt}\setlength{\fboxrule}{0.3pt}\fbox{\includegraphics[width=0.32\linewidth]{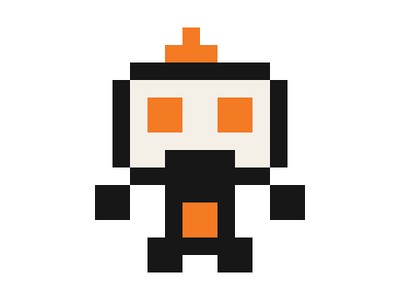}}}
\captionsetup{font=footnotesize,skip=3pt}\captionof{figure}{\textbf{Partial assets.} The supplied pixel-art mascot reference (one of four visual files; the others are an SVG version and copies).}
\end{minipage}
\par\vspace{4pt}
\begin{l2box}{C03}{Example L2 check}{G\_STORY $\cdot$ core $\cdot$ one of 46 checks}
\footnotesize\renewcommand{\arraystretch}{1.08}\setlength{\tabcolsep}{0pt}
\begin{tabular}{@{}>{\bfseries}p{5.3em}>{\raggedright\arraybackslash}p{\dimexpr\linewidth-5.3em\relax}@{}}
Pass & From b2.5 to b4, continuously push in from the original paper orange dot, until orange fills the screen at b4 and connects to the meter fill; there must be no switch to an unrelated background during the entire push-in, and correct endpoints alone cannot prove a continuous handoff. \\
Prompt & \textit{``At b2.5, push the camera into the orange dot; by b4, orange fills the frame and connects into the skill-meter fill. Do not cut to unrelated background during the push.''} \\
\end{tabular}
\end{l2box}
\par\vspace{5pt}\noindent\ico{trophy}\hspace{0.3em}\textbf{L3 quality rubric:} task-specific anchors for a score of 5 in all five dimensions:
\par\vspace{2pt}{\footnotesize\setlength{\tabcolsep}{4pt}\renewcommand{\arraystretch}{1.08}
\noindent\begin{tabular}{@{}>{\raggedright\arraybackslash}p{0.17\textwidth}>{\raggedright\arraybackslash}p{\dimexpr0.83\textwidth-8pt\relax}@{}}
\toprule
\textbf{VQ} Visual quality & Clear fine text and dot matrices/holes consistently and effectively support the purpose of this task. \\
\textbf{VA} Visual aesthetics & The paper resume and unified orange visual consistently and effectively support the purpose of this task. \\
\textbf{CL} Content logic & The five checkmarks corresponding to the five proofs consistently and effectively support the purpose of this task. \\
\textbf{TM} Timing \& motion & The spring-motion settling and orange handoff consistently and effectively support the purpose of this task. \\
\textbf{AQ} Audio \& A/V sync & The synthesized musical layers are clear, the beat points are discernible, and the dynamics are natural; the subdivided energy is coordinated with the five proofs and the resume return. \\
\bottomrule
\end{tabular}}
\clearpage
\subsection{Concept \& pipeline explainers (C04)}
\label{app:case:D0150}
\noindent\caticon{C04}\hspace{0.4em}{\bfseries How Wood Walks: Transmission and Foot Placement of an Eight-Legged Walker}\par\vspace{4pt}
\noindent\begin{minipage}[t]{0.42\textwidth}\vspace{0pt}
{\color{catC04}\rule{\linewidth}{1.2pt}}\par\vspace{1.5pt}
\includegraphics[width=\linewidth]{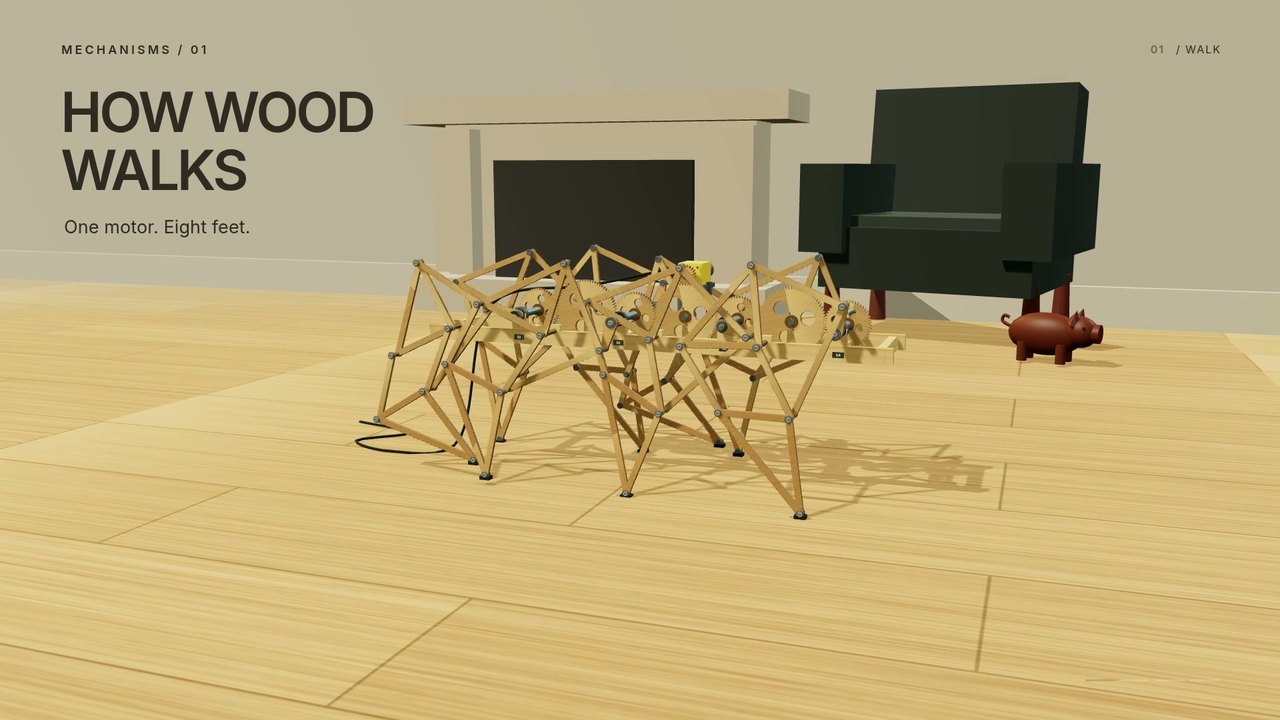}\par\vspace{2pt}
{\scriptsize\color{black!70}Frame at 3.8\,s of the video delivered by \logo{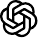}\,GPT-6 Astra.}
\end{minipage}\hfill
\begin{minipage}[t]{0.555\textwidth}\vspace{0pt}
{\footnotesize\setlength{\tabcolsep}{3pt}\renewcommand{\arraystretch}{1.05}
\begin{tabular}{@{}l>{\raggedright\arraybackslash}p{0.57\linewidth}rr@{}}
\toprule Goal & Core goal & L2 & core \\ \midrule
G1 & One complete machine unit across the four segments & 4 & 3 \\
G2 & Eight-legged closed chain, drive and leg phases & 23 & 22 \\
G3 & Support and displacement, elevated and on the ground & 16 & 15 \\
G4 & Material, identity and expression boundaries respected & 16 & 12 \\
G\_DELIVERY & Complete delivery: specified media parameters, editable project and task-specified attachments & 0 & 0 \\
\midrule & \textit{All L2 checks} & 59 & 52 \\ \bottomrule
\end{tabular}}
\end{minipage}\par\vspace{6pt}
\begin{specbox}{C04}{Task specification}{opening of 1,261 words}
\footnotesize\setlength{\parindent}{0pt}\setlength{\parskip}{2pt}
How Wood Walks: Transmission and Foot Placement of an Eight-Legged Walker\par
For general viewers interested in mechanical motion who have no background in linkage analysis, use a complete machine unit to answer how a continuously rotating motor makes a wooden machine take alternating steps, with the transmission, joints, and ground support visibly corresponding on the same machine.\\
Format: structural-constraint task; structure: observe walking on the floor---observe the leg cycle while elevated---trace the drive along the motor and gears---return to the floor to verify.\par
Content input\\
Use only the complete 01:13---01:50 walker unit retained from the original compilation, and independently produce a 37---45-second 3D short; do not mix in other electronic products, software tutorials, or the original channel intro. Preserve the complete structure of floor walking, elevated-base observation, motor/gear/bearing/yellow-motor close-ups, and return to the floor; the original footage is not a production input.\\
Use an unbranded eight-legged wooden walker; do not claim to reproduce the original clip's model or exact CAD. Use the supplied input data for all rod lengths, node connections, circle-intersection branches, gear phases, and leg phases. Model all scenes and mechanical parts independently; plywood texture may be procedural, with no external photos or models needed.\\
The machine has four jointed legs on each side, a single through-frame chassis, four transverse crankshafts, a small yellow motor, a visible rear cable, plywood gears, and metal pins/bearings. Keep stable leg IDs L1---L4 and R1---R4. {[}\ldots{]}
\end{specbox}
\runin{package}{Supplied assets} (2 files, 9\,KB), read-only in the sandbox:
\par\vspace{2pt}\noindent\begin{minipage}{0.86\textwidth}
{\footnotesize\setlength{\tabcolsep}{4pt}\renewcommand{\arraystretch}{1.12}
\begin{tabularx}{\linewidth}{@{}c l r >{\raggedright\arraybackslash}X@{}}
\ico{input-numbers} & Data & 2 & the walker's linkage (coordinates, bar lengths, fixed pivots, moving joints, rigid triangles) and a manifest \\
\end{tabularx}}
\end{minipage}
\par\vspace{4pt}
\begin{l2box}{C04}{Example L2 check}{G2 $\cdot$ core $\cdot$ one of 59 checks}
\footnotesize\renewcommand{\arraystretch}{1.08}\setlength{\tabcolsep}{0pt}
\begin{tabular}{@{}>{\bfseries}p{5.3em}>{\raggedright\arraybackslash}p{\dimexpr\linewidth-5.3em\relax}@{}}
Pass & Each connecting rod changes its angle around the actual pin shaft, with no rubber tube bending or stretching substituting for joint motion \\
Prompt & \textit{``Use the input's Jansen-style closed chain: two rigid triangular pieces and connecting rods change relative angles around pin joints; do not deform them like rubber tubes or let rod ends leave their joints.''} \\
\end{tabular}
\end{l2box}
\par\vspace{5pt}\noindent\ico{trophy}\hspace{0.3em}\textbf{L3 quality rubric:} task-specific anchors for a score of 5 in all five dimensions:
\par\vspace{2pt}{\footnotesize\setlength{\tabcolsep}{4pt}\renewcommand{\arraystretch}{1.08}
\noindent\begin{tabular}{@{}>{\raggedright\arraybackslash}p{0.17\textwidth}>{\raggedright\arraybackslash}p{\dimexpr0.83\textwidth-8pt\relax}@{}}
\toprule
\textbf{VQ} Visual quality & The same transmission chain and foot ends remain discernible throughout. \\
\textbf{VA} Visual aesthetics & The plywood-layer hole edges and metal bearings form a clear mechanical hierarchy throughout. \\
\textbf{CL} Content logic & The distinction between walking on the ground and being airborne, and the relationship from drive to foot end, remain clear throughout. \\
\textbf{TM} Timing \& motion & Rigid rods and pins move compatibly, and the feet's high return and low backsweep remain trackable throughout. \\
\textbf{AQ} Audio \& A/V sync & The sound actually used never interferes with mechanical understanding. \\
\bottomrule
\end{tabular}}
\clearpage
\subsection{Data \& fact stories (C05)}
\label{app:case:W0384}
\noindent\caticon{C05}\hspace{0.4em}{\bfseries Fifteen-Second Boomerang: LLM History and Imagined RSI}\par\vspace{4pt}
\noindent\begin{minipage}[t]{0.42\textwidth}\vspace{0pt}
{\color{catC05}\rule{\linewidth}{1.2pt}}\par\vspace{1.5pt}
\includegraphics[width=\linewidth]{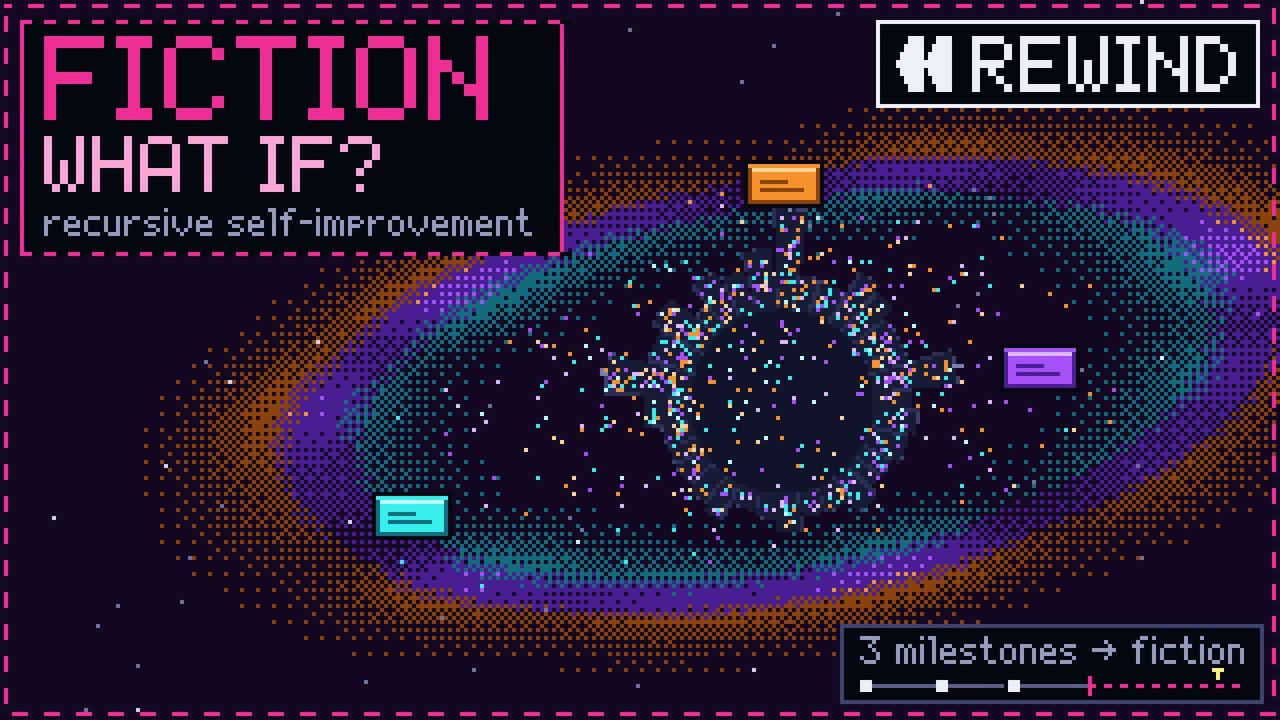}\par\vspace{2pt}
{\scriptsize\color{black!70}Frame at 7.9\,s of the video delivered by \logo{claude}\,Claude Opus 5.5.}
\end{minipage}\hfill
\begin{minipage}[t]{0.555\textwidth}\vspace{0pt}
{\footnotesize\setlength{\tabcolsep}{3pt}\renewcommand{\arraystretch}{1.05}
\begin{tabular}{@{}l>{\raggedright\arraybackslash}p{0.68\linewidth}rr@{}}
\toprule Goal & Core goal & L2 & core \\ \midrule
G1 & Three historical stages defined by the content clock u: u=0--1.5 seconds is 2003 Neural LM, u=1.5--3 seconds is 2017 Transformer, and u=3--4.5 seconds is 2020 GPT-3 {[}\ldots{]} & 9 & 9 \\
G2 & At u=4.5, when entering the future, a clear dashed boundary, FICTION, and RSI? appear; retain the question mark. RSI here is hypothetical feedback for recursive self-improvement {[}\ldots{]} & 7 & 7 \\
G3 & The complete forward progression runs from 0--7.5 seconds to the climax {[}\ldots{]} & 8 & 7 \\
\midrule & \textit{All L2 checks} & 24 & 23 \\ \bottomrule
\end{tabular}}
\end{minipage}\par\vspace{6pt}
\begin{specbox}{C05}{Task specification}{opening of 830 words}
\footnotesize\setlength{\parindent}{0pt}\setlength{\parskip}{2pt}
Pixel-art boomerang render of the history of LLMs. Then RSI. Then it goes fucking crazy. Create a 15-second pixel-art boomerang video for viewers with a basic interest in AI development. It should sequentially present milestones in LLM history, a fictional recursive-self-improvement feedback escalation, and a complete rewind back to the opening state.\par
Content and visuals\\
Select three representative milestones to give an overview of LLM prehistory, a key architecture, and scaling; do not claim to exhaust the history: 2003 Neural LM, 2017 Transformer, 2020 GPT-3. The 2003 segment depicts the prehistory of early neural language models, not the appearance of today's large models; 2017 depicts the Transformer architecture; 2020 depicts GPT-3 and its scaling phase. Do not treat the originally titled Opus5.5 as a production version for this task or as a historical milestone. The word blocks from each preceding milestone must continue transforming in the next; do not make three unrelated logos.\par
In the future segment, define RSI as Recursive Self-Improvement: an imagined feedback loop in which a system improves its own capabilities. The model proposing a plan, testing it, and forming the next version is only an artistic hypothesis; do not claim that autonomous unlimited intelligence growth has been achieved. When the forward content clock u reaches 4.5 seconds, FICTION and RSI? must clearly appear, with the question mark retained. During the rewind, as u falls below 4.5, FICTION and RSI? must exit with the original transition; do not leave them permanently visible based on finished-video time t$\geq$4.5. {[}\ldots{]}
\end{specbox}
\runin{package}{Supplied assets} (2 files, 11\,KB), read-only in the sandbox:
\par\vspace{2pt}\noindent\begin{minipage}{0.86\textwidth}
{\footnotesize\setlength{\tabcolsep}{4pt}\renewcommand{\arraystretch}{1.12}
\begin{tabularx}{\linewidth}{@{}c l r >{\raggedright\arraybackslash}X@{}}
\ico{input-numbers} & Data & 2 & the three historical stages with their rules and sources, and a structured brief \\
\end{tabularx}}
\end{minipage}
\par\vspace{4pt}
\begin{l2box}{C05}{Example L2 check}{G1 $\cdot$ core $\cdot$ one of 24 checks}
\footnotesize\renewcommand{\arraystretch}{1.08}\setlength{\tabcolsep}{0pt}
\begin{tabular}{@{}>{\bfseries}p{5.3em}>{\raggedright\arraybackslash}p{\dimexpr\linewidth-5.3em\relax}@{}}
Pass & The 2020 same-grid board expands, and after the example card enters, a new text block continues to emerge, expressing in-context few-shot; parameters are not changed during the demonstration \\
Fail & The animation trains the prompt example into the weights; only the name GPT-3 appears with no continuation process; or unprovided capability rankings are added. \\
Prompt & \textit{``Expand the same grid; after the example card enters, continue with new text blocks; do not animate parameter changes to represent few-shot.''} \\
\end{tabular}
\end{l2box}
\par\vspace{5pt}\noindent\ico{trophy}\hspace{0.3em}\textbf{L3 quality rubric:} task-specific anchors for a score of 5 in all five dimensions:
\par\vspace{2pt}{\footnotesize\setlength{\tabcolsep}{4pt}\renewcommand{\arraystretch}{1.08}
\noindent\begin{tabular}{@{}>{\raggedright\arraybackslash}p{0.17\textwidth}>{\raggedright\arraybackslash}p{\dimexpr0.83\textwidth-8pt\relax}@{}}
\toprule
\textbf{VQ} Visual quality & The key graphics from characters to city/nebula are clear and stable; the fictional and replay labels are discernible in the specified time windows. \\
\textbf{VA} Visual aesthetics & The cyan, orange, and purple visual language runs through the technical nodes, city, and nebula, with changes in form and layering serving a clear expressive purpose. \\
\textbf{CL} Content logic & Three real technology anchors lead to clearly labeled future speculation, with feedback and replay forming an understandable chain of imagery. \\
\textbf{TM} Timing \& motion & The 7.5-second forward motion and 7.5-second reverse motion return precisely along the same clock; states, labels, and the final fade-out form a clear rhythm. \\
\textbf{AQ} Audio \& A/V sync & The music/sound effects actually used support the forward/reverse imagery and do not interfere with the loop; when silence is permitted and actually used, AQ is N/A; pronunciation is not evaluated when there is no speech. \\
\bottomrule
\end{tabular}}
\clearpage
\subsection{Visual art \& form (C06)}
\label{app:case:W0298}
\noindent\caticon{C06}\hspace{0.4em}{\bfseries Painting the Night: A 50-Second Image in Paper, Graphite, and Moonlight}\par\vspace{4pt}
\noindent\begin{minipage}[t]{0.42\textwidth}\vspace{0pt}
{\color{catC06}\rule{\linewidth}{1.2pt}}\par\vspace{1.5pt}
\includegraphics[width=\linewidth]{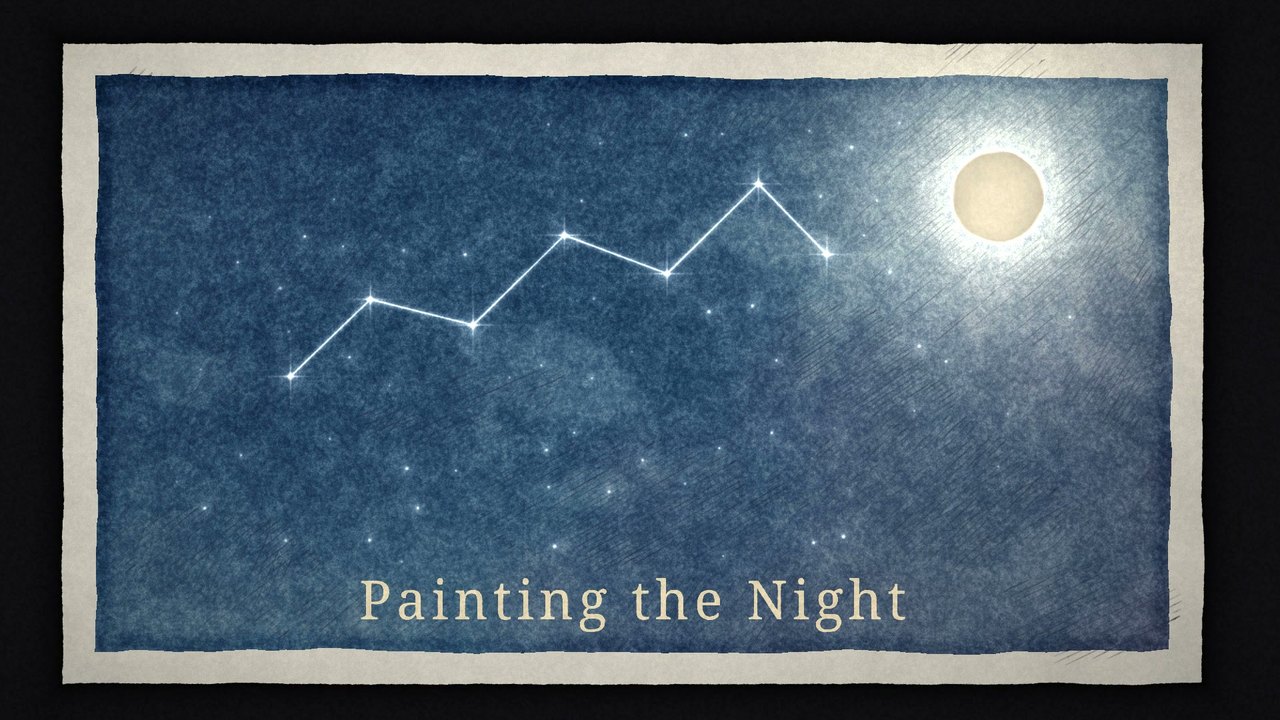}\par\vspace{2pt}
{\scriptsize\color{black!70}Frame at 50.0\,s of the video delivered by \logo{claude}\,Claude Sonnet 5.5.}
\end{minipage}\hfill
\begin{minipage}[t]{0.555\textwidth}\vspace{0pt}
{\footnotesize\setlength{\tabcolsep}{3pt}\renewcommand{\arraystretch}{1.05}
\begin{tabular}{@{}l>{\raggedright\arraybackslash}p{0.57\linewidth}rr@{}}
\toprule Goal & Core goal & L2 & core \\ \midrule
G\_MATERIAL & First establish visible warm-colored blank paper and fine-grained fibrous paper texture / Fix the paper texture to the same sheet, transforming consistently with the entire sheet {[}\ldots{]} & 9 & 9 \\
G\_MEANING & The same positions on the given paper surface as T1 through T7 become 7 main stars / The 6 connections follow T1$\rightarrow$T2$\rightarrow$T3$\rightarrow$T4$\rightarrow$T5$\rightarrow$T6$\rightarrow$T7 / The main stars and background stars have hierarchy {[}\ldots{]} & 11 & 11 \\
G\_AUDIO & There is complete 50-second original creative music, with soft percussion onset, piano-key pitch, and low/high dynamics {[}\ldots{]} & 5 & 5 \\
\midrule & \textit{All L2 checks} & 25 & 25 \\ \bottomrule
\end{tabular}}
\end{minipage}\par\vspace{6pt}
\begin{specbox}{C06}{Task specification}{opening of 1,707 words}
\footnotesize\setlength{\parindent}{0pt}\setlength{\parskip}{2pt}
Create a 50-second generative-art short for a general audience interested in generative art and short films, titled Painting the Night. On a single sheet of paper, continuously show a quiet sense of wonder developing from tentative marks into a complete night sky: materials accumulate, graphite imagery becomes a constellation, moonlight completes the sky, and the film settles into a quiet afterglow. Use zero image assets; paper grain, graphite hatching, the appearance of pigment, stars, the constellation, moonlight, and felt-piano music must all be generated at runtime. Implement the primary visuals with native JavaScript and WebGL.\par
Create one night sky on one sheet of paper: first show warm blank paper, then let tentative graphite hatching and a few dots accumulate on that same sheet; indigo pigment advances along irregular wet edges while the original paper grain and some graphite remain visible; the early dots become stars, and their existing connections become an original constellation; the early reserved circle gradually becomes a luminous moon, whose light affects the surrounding pigment and paper; finally show the complete painting and settle quietly. The key is the transformation of materials and meaning within the same motif. Do not present the paper grain, particles, and glow as three separate images pasted in one after another. No characters, conflict, competition, or scientific instruction are required.\par
{[}\ldots{]}\par
Geometry and identity: The paper design rectangle is {[}160,90,1600,900{]}, with UV increasing rightward and downward; measure circular distances in units normalized to paper height, and account for the horizontal 1600/900 ratio; {[}\ldots{]}
\end{specbox}
\runin{package}{Supplied assets} (14 files, 41\,KB), read-only in the sandbox:
\par\vspace{2pt}\noindent\begin{minipage}{0.86\textwidth}
{\footnotesize\setlength{\tabcolsep}{4pt}\renewcommand{\arraystretch}{1.12}
\begin{tabularx}{\linewidth}{@{}c l r >{\raggedright\arraybackslash}X@{}}
\ico{input-numbers} & Data & 9 & art direction (stages, palette, text), paper and constellation geometry, music, measurement and runtime contracts, and sources \\
\ico{hammer-and-wrench} & Scripts & 4 & optional checkers and an offline renderer \\
\ico{page-facing-up} & Notes & 1 & run notes \\
\end{tabularx}}
\end{minipage}
\par\vspace{4pt}
\begin{l2box}{C06}{Example L2 check}{G\_MATERIAL $\cdot$ core $\cdot$ one of 25 checks}
\footnotesize\renewcommand{\arraystretch}{1.08}\setlength{\tabcolsep}{0pt}
\begin{tabular}{@{}>{\bfseries}p{5.3em}>{\raggedright\arraybackslash}p{\dimexpr\linewidth-5.3em\relax}@{}}
Pass & The seven guide points are first established as graphite marks, then converted into the main stars. \\
Prompt & \textit{``The early dots become stars, and their existing connections become an original constellation.''} \\
\end{tabular}
\end{l2box}
\par\vspace{5pt}\noindent\ico{trophy}\hspace{0.3em}\textbf{L3 quality rubric:} task-specific anchors for a score of 5 in all five dimensions:
\par\vspace{2pt}{\footnotesize\setlength{\tabcolsep}{4pt}\renewcommand{\arraystretch}{1.08}
\noindent\begin{tabular}{@{}>{\raggedright\arraybackslash}p{0.17\textwidth}>{\raggedright\arraybackslash}p{\dimexpr0.83\textwidth-8pt\relax}@{}}
\toprule
\textbf{VQ} Visual quality & The seven-point hexagon and full moon remain stable and clear during the material progression; the title has a display interval within the final 3 seconds sufficient for reading (it is not required to run throughout the material development). \\
\textbf{VA} Visual aesthetics & The sparse stars and moon, together with the negative space, are orderly, and soft moonlight naturally opens up the space. \\
\textbf{CL} Content logic & The early points and connections on the same paper surface gradually become a constellation, and the reserved circle gradually becomes the moon; their relationship is clear and together they complete the night-sky imagery. {[}\ldots{]} \\
\textbf{TM} Timing \& motion & The local drawn lines and paint edges have a sense of growth, and the completion of the stars and moon and the ending are unhurried. \\
\textbf{AQ} Audio \& A/V sync & The felt-piano soft attacks and resonance are clear; the progression from sparse to open and then settling supports the material changes. \\
\bottomrule
\end{tabular}}
\clearpage
\subsection{Games \& sports (C07)}
\label{app:case:CG02}
\noindent\caticon{C07}\hspace{0.4em}{\bfseries Plants vs. Zombies: One Lawn Battle from 2D into 3D}\par\vspace{4pt}
\noindent\begin{minipage}[t]{0.42\textwidth}\vspace{0pt}
{\color{catC07}\rule{\linewidth}{1.2pt}}\par\vspace{1.5pt}
\includegraphics[width=\linewidth]{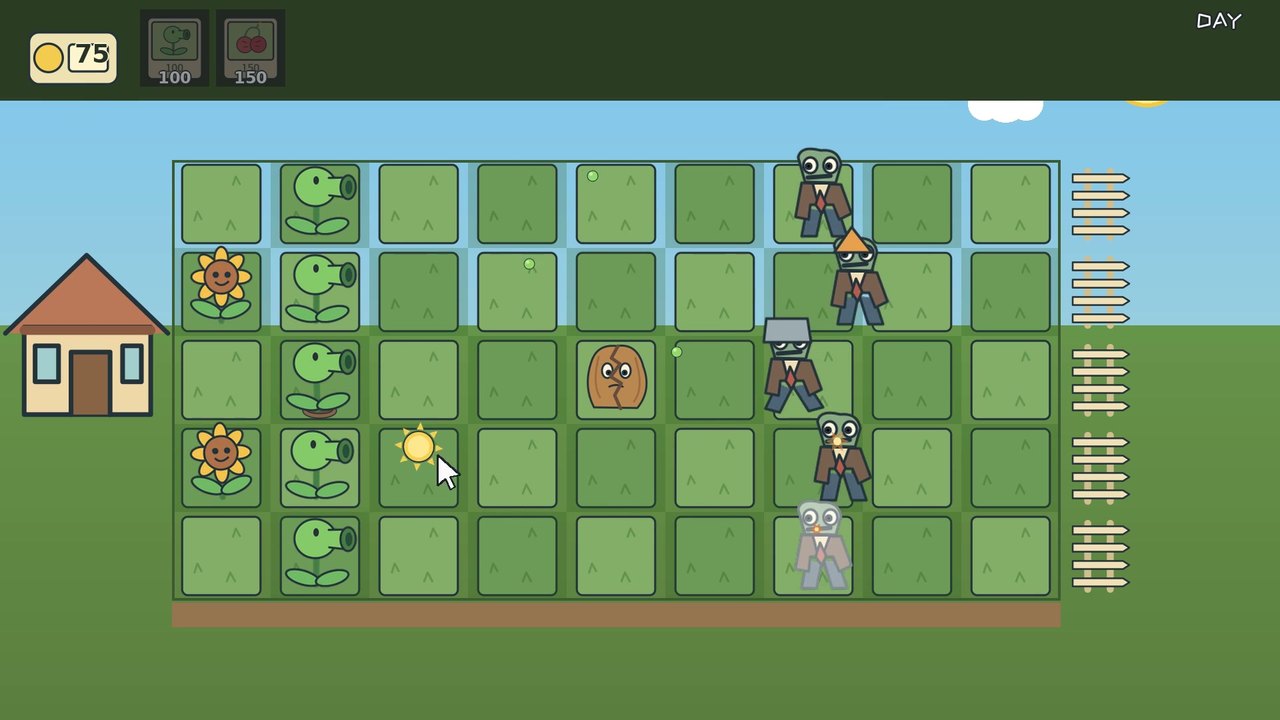}\par\vspace{2pt}
{\scriptsize\color{black!70}Frame at 12.9\,s of the video delivered by \logo{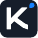}\,Kimi K3.}
\par\vspace{6pt}\centering
{\setlength{\fboxsep}{0pt}\setlength{\fboxrule}{0.3pt}\fbox{\includegraphics[width=0.18\linewidth]{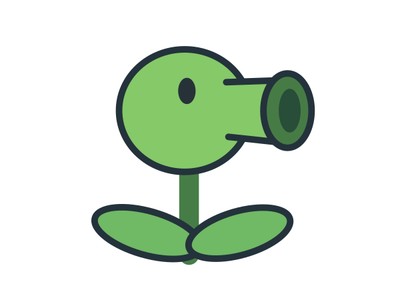}}\hspace{2pt}\fbox{\includegraphics[width=0.18\linewidth]{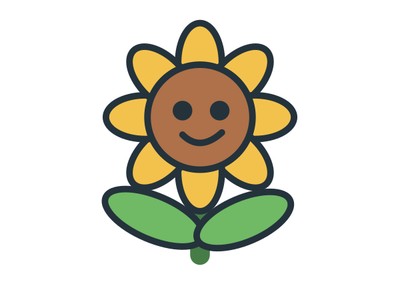}}\hspace{2pt}\fbox{\includegraphics[width=0.18\linewidth]{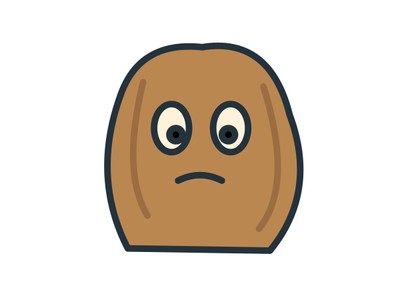}}\hspace{2pt}\fbox{\includegraphics[width=0.18\linewidth]{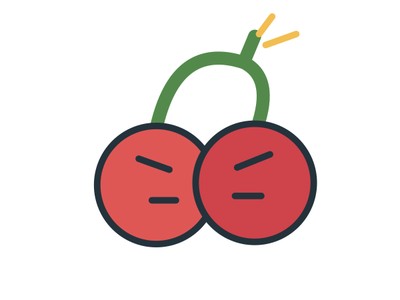}}\hspace{2pt}\fbox{\includegraphics[width=0.18\linewidth]{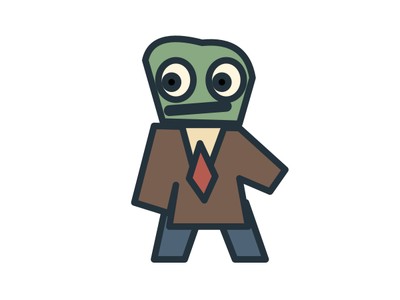}}\\[2pt]
\fbox{\includegraphics[width=0.18\linewidth]{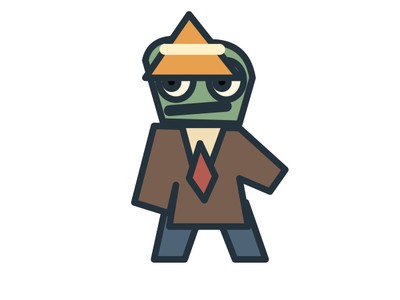}}\hspace{2pt}\fbox{\includegraphics[width=0.18\linewidth]{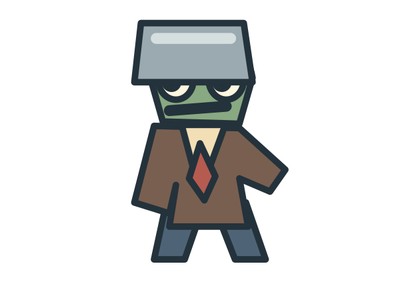}}\hspace{2pt}\fbox{\includegraphics[width=0.18\linewidth]{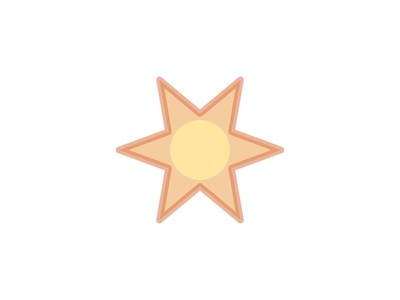}}\hspace{2pt}\fbox{\includegraphics[width=0.18\linewidth]{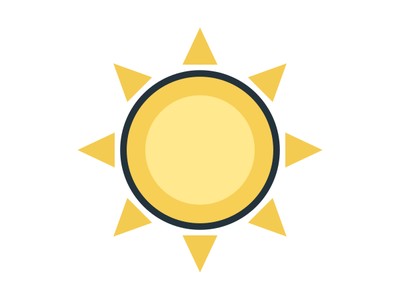}}\hspace{2pt}\fbox{\includegraphics[width=0.18\linewidth]{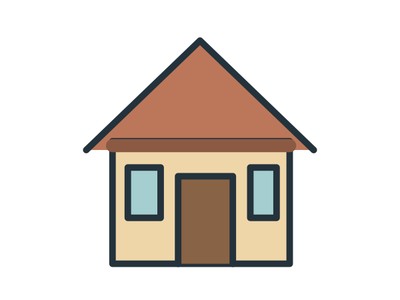}}}
\captionsetup{font=footnotesize,skip=3pt}\captionof{figure}{\textbf{Partial assets.} Ten of the 75 supplied SVG sprites, one per character or effect: plants, zombies, an explosion frame, the sun token and the house.}
\end{minipage}\hfill
\begin{minipage}[t]{0.555\textwidth}\vspace{0pt}
{\footnotesize\setlength{\tabcolsep}{3pt}\renewcommand{\arraystretch}{1.05}
\begin{tabular}{@{}l>{\raggedright\arraybackslash}p{0.68\linewidth}rr@{}}
\toprule Goal & Core goal & L2 & core \\ \midrule
G1 & Establish the classic five-row, nine-column daytime lawn with the specified initial formation, enemy identities, and gaps. & 6 & 6 \\
G2 & Complete P5 and CB1 through genuinely visible card selection, collection, and planting, strictly following the single sun ledger. & 8 & 8 \\
G3 & Make same-lane attacks, zombie movement, contact damage, and Wall-nut blocking form a traceable battle. & 7 & 7 \\
G4 & Transition from 2D to actual 3D within the same ongoing battle, and continue the identities, states, and clock. & 9 & 9 \\
G5 & Present the Wall-nut being chewed, three-lane gathering, effective Cherry rescue, and attacks clearing enemies under 3D pressure. & 8 & 8 \\
G6 & Conclude with a complete victory in this wave, with all 8 enemies eliminated and the residence unbreached. & 3 & 3 \\
G7 & Support gameplay with the event sounds required by the prompt, while preserving freedom to include optional music and narration. & 9 & 8 \\
G8 & Deliver an MP4 with sound in the specified specifications, an editable code project, the local assets used, a timeline, and rendering instructions. & 0 & 0 \\
\midrule & \textit{All L2 checks} & 50 & 49 \\ \bottomrule
\end{tabular}}
\end{minipage}\par\vspace{6pt}
\begin{specbox}{C07}{Task specification}{opening of 660 words}
\footnotesize\setlength{\parindent}{0pt}\setlength{\parskip}{2pt}
Plants vs. Zombies: One Lawn Battle from 2D into 3D\par
Creative brief\\
Use code to create a 90--120 second Plants vs. Zombies gameplay expansion video. Build around the classic daytime five-row, nine-column lawn, Sunflowers, Peashooters, Wall-nuts and Cherry Bombs, with regular, Conehead and Buckethead Zombies. Establish an active multi-lane defense in a clear 2D view, then transform that same ongoing battle into a 3D battlefield with visible volume and spatial relationships. Continue the defense to its conclusion so viewers recognize the classic gameplay and understand that the battle did not restart. {[}\ldots{]}
\end{specbox}
\runin{package}{Supplied assets} (141 files, 3.8\,MB), read-only in the sandbox:
\par\vspace{2pt}\noindent\begin{minipage}{0.86\textwidth}
{\footnotesize\setlength{\tabcolsep}{4pt}\renewcommand{\arraystretch}{1.12}
\begin{tabularx}{\linewidth}{@{}c l r >{\raggedright\arraybackslash}X@{}}
\ico{framed-picture} & Images & 75 & SVG sprites of every plant, zombie, effect and interface element, with animation frames \\
\ico{package} & 3D models & 25 & glTF models of the same characters, static and animated \\
\ico{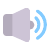} & Audio & 11 & sound effects for shooting, chewing, cracking, explosions, planting, sun pickup, transitions and victory \\
\ico{input-numbers} & Data & 29 & the level scenario, an asset manifest and supplement, sprite sequences and mesh descriptions \\
\ico{page-facing-up} & Notes & 1 & an asset guide \\
\end{tabularx}}
\end{minipage}
\par\vspace{4pt}
\begin{l2box}{C07}{Example L2 check}{G2 $\cdot$ core $\cdot$ one of 50 checks}
\footnotesize\renewcommand{\arraystretch}{1.08}\setlength{\tabcolsep}{0pt}
\begin{tabular}{@{}>{\bfseries}p{5.3em}>{\raggedright\arraybackslash}p{\dimexpr\linewidth-5.3em\relax}@{}}
Pass & First use 100 of 125 to plant P5, leaving 25; the five specified sun collection events in this round occur after that charge. \\
Fail & After planting, 100 was not deducted, the required sunlight was collected before payment, the cherry was planted first, or the balance cannot be matched to the actual operation. \\
Look at & From before P5 is selected until the first specified sunlight collection. \\
Prompt & \textit{``First spend 100 to add the Peashooter, collect five sun tokens to reach 150, then plant the Cherry Bomb and finish at 0.''} \\
\end{tabular}
\end{l2box}
\par\vspace{5pt}\noindent\ico{trophy}\hspace{0.3em}\textbf{L3 quality rubric:} task-specific anchors for a score of 5 in all five dimensions:
\par\vspace{2pt}{\footnotesize\setlength{\tabcolsep}{4pt}\renewcommand{\arraystretch}{1.08}
\noindent\begin{tabular}{@{}>{\raggedright\arraybackslash}p{0.17\textwidth}>{\raggedright\arraybackslash}p{\dimexpr0.83\textwidth-8pt\relax}@{}}
\toprule
\textbf{VQ} Visual quality & Key content is clear, the image is stable, and visual details support viewing. \\
\textbf{VA} Visual aesthetics & The visual language is consistent, emphasis is clear, and details are suitable for this task's use. \\
\textbf{CL} Content logic & Core relationships and development are clear, the content is coherently organized, and it meets audience needs. \\
\textbf{TM} Timing \& motion & Key changes are clear, with motion, transitions, and holds naturally serving the task. \\
\textbf{AQ} Audio \& A/V sync & The sound is clear, with natural layering; its style and coordination with events effectively support expression. \\
\bottomrule
\end{tabular}}
\clearpage
\subsection{Story \& visual comedy (C08)}
\label{app:case:D0173}
\noindent\caticon{C08}\hspace{0.4em}{\bfseries The Snowfield Cache and the Overrun Camp: One Failed Search by Six Survivors}\par\vspace{4pt}
\noindent\begin{minipage}[t]{0.42\textwidth}\vspace{0pt}
{\color{catC08}\rule{\linewidth}{1.2pt}}\par\vspace{1.5pt}
\includegraphics[width=\linewidth]{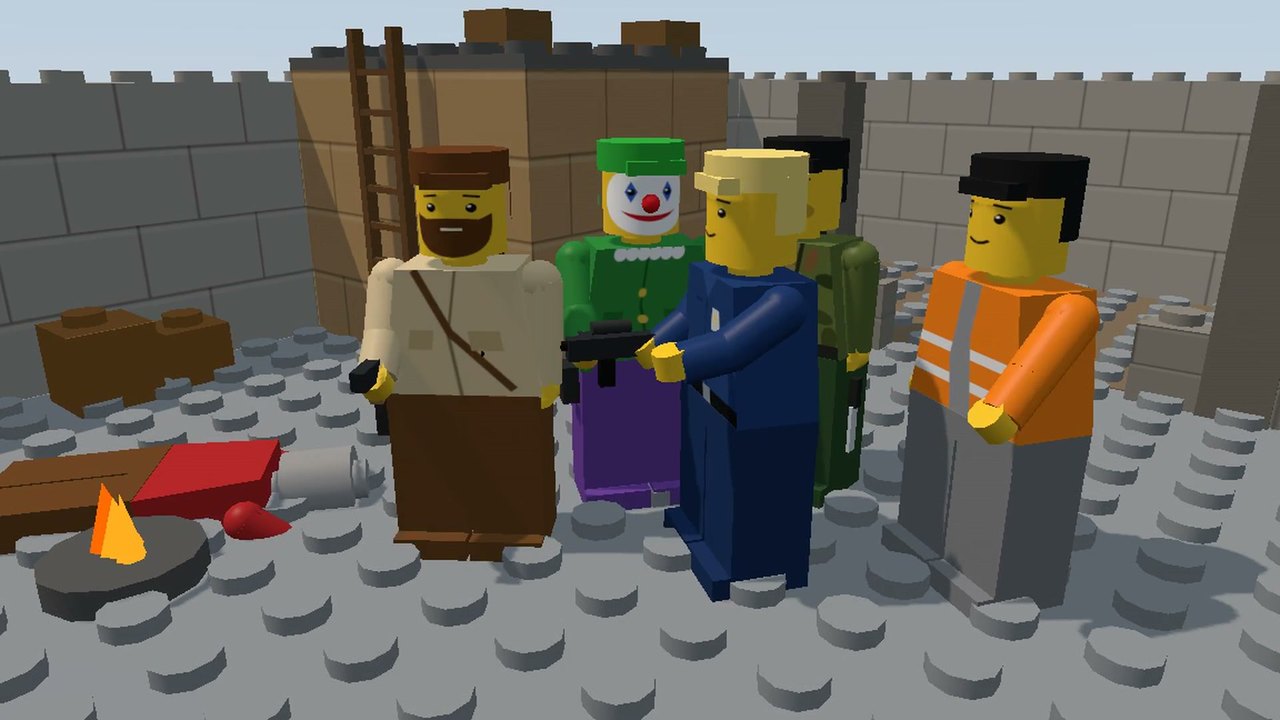}\par\vspace{2pt}
{\scriptsize\color{black!70}Frame at 41.0\,s of the video delivered by \logo{claude}\,Claude Opus 5.5.}
\end{minipage}\hfill
\begin{minipage}[t]{0.555\textwidth}\vspace{0pt}
{\footnotesize\setlength{\tabcolsep}{3pt}\renewcommand{\arraystretch}{1.05}
\begin{tabular}{@{}l>{\raggedright\arraybackslash}p{0.68\linewidth}rr@{}}
\toprule Goal & Core goal & L2 & core \\ \midrule
G1 & Independent dual storylines and the causality of failure & 12 & 11 \\
G2 & Identity, injuries, and prop states & 31 & 31 \\
G3 & Three-dimensional building-block action and space & 20 & 19 \\
G4 & Ambush, disarmament, and everyone's capture & 23 & 23 \\
G5 & Actual complete delivery & 0 & 0 \\
\midrule & \textit{All L2 checks} & 86 & 84 \\ \bottomrule
\end{tabular}}
\end{minipage}\par\vspace{6pt}
\begin{specbox}{C08}{Task specification}{opening of 1,707 words}
\footnotesize\setlength{\parindent}{0pt}\setlength{\parskip}{2pt}
The Snowfield Cache and the Overrun Camp: One Failed Search by Six Survivors\par
For viewers watching a complete brick adventure with action performance and causal storytelling, adapt the complete caption into a self-contained brick stop-motion-style short film that does not require a previous installment. The characters' identities, injuries, prop ownership, and failed ending must all be traceable from the images.\\
Narrative order: clearing and splitting up --- cross-cutting between the snowfield search and the camp injury --- ambush on the return --- the entire group captured.\par
Content input\par
{[}\ldots{]}\par
Based on the entire {[}task{]} story, make a complete 180--200 second video. Do not copy the source-film title, any watch-the-previous-installment prompt, or a list of other makers. Independent premise: six survivors are temporarily staying at an abandoned camp. After clearing zombies from the surrounding area, they send two people to a snowfield watchtower to search for supplies. The search ultimately fails and the entire group is captured. You may use THE WHITE CACHE as a creator-original title.\\
For any non-subject display labels in the supplied source file, use the English role, appearance, fate, scene, prop, and event wording in this task for production. Preserve every source ID, filename, coordinate, numerical value, and state value unchanged.\\
The six survivors are fixed as follows: P is a blond-haired police officer in a blue uniform; B is a brown-bearded explorer in beige; {[}\ldots{]}
\end{specbox}
\runin{package}{Supplied assets} (7 files, 43\,KB), read-only in the sandbox:
\par\vspace{2pt}\noindent\begin{minipage}{0.86\textwidth}
{\footnotesize\setlength{\tabcolsep}{4pt}\renewcommand{\arraystretch}{1.12}
\begin{tabularx}{\linewidth}{@{}c l r >{\raggedright\arraybackslash}X@{}}
\ico{input-numbers} & Data & 4 & characters, hostiles, props, scene and events; a ledger of the survivors' states; font provenance; a manifest \\
\ico{input-latin-letters} & Font & 1 & a monospaced display font \\
\ico{page-facing-up} & Notes & 2 & font licence and usage notes \\
\end{tabularx}}
\end{minipage}
\par\vspace{4pt}
\begin{l2box}{C08}{Example L2 check}{G1 $\cdot$ core $\cdot$ one of 86 checks}
\footnotesize\renewcommand{\arraystretch}{1.08}\setlength{\tabcolsep}{0pt}
\begin{tabular}{@{}>{\bfseries}p{5.3em}>{\raggedright\arraybackslash}p{\dimexpr\linewidth-5.3em\relax}@{}}
Pass & Only after O has stood up holding the bone staff in the left hand and tried walking does R leave through the previously established guard-post door carrying a shotgun, without passing through a wall with no entrance; no interval, number of shots, or exit posture is specified. \\
Fail & Fail if R leaves the sentry opening before O completes standing up and attempting to walk while holding the bone cane in the left hand, or is not holding a shotgun, or actually passes through a wall without a doorway; do not deduct points for unspecified poses, intervals, or numbers of shots. \\
Prompt & \textit{``O braces with the left hand, stands, and tests walking while the right arm remains missing. {[}...{]} R leaves the lookout with the shotgun.''} \\
\end{tabular}
\end{l2box}
\par\vspace{5pt}\noindent\ico{trophy}\hspace{0.3em}\textbf{L3 quality rubric:} task-specific anchors for a score of 5 in all five dimensions:
\par\vspace{2pt}{\footnotesize\setlength{\tabcolsep}{4pt}\renewcommand{\arraystretch}{1.08}
\noindent\begin{tabular}{@{}>{\raggedright\arraybackslash}p{0.17\textwidth}>{\raggedright\arraybackslash}p{\dimexpr0.83\textwidth-8pt\relax}@{}}
\toprule
\textbf{VQ} Visual quality & The block identity and left/right injuries remain discernible and effectively support this task's purpose. \\
\textbf{VA} Visual aesthetics & The plastic-stage atmosphere in the snow and camp consistently and effectively supports the purpose of this task. \\
\textbf{CL} Content logic & The causality behind the convergence of the two storylines and the outcome of failure consistently and effectively supports the purpose of this task. \\
\textbf{TM} Timing \& motion & Joints/climbing/bone-staff weight-bearing and handoff consistently and effectively support the purpose of this task. \\
\textbf{AQ} Audio \& A/V sync & The sound actually used is clear, with natural layering and timing; there is continuous positive evidence that optional sound supports threat and brief relief; do not gain points by adding more sound layers. \\
\bottomrule
\end{tabular}}
\clearpage
\subsection{Building \& process (C09)}
\label{app:case:W0137}
\noindent\caticon{C09}\hspace{0.4em}{\bfseries Embed Voxel Explosions into a Real Street: From 15 TNT Blocks to Two Road Craters}\par\vspace{4pt}
\noindent\begin{minipage}[t]{0.42\textwidth}\vspace{0pt}
{\color{catC09}\rule{\linewidth}{1.2pt}}\par\vspace{1.5pt}
\includegraphics[width=\linewidth]{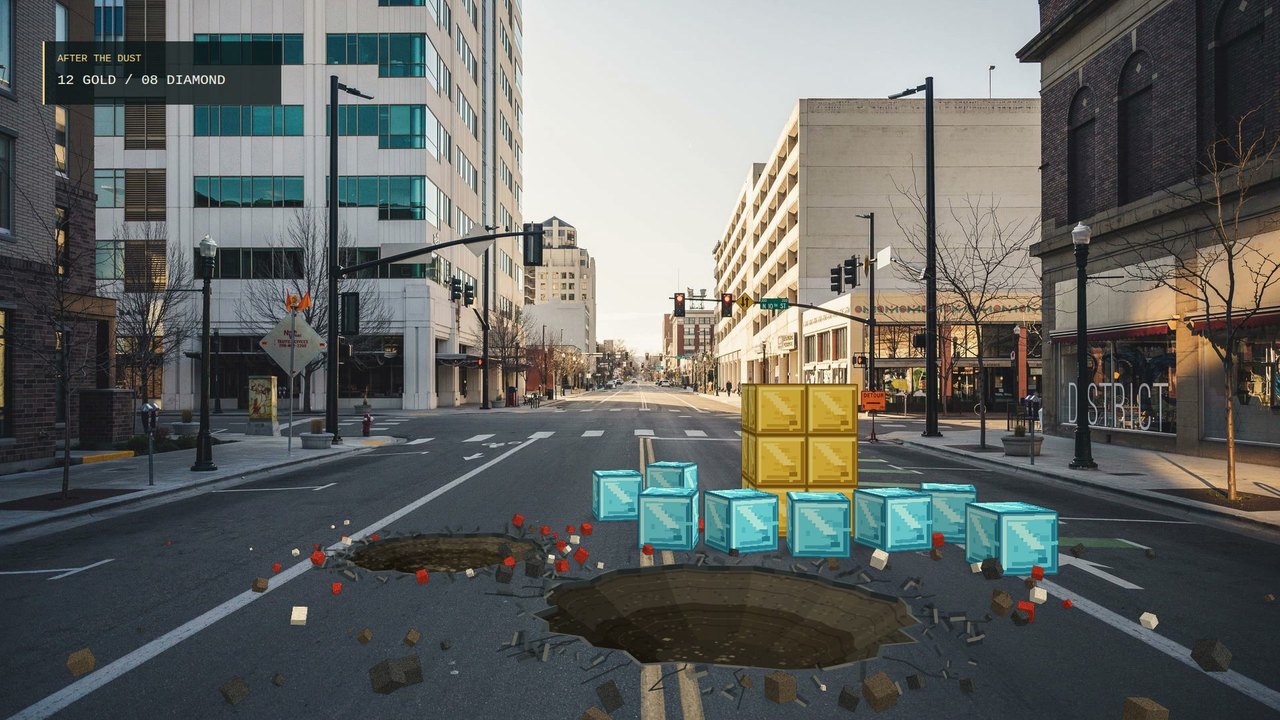}\par\vspace{2pt}
{\scriptsize\color{black!70}Frame at 30.4\,s of the video delivered by \logo{openai}\,GPT-6 Astra.}
\end{minipage}\hfill
\begin{minipage}[t]{0.555\textwidth}\vspace{0pt}
{\footnotesize\setlength{\tabcolsep}{3pt}\renewcommand{\arraystretch}{1.05}
\begin{tabular}{@{}l>{\raggedright\arraybackslash}p{0.68\linewidth}rr@{}}
\toprule Goal & Core goal & L2 & core \\ \midrule
G1 & The background uses the attached real street photograph & 12 & 12 \\
G2 & Ordinary TNT appears on the left side of the road & 32 & 30 \\
G3 & The openings of the two final pits are separated from each other & 16 & 16 \\
\midrule & \textit{All L2 checks} & 60 & 58 \\ \bottomrule
\end{tabular}}
\end{minipage}\par\vspace{6pt}
\begin{specbox}{C09}{Task specification}{opening of 1,486 words}
\footnotesize\setlength{\parindent}{0pt}\setlength{\parskip}{2pt}
Create a Minecraft-style voxel animation that integrates block construction and destruction into a real street scene. Use the street photograph in the input package as the full-frame background. Stack 15 TNT blocks one by one on the foreground road, build a tower from gold and diamond blocks, then have a giant TNT block fall from the sky and trigger an explosion that creates two craters in the actual road surface. The output must be a complete, coherent, directly playable video.\par
{[}\ldots{]}\par
First let the viewer clearly see the photographed street, including the real buildings on both sides, the empty foreground road, and the yellow road centerline. People and vehicles visible in the distance remain part of the background; {[}\ldots{]}
\end{specbox}
\runin{package}{Supplied assets} (9 files, 0.4\,MB), read-only in the sandbox:
\par\vspace{2pt}\noindent\begin{minipage}[t]{0.6\textwidth}\vspace{0pt}
{\footnotesize\setlength{\tabcolsep}{4pt}\renewcommand{\arraystretch}{1.12}
\begin{tabularx}{\linewidth}{@{}c l r >{\raggedright\arraybackslash}X@{}}
\ico{framed-picture} & Images & 2 & a real street photograph and a ground reference that outlines the crater sites on it \\
\ico{input-numbers} & Data & 4 & blocks, materials, build order, craters and story; photo and font provenance; a manifest \\
\ico{input-latin-letters} & Font & 1 & a monospaced display font \\
\ico{page-facing-up} & Notes & 2 & font licence, source and usage notes \\
\end{tabularx}}
\end{minipage}\hfill\begin{minipage}[t]{0.37\textwidth}\vspace{0pt}\centering
{\setlength{\fboxsep}{0pt}\setlength{\fboxrule}{0.3pt}\fbox{\includegraphics[width=0.48\linewidth]{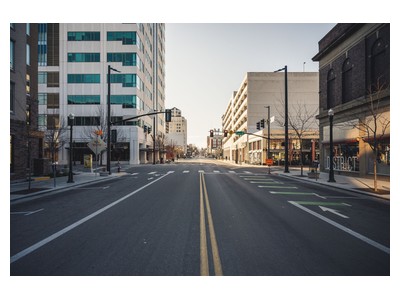}}\hspace{2pt}\fbox{\includegraphics[width=0.48\linewidth]{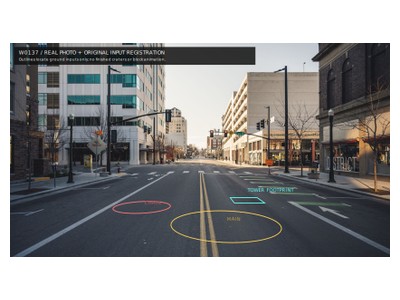}}}
\captionsetup{font=footnotesize,skip=3pt}\captionof{figure}{\textbf{Supplied images.} The street photograph that is the full-frame background (left) and the ground reference that outlines the crater sites on it (right).}
\end{minipage}
\par\vspace{4pt}
\begin{l2box}{C09}{Example L2 check}{G2 $\cdot$ core $\cdot$ one of 60 checks}
\footnotesize\renewcommand{\arraystretch}{1.08}\setlength{\tabcolsep}{0pt}
\begin{tabular}{@{}>{\bfseries}p{5.3em}>{\raggedright\arraybackslash}p{\dimexpr\linewidth-5.3em\relax}@{}}
Pass & 15 ordinary TNT blocks can be individually tracked from actual entry through placement, without using a counter or a screenshot of the final pile in place of 15 entries \\
Look at & Track each of T01--T15 entering, contacting support and settling, without requiring every block to be visible in the final frame or relying on a counting caption. \\
Prompt & \textit{``For the 15-block TNT stack, the viewer must still see every block enter and settle, followed by a brief view of the complete stack. A numeric counter is optional, but do not merely draw a pile and label it ``15.''''} \\
\end{tabular}
\end{l2box}
\par\vspace{5pt}\noindent\ico{trophy}\hspace{0.3em}\textbf{L3 quality rubric:} task-specific anchors for a score of 5 in all five dimensions:
\par\vspace{2pt}{\footnotesize\setlength{\tabcolsep}{4pt}\renewcommand{\arraystretch}{1.08}
\noindent\begin{tabular}{@{}>{\raggedright\arraybackslash}p{0.17\textwidth}>{\raggedright\arraybackslash}p{\dimexpr0.83\textwidth-8pt\relax}@{}}
\toprule
\textbf{VQ} Visual quality & The visuals are clear and stable; composition and handling of occlusion/exposure consistently support viewing, and key details are in place. \\
\textbf{VA} Visual aesthetics & Visual style, color, material, and composition remain consistently coordinated and effectively serve viewing of this task's content. \\
\textbf{CL} Content logic & Information is clearly organized, key points are prominent, and the burden is moderate, enabling viewers to consistently understand the main relationships in this task. \\
\textbf{TM} Timing \& motion & Action trajectories and state changes are consistently coherent and easy to track; repeated sections are compressed naturally, and speed and holds leave appropriate viewing time for key actions and results. \\
\textbf{AQ} Audio \& A/V sync & The actually selected sound remains clear, natural, and appropriately layered {[}\ldots{]} \\
\bottomrule
\end{tabular}}
\clearpage
\subsection{Music \& sound-driven (C10)}
\label{app:case:W0225}
\noindent\caticon{C10}\hspace{0.4em}{\bfseries A 15-Second Code Film About Its Own Prompt: Score, Pure Time-Based Drawing, Real FFT, and Beat-by-Beat Self-Audit}\par\vspace{4pt}
\noindent\begin{minipage}[t]{0.42\textwidth}\vspace{0pt}
{\color{catC10}\rule{\linewidth}{1.2pt}}\par\vspace{1.5pt}
\includegraphics[width=\linewidth]{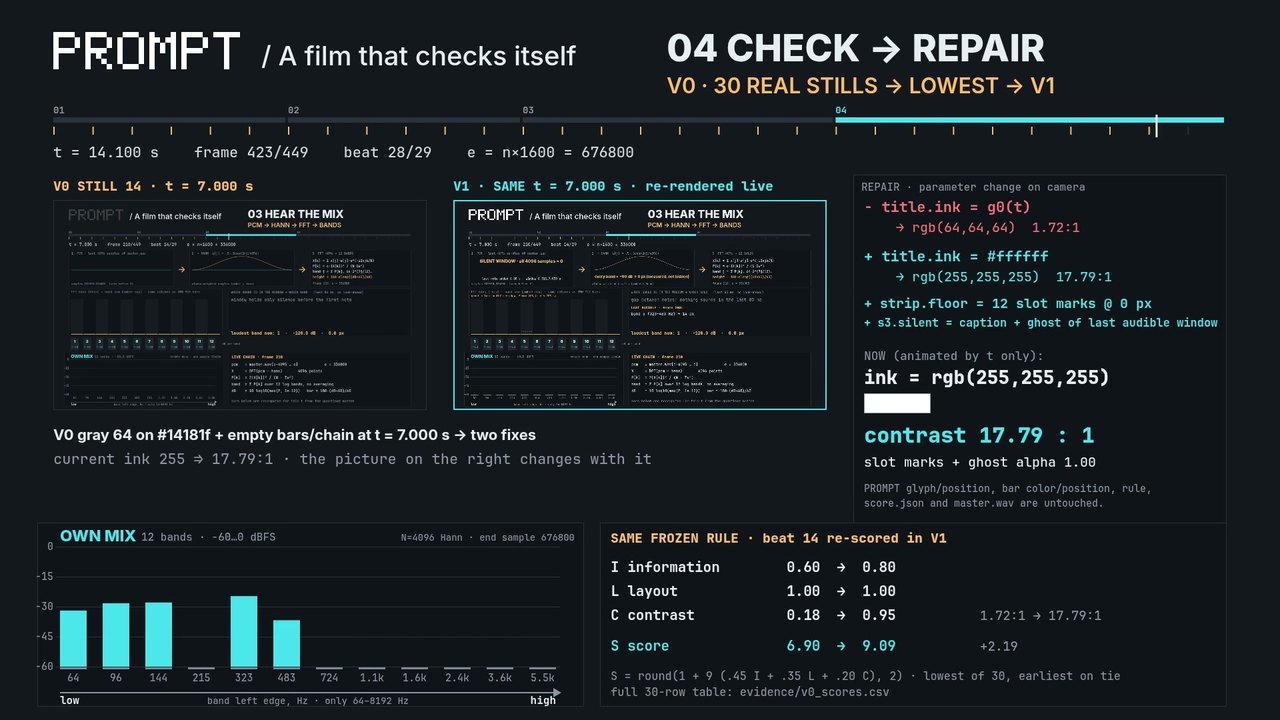}\par\vspace{2pt}
{\scriptsize\color{black!70}Frame at 14.1\,s of the video delivered by \logo{claude}\,Claude Sonnet 5.5.}
\end{minipage}\hfill
\begin{minipage}[t]{0.555\textwidth}\vspace{0pt}
{\footnotesize\setlength{\tabcolsep}{3pt}\renewcommand{\arraystretch}{1.05}
\begin{tabular}{@{}l>{\raggedright\arraybackslash}p{0.68\linewidth}rr@{}}
\toprule Goal & Core goal & L2 & core \\ \midrule
G1 & Complete 15-second four-step fixed time window, original code score, and genuinely audible music. & 25 & 19 \\
G2 & Purely time-deterministic seek, fixed runtime verification, and authentic FFT of the actual mix. & 25 & 21 \\
G3 & Freeze V0, use the lowest score from all 30 real scores, make a real repair attempt at the same t, and report the result honestly & 29 & 24 \\
\midrule & \textit{All L2 checks} & 79 & 64 \\ \bottomrule
\end{tabular}}
\end{minipage}\par\vspace{6pt}
\begin{specbox}{C10}{Task specification}{opening of 2,215 words}
\footnotesize\setlength{\parindent}{0pt}\setlength{\parskip}{2pt}
Create a complete 15-second code film for general viewers interested in generative audiovisual work and reproducible creative processes. The film must help viewers understand how audiovisuals are generated from the film's own prompt and must show real screenshots of checking and repair. Present the four steps in this order: score first, pure seek-based drawing, the film's own mix spectrum, and beat-by-beat scoring of candidate versions followed by actual lowest-score-frame repair. The viewer must see the actual object or state change at every step; do not merely replace four title cards.\par
{[}\ldots{]}\par
All four steps must be executed. The 15 seconds must be a complete short film with a beginning, production logic, checking, and a repair resolution, not an excerpt from a longer work. Do not add model, API, or paid-tool requirements. The C10 sound-driven component is central: the score, drawing, spectrum, and self-audit mechanisms must be understandable from the actual picture and sound.\par
Implement 0 samples as using no prerecorded audio clips, sample libraries, or recorded voice; digital PCM still contains discrete samples and must not be described as having no sample rate. Implement 0 keyframes as using no manual animation-keyframe system; note events, beat data, and analytic time functions are not animation keyframes. Do not use random numbers or a seeded PRNG, and do not drive drawing through a wall clock or timer.\par
Deliver a 1920$\times$1080, 30fps, exactly 450-frame, 15-second complete MP4. Frame n has time n/30, n=0\ldots{}449; do not output a terminal frame at t=15. 120 BPM equals 0.5 seconds per beat, or 30 beats; 4/4 time is 7 complete bars plus 2 beats, not 8 complete bars. {[}\ldots{]}
\end{specbox}
\runin{package}{Supplied assets} (18 files, 0.1\,MB), read-only in the sandbox:
\par\vspace{2pt}\noindent\begin{minipage}{0.86\textwidth}
{\footnotesize\setlength{\tabcolsep}{4pt}\renewcommand{\arraystretch}{1.12}
\begin{tabularx}{\linewidth}{@{}c l r >{\raggedright\arraybackslash}X@{}}
\ico{input-numbers} & Data & 10 & the film (stages, timing, palette), the original score (events, oscillators, envelope, mix), production inputs, and rules and templates for the FFT and the self-audit \\
\ico{hammer-and-wrench} & Scripts & 6 & optional checkers and measuring tools for audio, FFT, audit and media \\
\ico{page-facing-up} & Notes & 2 & a README, licence and credits \\
\end{tabularx}}
\end{minipage}
\par\vspace{4pt}
\begin{l2box}{C10}{Example L2 check}{G1 $\cdot$ core $\cdot$ one of 79 checks}
\footnotesize\renewcommand{\arraystretch}{1.08}\setlength{\tabcolsep}{0pt}
\begin{tabular}{@{}>{\bfseries}p{5.3em}>{\raggedright\arraybackslash}p{\dimexpr\linewidth-5.3em\relax}@{}}
Pass & Even-beat low kick and bass, odd-beat composite hits, and the melody/tick relationship on every beat are audible. \\
Prompt & \textit{``Every beat must have an audible rhythmic anchor.''} \\
\end{tabular}
\end{l2box}
\par\vspace{5pt}\noindent\ico{trophy}\hspace{0.3em}\textbf{L3 quality rubric:} task-specific anchors for a score of 5 in all five dimensions:
\par\vspace{2pt}{\footnotesize\setlength{\tabcolsep}{4pt}\renewcommand{\arraystretch}{1.08}
\noindent\begin{tabular}{@{}>{\raggedright\arraybackslash}p{0.17\textwidth}>{\raggedright\arraybackslash}p{\dimexpr0.83\textwidth-8pt\relax}@{}}
\toprule
\textbf{VQ} Visual quality & The same-track spectrum, version labels, and actual review images are clear and stable. \\
\textbf{VA} Visual aesthetics & The abstract geometry is unified, with precise hierarchy for the overview and enlarged minimum-issue view. \\
\textbf{CL} Content logic & The score-first approach, true spectrum, frozen scoring, and organization of same-t actual attempts/results are clear. \\
\textbf{TM} Timing \& motion & The current responses of the 12 specified frequency bands are discernible and present corresponding changes with the actual mix (not every band is required to change, and full-spectrum coverage is not claimed). \\
\textbf{AQ} Audio \& A/V sync & The coded melody, bass, and beat are distinct, and the mix itself is coordinated with the spectral changes. \\
\bottomrule
\end{tabular}}
\clearpage
\subsection{Classroom lessons (C11)}
\label{app:case:ZH26}
\noindent\caticon{C11}\hspace{0.4em}{\bfseries Digitizing images}\par\vspace{4pt}
\noindent\begin{minipage}[t]{0.42\textwidth}\vspace{0pt}
{\color{catC11}\rule{\linewidth}{1.2pt}}\par\vspace{1.5pt}
\includegraphics[width=\linewidth]{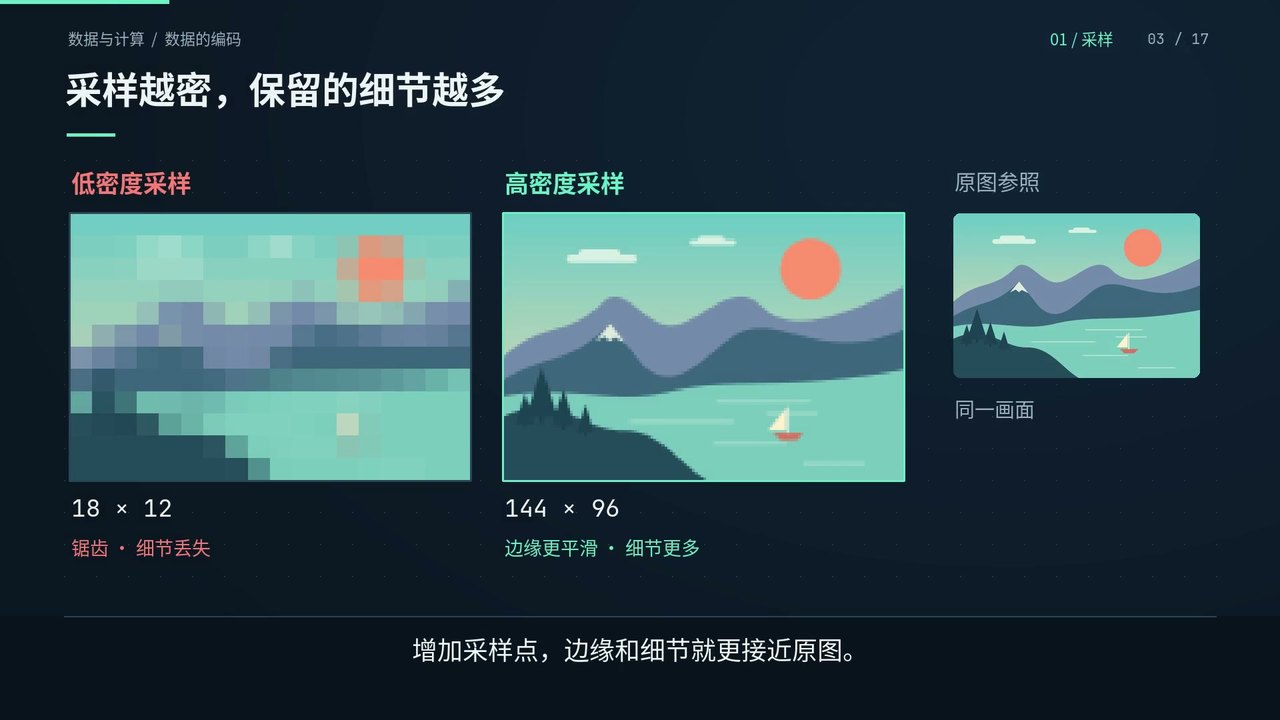}\par\vspace{2pt}
{\scriptsize\color{black!70}Frame at 37.9\,s of the video delivered by \logo{openai}\,GPT-6 Astra.}
\end{minipage}\hfill
\begin{minipage}[t]{0.555\textwidth}\vspace{0pt}
{\scriptsize\setlength{\tabcolsep}{3pt}\renewcommand{\arraystretch}{1.05}
\begin{tabular}{@{}l>{\raggedright\arraybackslash}p{0.68\linewidth}rr@{}}
\toprule Goal & Core goal & L2 & core \\ \midrule
G1 & The two key stages of digitization, sampling and quantization, and why and with what result a continuous image becomes a digital one. & 9 & 8 \\
G2 & Pixel, resolution, total pixel count, bit depth, number of colors, the typical meaning of 1/8/24 bits, and how the three RGB channels relate to true color. & 14 & 11 \\
G3 & Derive and apply the storage formula of an uncompressed bitmap, with B/KB/MB conversion and the 1024$\times$768 worked example. & 8 & 6 \\
G4 & How file size relates to resolution, bit depth and compression, correcting common misconceptions. & 9 & 7 \\
G5 & Diagrams or animation for the relations of Section 6: pixel grid, sampling density, quantization levels, binary mapping, RGB mixing and the storage calculation. & 6 & 6 \\
G6 & Chinese narration carries the explanation, synchronized with the picture. & 13 & 10 \\
\midrule & \textit{All L2 checks} & 59 & 48 \\ \bottomrule
\end{tabular}}
\end{minipage}\par\vspace{6pt}
\begin{specbox}{C11}{Task specification}{opening of 186 words $\cdot$ translated from Chinese}
\footnotesize\setlength{\parindent}{0pt}\setlength{\parskip}{2pt}
Digitizing images\par
Make an explainer video for 10th-grade students that makes "digitizing images" clear, based on the lecture note in the supplied file.\par
The note's core concepts, principles and derivation steps, worked examples and common misconceptions must all be covered, and covered correctly; the relationships listed in Section 6 of the note should be made clear with diagrams or animation, so that students can see the spatial, process or causal relations in them.\par
The whole video must have Chinese narration that carries the main explanation and is synchronized with the picture: whatever step the narration is on, the picture shows that step.\par
The order of explanation, visual style, camera, layout and color scheme are up to you. You need not read the note word for word and may add intuitive examples or analogies beyond it, but you must not contradict the note or make scientific errors. {[}\ldots{]}
\end{specbox}
\runin{package}{Supplied assets} (1 file, 8\,KB), read-only in the sandbox:
\par\vspace{2pt}\noindent\begin{minipage}{0.86\textwidth}
{\footnotesize\setlength{\tabcolsep}{4pt}\renewcommand{\arraystretch}{1.12}
\begin{tabularx}{\linewidth}{@{}c l r >{\raggedright\arraybackslash}X@{}}
\ico{page-facing-up} & Lecture note & 1 & the lecture note the lesson is based on (Markdown, in Chinese) \\
\end{tabularx}}
\end{minipage}
\par\vspace{4pt}
\begin{l2box}{C11}{Example L2 check}{G3 $\cdot$ core $\cdot$ one of 59 checks $\cdot$ translated from Chinese}
\footnotesize\renewcommand{\arraystretch}{1.08}\setlength{\tabcolsep}{0pt}
\begin{tabular}{@{}>{\bfseries}p{5.3em}>{\raggedright\arraybackslash}p{\dimexpr\linewidth-5.3em\relax}@{}}
Pass & Narration or picture gives the 24-bit data size as 2,359,296 bytes (from 786,432$\times$24$\div$8 or 786,432$\times$3), converted to 2,304 KB, about 2.25 MB; if the total of 18,874,368 bits is shown, its value must be correct. \\
Fail & The data size is miscalculated; or the division by 8 is forgotten and 18,874,368 is taken as the byte count; or the MB result is not about 2.25; or the 24-bit size and the unit conversion are not given. \\
Look at & The worked example's 24-bit size and unit conversion. \\
Lecture note & \textit{``Total bits = 786432$\times$24 = 18874368 bits; total bytes = 18874368$\div$8 = 2359296 bytes (or 786432$\times$3 = 2359296 bytes); 2359296$\div$1024 = 2304 KB; 2304$\div$1024 = 2.25 MB.''} \\
\end{tabular}
\end{l2box}
\par\vspace{5pt}\noindent\ico{trophy}\hspace{0.3em}\textbf{L3 quality rubric:} task-specific anchors for a score of 5 in all five dimensions (translated from Chinese):
\par\vspace{2pt}{\scriptsize\setlength{\tabcolsep}{4pt}\renewcommand{\arraystretch}{1.08}
\noindent\begin{tabular}{@{}>{\raggedright\arraybackslash}p{0.17\textwidth}>{\raggedright\arraybackslash}p{\dimexpr0.83\textwidth-8pt\relax}@{}}
\toprule
\textbf{VQ} Visual quality & At 1920$\times$1080 the picture is sharp and stable; text, formulas, arrows, grids, RGB mixing and the storage flow chart are clearly legible and still identifiable in sampled frames; no occlusion or unintended flicker. \\
\textbf{VA} Visual aesthetics & A unified, simple teaching design whose visual metaphors (grids, gray-level strips, binary mapping, RGB sliders, the storage flow) explain image digitization and engage 10th-graders without distracting. \\
\textbf{CL} Content logic & A clear path from analog versus digital through sampling, quantization, pixels, bit depth, RGB, the storage formula, the worked example and the misconceptions; narration and picture advance together, with a summary. \\
\textbf{TM} Timing \& motion & Motion and pacing match the Chinese narration; the sampling grid, gray-level steps, RGB mixing and unit conversion unfold step by step with the explanation; key points get pauses, and the whole is compact but not rushed within 120--300 s. \\
\textbf{AQ} Audio \& A/V sync & Clear, natural Chinese narration throughout reads terms, numbers and formulas correctly at a pace that allows following the derivation, keeps step with the picture, and is never covered by music or effects. \\
\bottomrule
\end{tabular}}
\clearpage
\subsection{Paper explainers (C12)}
\label{app:case:PA06}
\noindent\caticon{C12}\hspace{0.4em}{\bfseries Explaining "Auto-Encoding Variational Bayes"}\par\vspace{4pt}
\noindent\begin{minipage}[t]{0.42\textwidth}\vspace{0pt}
{\color{catC12}\rule{\linewidth}{1.2pt}}\par\vspace{1.5pt}
\includegraphics[width=\linewidth]{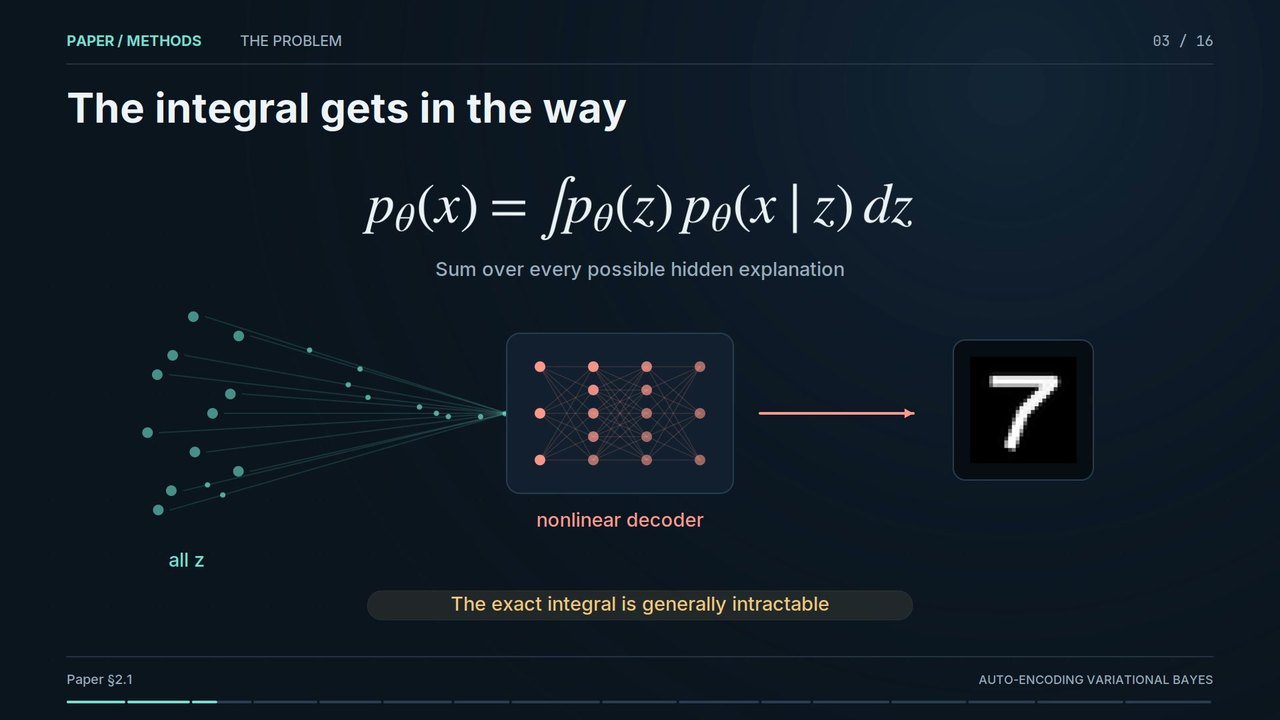}\par\vspace{2pt}
{\scriptsize\color{black!70}Frame at 59.8\,s of the video delivered by \logo{openai}\,GPT-6 Astra.}
\par\vspace{6pt}\centering
{\setlength{\fboxsep}{0pt}\setlength{\fboxrule}{0.3pt}\fbox{\includegraphics[width=0.3\linewidth]{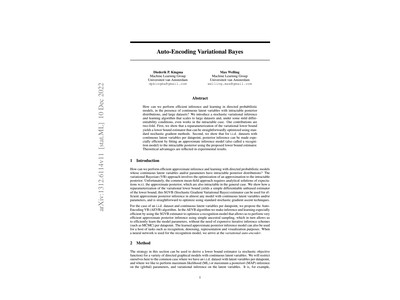}}}
\captionsetup{font=footnotesize,skip=3pt}\captionof{figure}{\textbf{Supplied paper.} First page of the PDF the explainer is based on.}
\end{minipage}\hfill
\begin{minipage}[t]{0.555\textwidth}\vspace{0pt}
{\footnotesize\setlength{\tabcolsep}{3pt}\renewcommand{\arraystretch}{1.05}
\begin{tabular}{@{}l>{\raggedright\arraybackslash}p{0.68\linewidth}rr@{}}
\toprule Goal & Core goal & L2 & core \\ \midrule
G1 & Explain the problem and motivation: efficient inference and learning in directed probabilistic models with continuous latent {[}\ldots{]} & 6 & 4 \\
G2 & Explain the key idea: reparameterizing the variational lower bound yields the differentiable SGVB estimator, enabling AEVB and the variational auto-encoder. & 4 & 3 \\
G3 & Explain how the method works: generative model, recognition model, variational bound, reparameterization trick, SGVB estimators {[}\ldots{]} & 24 & 11 \\
G4 & Report the paper's main results, supporting evidence, and stated limitations or open questions accurately. & 12 & 4 \\
G5 & Use diagrams and animations to make mechanisms, data flow, equations, and result comparisons clear and consistent with the paper. & 6 & 3 \\
G6 & English narration carries the explanation throughout and stays in sync with the visuals. & 12 & 9 \\
\midrule & \textit{All L2 checks} & 64 & 34 \\ \bottomrule
\end{tabular}}
\end{minipage}\par\vspace{6pt}
\begin{specbox}{C12}{Task specification}{opening of 205 words}
\footnotesize\setlength{\parindent}{0pt}\setlength{\parskip}{2pt}
Explaining "Auto-Encoding Variational Bayes"\par
Make an explainer video about the paper in the supplied file (arXiv:1312.6114) for first-year graduate students who know basic machine learning (neural networks, gradient descent) but have not read this paper.\par
After watching, a viewer should understand the problem the paper addresses and why it matters, the key idea and how the method actually works, the main results and what evidence supports them, and the paper's limitations or open questions. Explain the key mechanisms with diagrams or animation rather than only reading text aloud; equations, figures, and numbers you present must match the paper. {[}\ldots{]}
\end{specbox}
\runin{package}{Supplied assets} (1 file, 3.7\,MB), read-only in the sandbox:
\par\vspace{2pt}\noindent\begin{minipage}{0.86\textwidth}
{\footnotesize\setlength{\tabcolsep}{4pt}\renewcommand{\arraystretch}{1.12}
\begin{tabularx}{\linewidth}{@{}c l r >{\raggedright\arraybackslash}X@{}}
\ico{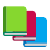} & Paper & 1 & the paper itself (PDF) \\
\end{tabularx}}
\end{minipage}
\par\vspace{4pt}
\begin{l2box}{C12}{Example L2 check}{G3 $\cdot$ core $\cdot$ one of 64 checks}
\footnotesize\renewcommand{\arraystretch}{1.08}\setlength{\tabcolsep}{0pt}
\begin{tabular}{@{}>{\bfseries}p{5.3em}>{\raggedright\arraybackslash}p{\dimexpr\linewidth-5.3em\relax}@{}}
Pass & Narration states the two-step generative process from prior latent z to observed x. \\
Fail & The video reverses the generative order, omits the prior, or says x generates z in the generative model. \\
Look at & Method section where the generative model is introduced. \\
Paper & \textit{``The process consists of two steps: (1) a value $\mathbf{z}^{(i)}$ is generated from some prior distribution $p_{\boldsymbol{\theta}^*}(\mathbf{z})$; (2) a value $\mathbf{x}^{(i)}$ is generated from some conditional distribution {[}\ldots{]}''} \\
\end{tabular}
\end{l2box}
\par\vspace{5pt}\noindent\ico{trophy}\hspace{0.3em}\textbf{L3 quality rubric:} task-specific anchors for a score of 5 in all five dimensions:
\par\vspace{2pt}{\footnotesize\setlength{\tabcolsep}{4pt}\renewcommand{\arraystretch}{1.08}
\noindent\begin{tabular}{@{}>{\raggedright\arraybackslash}p{0.17\textwidth}>{\raggedright\arraybackslash}p{\dimexpr0.83\textwidth-8pt\relax}@{}}
\toprule
\textbf{VQ} Visual quality & All equations, labels, axes, and diagram text are clearly legible at 1920$\times$1080, with sufficient size, contrast, and on-screen duration for first-year graduate viewers to read key symbols and numbers. \\
\textbf{VA} Visual aesthetics & Visuals are purpose-built for this paper: the graphical model, encoder-decoder flow, reparameterization, lower-bound decomposition, training loop, and results charts are represented with clear, task-specific diagrams or animations that directly support understanding. {[}\ldots{]} \\
\textbf{CL} Content logic & Content is organized as a clear explanatory arc: problem and why it matters, key idea, step-by-step method, evidence-based results, then limitations or open questions, with smooth transitions that make the logical dependencies easy to follow. \\
\textbf{TM} Timing \& motion & Motion and pacing are carefully synced with narration: each mechanism is revealed as it is described, dense equations get enough dwell time, and transitions are smooth without distracting motion. \\
\textbf{AQ} Audio \& A/V sync & English narration is clear and intelligible throughout, technical terms, equations, and numbers are read correctly, music or sound effects never mask the voice, and narration and visuals are tightly synchronized so every described step appears when spoken. \\
\bottomrule
\end{tabular}}


{\small\begin{thebibliography}{64}
\providecommand{\natexlab}[1]{#1}
\providecommand{\url}[1]{\texttt{#1}}
\expandafter\ifx\csname urlstyle\endcsname\relax
  \providecommand{\doi}[1]{doi: #1}\else
  \providecommand{\doi}{doi: \begingroup \urlstyle{rm}\Url}\fi

\bibitem[{Anthropic}(2025)]{anthropic2025claudecode}
{Anthropic}.
\newblock Claude code.
\newblock \url{https://docs.anthropic.com/en/docs/claude-code}, 2025.

\bibitem[{Anthropic}(2026{\natexlab{a}})]{anthropic2026opus55}
{Anthropic}.
\newblock Introducing {Claude Opus 5.5}.
\newblock \url{https://www.anthropic.com/claude-opus-5-5}, 2026{\natexlab{a}}.

\bibitem[{Anthropic}(2026{\natexlab{b}})]{anthropic2026sonnet55}
{Anthropic}.
\newblock Introducing {Claude Sonnet 5.5}.
\newblock \url{https://www.anthropic.com/claude-sonnet-5-5},
  2026{\natexlab{b}}.

\bibitem[Arora et~al.(2025)Arora, Wei, Hicks, Bowman, Qui{\~n}onero-Candela,
  Tsimpourlas, et~al.]{arora2025healthbench}
Rahul~K. Arora, Jason Wei, Rebecca~Soskin Hicks, Preston Bowman, Joaquin
  Qui{\~n}onero-Candela, Foivos Tsimpourlas, et~al.
\newblock {HealthBench}: Evaluating large language models towards improved
  human health.
\newblock \emph{arXiv preprint arXiv:2505.08775}, 2025.

\bibitem[Bansal et~al.(2025)Bansal, Lin, Xie, Zong, Yarom, Bitton, Jiang, Sun,
  Chang, and Grover]{bansal2025videophy}
Hritik Bansal, Zongyu Lin, Tianyi Xie, Zeshun Zong, Michal Yarom, Yonatan
  Bitton, Chenfanfu Jiang, Yizhou Sun, Kai-Wei Chang, and Aditya Grover.
\newblock {VideoPhy}: Evaluating physical commonsense for video generation.
\newblock In \emph{International Conference on Learning Representations
  (ICLR)}, 2025.

\bibitem[Bradley and Terry(1952)]{bradley1952rank}
Ralph~Allan Bradley and Milton~E. Terry.
\newblock Rank analysis of incomplete block designs: {I}. the method of paired
  comparisons.
\newblock \emph{Biometrika}, 39\penalty0 (3/4):\penalty0 324--345, 1952.

\bibitem[Brooks et~al.(2024)Brooks, Peebles, Holmes, DePue, Guo, Jing, Schnurr,
  Taylor, Luhman, Luhman, Ng, Wang, and Ramesh]{brooks2024sora}
Tim Brooks, Bill Peebles, Connor Holmes, Will DePue, Yufei Guo, Li Jing, David
  Schnurr, Joe Taylor, Troy Luhman, Eric Luhman, Clarence Ng, Ricky Wang, and
  Aditya Ramesh.
\newblock Video generation models as world simulators.
\newblock OpenAI technical report, 2024.

\bibitem[Cao et~al.(2026)Cao, Zheng, Song, and Hu]{cao2026agenticvbench}
Zongheng Cao, Yi Zheng, Rui Song, and Xinyu Hu.
\newblock {AgenticVBench}: Can {AI} agents complete real-world post-production
  tasks?
\newblock \emph{arXiv preprint arXiv:2605.27705}, 2026.

\bibitem[Chen et~al.(2024)Chen, Chen, Zhang, Wang, Liu, Zhou,
  et~al.]{chen2024mllmjudge}
Dongping Chen, Ruoxi Chen, Shilin Zhang, Yaochen Wang, Yinuo Liu, Huichi Zhou,
  et~al.
\newblock {MLLM-as-a-Judge}: Assessing multimodal {LLM-as-a-Judge} with
  vision-language benchmark.
\newblock In \emph{International Conference on Machine Learning (ICML)}, 2024.

\bibitem[Chen et~al.(2025{\natexlab{a}})Chen, Shi, Huang, Li, Pan, Chen, and
  He]{chen2025visualedu}
Hao Chen, Tianyu Shi, Pengran Huang, Zeyuan Li, Jiahui Pan, Qianglong Chen, and
  Lewei He.
\newblock {VisualEDU}: A benchmark for assessing coding and visual
  comprehension through educational problem-solving video generation.
\newblock In \emph{Findings of the Association for Computational Linguistics:
  EMNLP 2025}, pages 16363--16394, 2025{\natexlab{a}}.

\bibitem[Chen et~al.(2025{\natexlab{b}})]{chen2025svgenius}
Siqi Chen et~al.
\newblock {SVGenius}: Benchmarking {LLMs} in {SVG} understanding, editing and
  generation.
\newblock \emph{arXiv preprint arXiv:2506.03139}, 2025{\natexlab{b}}.

\bibitem[Chen et~al.(2026{\natexlab{a}})Chen, Lin, and
  Shou]{chen2026code2video}
Yanzhe Chen, Kevin~Qinghong Lin, and Mike~Zheng Shou.
\newblock {Code2Video}: A code-centric paradigm for educational video
  generation.
\newblock In \emph{International Conference on Machine Learning (ICML)},
  2026{\natexlab{a}}.
\newblock arXiv:2510.01174.

\bibitem[Chen et~al.(2026{\natexlab{b}})Chen, Shao, Lin, Cheng, and
  Chen]{chen2026code2videocompass}
Zhenghan Chen, Zekai Shao, Xin Lin, Leting Cheng, and Siming Chen.
\newblock {Code2Video-Compass}: A comprehensive benchmark for general-purpose
  code-based video generation across {T2V}, {I2V}, and {V2V}.
\newblock \emph{arXiv preprint arXiv:2610.28416}, 2026{\natexlab{b}}.

\bibitem[Chiang et~al.(2024)Chiang, Zheng, Sheng, Angelopoulos, Li, Li,
  et~al.]{chiang2024chatbotarena}
Wei-Lin Chiang, Lianmin Zheng, Ying Sheng, Anastasios~Nikolas Angelopoulos,
  Tianle Li, Dacheng Li, et~al.
\newblock Chatbot arena: An open platform for evaluating {LLMs} by human
  preference.
\newblock In \emph{International Conference on Machine Learning (ICML)}, 2024.

\bibitem[{DeepSeek-AI}(2026{\natexlab{a}})]{deepseek2026dsh}
{DeepSeek-AI}.
\newblock {DeepSeek Harness}.
\newblock \url{https://github.com/deepseek-ai/deepseek-harness},
  2026{\natexlab{a}}.

\bibitem[{DeepSeek-AI}(2026{\natexlab{b}})]{deepseek2026v41flash}
{DeepSeek-AI}.
\newblock {DeepSeek-V4.1-Flash-1017}: Wish-driven programming made real.
\newblock \url{https://api-docs.deepseek.com/news/news261017/},
  2026{\natexlab{b}}.

\bibitem[Gong et~al.(2024)Gong, Feng, Li, Wang, Cheng, Yang,
  et~al.]{gong2024avodyssey}
Kaixiong Gong, Kaituo Feng, Bohao Li, Yibing Wang, Mofang Cheng, Shijia Yang,
  et~al.
\newblock {AV-Odyssey} bench: Can your multimodal {LLMs} really understand
  audio-visual information?
\newblock \emph{arXiv preprint arXiv:2412.02611}, 2024.

\bibitem[{Google}(2026{\natexlab{a}})]{google2026antigravity}
{Google}.
\newblock {Google Antigravity}.
\newblock \url{https://antigravity.google/}, 2026{\natexlab{a}}.

\bibitem[{Google}(2026{\natexlab{b}})]{google2026gemini38flash}
{Google}.
\newblock What's new in {Gemini 3.8 Flash}.
\newblock \url{https://ai.google.dev/gemini-api/docs/whats-new-gemini-3.8},
  2026{\natexlab{b}}.

\bibitem[{Google}(2026{\natexlab{c}})]{google2026gemini4}
{Google}.
\newblock {Gemini 4 Argon}: Built to solve your hardest problems.
\newblock
  \url{https://blog.google/innovation-and-ai/models-and-research/gemini-models/introducing-gemini-4-argon/},
  2026{\natexlab{c}}.

\bibitem[{Google DeepMind}(2025)]{veo3}
{Google DeepMind}.
\newblock Veo 3.
\newblock Model card and technical overview, 2025.

\bibitem[Gu et~al.(2025)Gu, Huang, Je, Yang, and Guibas]{gu2025blendergym}
Yunqi Gu, Ian Huang, Jihyeon Je, Guandao Yang, and Leonidas Guibas.
\newblock {BlenderGym}: Benchmarking foundational model systems for graphics
  editing.
\newblock In \emph{IEEE/CVF Conference on Computer Vision and Pattern
  Recognition (CVPR)}, 2025.

\bibitem[Gunjal et~al.(2025)Gunjal, Wang, Lau, Nath, He, Liu, and
  Hendryx]{gunjal2025rubrics}
Anisha Gunjal, Anthony Wang, Elaine Lau, Vaskar Nath, Yunzhong He, Bing Liu,
  and Sean Hendryx.
\newblock Rubrics as rewards: Reinforcement learning beyond verifiable domains.
\newblock \emph{arXiv preprint arXiv:2507.17746}, 2025.

\bibitem[He et~al.(2024)He, Jiang, Zhang, Ku, Soni, Siu,
  et~al.]{he2024videoscore}
Xuan He, Dongfu Jiang, Ge Zhang, Max Ku, Achint Soni, Sherman Siu, et~al.
\newblock {VideoScore}: Building automatic metrics to simulate fine-grained
  human feedback for video generation.
\newblock In \emph{Conference on Empirical Methods in Natural Language
  Processing (EMNLP)}, 2024.

\bibitem[{HeyGen Research}(2026)]{heygen2026code2videobench}
{HeyGen Research}.
\newblock Introducing the {Code2Video} benchmark.
\newblock Technical report and Kaggle leaderboard, 2026.

\bibitem[Hong et~al.(2025)Hong, Yan, Cai, Jiang, Hu, and
  Xie]{hong2025worldsense}
Jack Hong, Shilin Yan, Jiayin Cai, Xiaolong Jiang, Yao Hu, and Weidi Xie.
\newblock {WorldSense}: Evaluating real-world omnimodal understanding for
  multimodal {LLMs}.
\newblock \emph{arXiv preprint arXiv:2502.04326}, 2025.

\bibitem[Huang et~al.(2024)Huang, He, Yu, Zhang, Si, Jiang, Zhang, Wu, Jin,
  Chanpaisit, Wang, Chen, Wang, Lin, Qiao, and Liu]{huang2024vbench}
Ziqi Huang, Yinan He, Jiashuo Yu, Fan Zhang, Chenyang Si, Yuming Jiang, Yuanhan
  Zhang, Tianxing Wu, Qingyang Jin, Nattapol Chanpaisit, Yaohui Wang, Xinyuan
  Chen, Limin Wang, Dahua Lin, Yu Qiao, and Ziwei Liu.
\newblock {VBench}: Comprehensive benchmark suite for video generative models.
\newblock In \emph{IEEE/CVF Conference on Computer Vision and Pattern
  Recognition (CVPR)}, 2024.

\bibitem[Jiang et~al.(2024)Jiang, Ku, Li, Ni, Sun, Fan, and
  Chen]{jiang2024genaiarena}
Dongfu Jiang, Max Ku, Tianle Li, Yuansheng Ni, Shizhuo Sun, Rongqi Fan, and
  Wenhu Chen.
\newblock {GenAI Arena}: An open evaluation platform for generative models.
\newblock In \emph{Advances in Neural Information Processing Systems (NeurIPS),
  Datasets and Benchmarks Track}, 2024.

\bibitem[Jiang et~al.(2026)Jiang, Cai, Shao, Wang, Han, Song, Chen, An, Yang,
  and Yang]{jiang2026manimagent}
Wenjia Jiang, Zongyuan Cai, Yuanhang Shao, Chenru Wang, Boyan Han, Zhixue Song,
  Keyu Chen, Shengwei An, Xu Yang, and Zhou Yang.
\newblock {ManimAgent}: Self-evolving multimodal agents for visual education.
\newblock \emph{arXiv preprint arXiv:2606.30296}, 2026.
\newblock EMNLP 2026.

\bibitem[Jimenez et~al.(2024)Jimenez, Yang, Wettig, Yao, Pei, Press, and
  Narasimhan]{jimenez2024swebench}
Carlos~E. Jimenez, John Yang, Alexander Wettig, Shunyu Yao, Kexin Pei, Ofir
  Press, and Karthik Narasimhan.
\newblock {SWE-bench}: Can language models resolve real-world {GitHub} issues?
\newblock In \emph{International Conference on Learning Representations
  (ICLR)}, 2024.

\bibitem[Kim et~al.(2024)Kim, Shin, Cho, Jang, Longpre, Lee, Yun, Shin, Kim,
  Thorne, and Seo]{kim2024prometheus}
Seungone Kim, Jamin Shin, Yejin Cho, Joel Jang, Shayne Longpre, Hwaran Lee,
  Sangdoo Yun, Seongjin Shin, Sungdong Kim, James Thorne, and Minjoon Seo.
\newblock Prometheus: Inducing fine-grained evaluation capability in language
  models.
\newblock In \emph{International Conference on Learning Representations
  (ICLR)}, 2024.

\bibitem[Kong et~al.(2024)Kong, Tian, Zhang, Min, Dai, Zhou,
  et~al.]{kong2024hunyuanvideo}
Weijie Kong, Qi Tian, Zijian Zhang, Rox Min, Zuozhuo Dai, Jin Zhou, et~al.
\newblock {HunyuanVideo}: A systematic framework for large video generative
  models.
\newblock \emph{arXiv preprint arXiv:2412.03603}, 2024.

\bibitem[Ku et~al.(2025)Ku, Chong, Leung, Shah, Yu, and
  Chen]{ku2025theoremexplainagent}
Max Ku, Cheuk~Hei Chong, Jonathan Leung, Krish Shah, Alvin Yu, and Wenhu Chen.
\newblock {TheoremExplainAgent}: Towards video-based multimodal explanations
  for {LLM} theorem understanding.
\newblock In \emph{Annual Meeting of the Association for Computational
  Linguistics (ACL)}, 2025.

\bibitem[{Kuaishou Technology}(2024)]{kling}
{Kuaishou Technology}.
\newblock Kling {AI}: Next-generation {AI} creative studio, 2024.

\bibitem[Li et~al.(2025)Li, Yang, Chen, Cao, Liu, Lu, Wu, Zhang, Ji, Tang, Wu, and Zhu]{li2025opusanimation}
Bozheng Li, Miao Yang, Zhenhan Chen, Jiawang Cao, Mushui Liu, Yi Lu,
  Yongliang Wu, Bin Zhang, Yangguang Ji, Licheng Tang, Jay Wu, and Wenbo Zhu.
\newblock {OpusAnimation}: Code-based dynamic chart generation.
\newblock \emph{arXiv preprint arXiv:2510.03341}, 2025.

\bibitem[Lin et~al.(2023)Lin, Zala, Cho, and Bansal]{lin2023videodirectorgpt}
Han Lin, Abhay Zala, Jaemin Cho, and Mohit Bansal.
\newblock {VideoDirectorGPT}: Consistent multi-scene video generation via
  {LLM}-guided planning.
\newblock \emph{arXiv preprint arXiv:2309.15091}, 2023.
\newblock COLM 2024.

\bibitem[Liu et~al.(2024)Liu, Cun, Liu, Wang, Zhang, Chen, Liu, Zeng, Chan, and
  Shan]{liu2024evalcrafter}
Yaofang Liu, Xiaodong Cun, Xuebo Liu, Xintao Wang, Yong Zhang, Haoxin Chen,
  Yang Liu, Tieyong Zeng, Raymond Chan, and Ying Shan.
\newblock {EvalCrafter}: Benchmarking and evaluating large video generation
  models.
\newblock In \emph{IEEE/CVF Conference on Computer Vision and Pattern
  Recognition (CVPR)}, 2024.

\bibitem[Meng et~al.(2024)Meng, Liao, Tan, Shao, Lu, Zhang,
  et~al.]{meng2024phygenbench}
Fanqing Meng, Jiaqi Liao, Xinyu Tan, Wenqi Shao, Quanfeng Lu, Kaipeng Zhang,
  et~al.
\newblock Towards world simulator: Crafting physical commonsense-based
  benchmark for video generation.
\newblock \emph{arXiv preprint arXiv:2410.05363}, 2024.

\bibitem[Merrill et~al.(2026)Merrill, Shaw, Carlini, Li, Raj, Bercovich,
  et~al.]{merrill2026terminalbench}
Mike~A. Merrill, Alexander~G. Shaw, Nicholas Carlini, Boxuan Li, Harsh Raj,
  Ivan Bercovich, et~al.
\newblock {Terminal-Bench}: Benchmarking agents on hard, realistic tasks in
  command line interfaces.
\newblock In \emph{International Conference on Learning Representations
  (ICLR)}, 2026.
\newblock arXiv:2601.11868.

\bibitem[{Moonshot AI}(2026{\natexlab{a}})]{moonshot2026kimicode}
{Moonshot AI}.
\newblock {Kimi Code}.
\newblock \url{https://www.kimi.com/code/docs/en/}, 2026{\natexlab{a}}.

\bibitem[{Moonshot AI}(2026{\natexlab{b}})]{moonshot2026kimik3}
{Moonshot AI}.
\newblock {Kimi K3} tech blog: Open frontier intelligence.
\newblock \url{https://www.kimi.com/en/blog/kimi-k3}, 2026{\natexlab{b}}.

\bibitem[{OpenAI}(2025)]{openai2025codexcli}
{OpenAI}.
\newblock Codex {CLI}.
\newblock \url{https://github.com/openai/codex}, 2025.

\bibitem[{OpenAI}(2026{\natexlab{a}})]{openai2026gpt56}
{OpenAI}.
\newblock {GPT-5.6}: Frontier intelligence that scales with your ambition.
\newblock \url{https://openai.com/index/gpt-5-6/}, 2026{\natexlab{a}}.

\bibitem[{OpenAI}(2026{\natexlab{b}})]{openai2026gpt6astra}
{OpenAI}.
\newblock {GPT-6 Astra}: A new generation of intelligence.
\newblock \url{https://openai.com/index/gpt-6-astra/}, 2026{\natexlab{b}}.

\bibitem[Patwardhan et~al.(2026)]{patwardhan2026gdpval}
Tejal Patwardhan et~al.
\newblock {GDPval}: Evaluating {AI} model performance on real-world
  economically valuable tasks.
\newblock In \emph{International Conference on Learning Representations
  (ICLR)}, 2026.
\newblock arXiv:2510.04374.

\bibitem[Polyak et~al.(2024)Polyak, Zohar, Brown, Tjandra, Sinha, Lee,
  et~al.]{polyak2024moviegen}
Adam Polyak, Amit Zohar, Andrew Brown, Andros Tjandra, Animesh Sinha, Ann Lee,
  et~al.
\newblock Movie gen: A cast of media foundation models.
\newblock \emph{arXiv preprint arXiv:2410.13720}, 2024.

\bibitem[{Qwen Team}(2025)]{qwen3tts}
{Qwen Team}.
\newblock {Qwen3-TTS-Flash}.
\newblock
  \url{https://qwen.ai/blog?id=b4264e11fb80b5e37350790121baf0a0f10daf82}, 2025.

\bibitem[{Qwen Team}(2026)]{qwen2026omni38flash}
{Qwen Team}.
\newblock {Qwen3.8-Omni-Flash: Omni Senses. Agentic Delivery}.
\newblock \url{https://qwen.ai/blog?id=qwen3.8-omni-flash}, 2026.

\bibitem[Shi et~al.(2025)Shi, Yang, Liu, Shui, Wang, Jing,
  et~al.]{shi2025chartmimic}
Chufan Shi, Cheng Yang, Yaxin Liu, Bo Shui, Junjie Wang, Mohan Jing, et~al.
\newblock {ChartMimic}: Evaluating {LMM}'s cross-modal reasoning capability via
  chart-to-code generation.
\newblock In \emph{International Conference on Learning Representations
  (ICLR)}, 2025.

\bibitem[Si et~al.(2025)Si, Zhang, Li, Yang, Liu, and Yang]{si2025design2code}
Chenglei Si, Yanzhe Zhang, Ryan Li, Zhengyuan Yang, Ruibo Liu, and Diyi Yang.
\newblock {Design2Code}: Benchmarking multimodal code generation for automated
  front-end engineering.
\newblock In \emph{Conference of the North American Chapter of the Association
  for Computational Linguistics (NAACL)}, 2025.

\bibitem[Starace et~al.(2025)Starace, Jaffe, Sherburn, Aung, Chan, Maksin,
  et~al.]{starace2025paperbench}
Giulio Starace, Oliver Jaffe, Dane Sherburn, James Aung, Jun~Shern Chan, Leon
  Maksin, et~al.
\newblock {PaperBench}: Evaluating {AI}'s ability to replicate {AI} research.
\newblock In \emph{International Conference on Machine Learning (ICML)}, 2025.

\bibitem[{Wan Team}(2025)]{wan2025wan}
{Wan Team}.
\newblock Wan: Open and advanced large-scale video generative models.
\newblock \emph{arXiv preprint arXiv:2503.20314}, 2025.

\bibitem[Wang et~al.(2024)Wang, Li, Chen, Cai, Zhu, Lin, Cao, Liu, Liu, and
  Sui]{wang2024notfair}
Peiyi Wang, Lei Li, Liang Chen, Zefan Cai, Dawei Zhu, Binghuai Lin, Yunbo Cao,
  Qi Liu, Tianyu Liu, and Zhifang Sui.
\newblock Large language models are not fair evaluators.
\newblock In \emph{Annual Meeting of the Association for Computational
  Linguistics (ACL)}, 2024.

\bibitem[Wu et~al.(2025)Wu, Zhu, and Shou]{wu2025movieagent}
Weijia Wu, Zeyu Zhu, and Mike~Zheng Shou.
\newblock Automated movie generation via multi-agent {CoT} planning.
\newblock \emph{arXiv preprint arXiv:2503.07314}, 2025.

\bibitem[Xie et~al.(2024)Xie, Zhang, Chen, Li, Zhao, Cao,
  et~al.]{xie2024osworld}
Tianbao Xie, Danyang Zhang, Jixuan Chen, Xiaochuan Li, Siheng Zhao, Ruisheng
  Cao, et~al.
\newblock {OSWorld}: Benchmarking multimodal agents for open-ended tasks in
  real computer environments.
\newblock In \emph{Advances in Neural Information Processing Systems (NeurIPS),
  Datasets and Benchmarks Track}, 2024.

\bibitem[Xu et~al.(2025)Xu, Guo, Hu, Chu, Wang, He, et~al.]{xu2025qwen3omni}
Jin Xu, Zhifang Guo, Hangrui Hu, Yunfei Chu, Xiong Wang, Jinzheng He, et~al.
\newblock {Qwen3-Omni} technical report.
\newblock \emph{arXiv preprint arXiv:2509.17765}, 2025.

\bibitem[Yang et~al.(2025)Yang, Teng, Zheng, Ding, Huang, Xu,
  et~al.]{yang2025cogvideox}
Zhuoyi Yang, Jiayan Teng, Wendi Zheng, Ming Ding, Shiyu Huang, Jiazheng Xu,
  et~al.
\newblock {CogVideoX}: Text-to-video diffusion models with an expert
  transformer.
\newblock In \emph{International Conference on Learning Representations
  (ICLR)}, 2025.

\bibitem[Zeng et~al.(2026)Zeng, Wang, Yu, Hua, and Luo]{zeng2026videoargus}
Ziyun Zeng, Zixuan Wang, Yongsheng Yu, Hang Hua, and Jiebo Luo.
\newblock {VideoArgus}: Agentic rubric-grounded unified evaluation for video
  generation and editing.
\newblock \emph{arXiv preprint arXiv:2608.05485}, 2026.

\bibitem[Zhang et~al.(2025)Zhang, Li, Xu, Liu, Liu, Zhou,
  et~al.]{zhang2025artifactsbench}
Chenchen Zhang, Yuhang Li, Can Xu, Jiaheng Liu, Ao Liu, Changzhi Zhou, et~al.
\newblock {ArtifactsBench}: Bridging the visual-interactive gap in {LLM} code
  generation evaluation.
\newblock \emph{arXiv preprint arXiv:2507.04952}, 2025.

\bibitem[Zheng et~al.(2025)Zheng, Huang, Liu, Zou, He, Zhang,
  et~al.]{zheng2025vbench2}
Dian Zheng, Ziqi Huang, Hongbo Liu, Kai Zou, Yinan He, Fan Zhang, et~al.
\newblock {VBench-2.0}: Advancing video generation benchmark suite for
  intrinsic faithfulness.
\newblock \emph{arXiv preprint arXiv:2503.21755}, 2025.

\bibitem[Zheng et~al.(2023)Zheng, Chiang, Sheng, Zhuang, Wu, Zhuang, Lin, Li,
  Li, Xing, Zhang, Gonzalez, and Stoica]{zheng2023judging}
Lianmin Zheng, Wei-Lin Chiang, Ying Sheng, Siyuan Zhuang, Zhanghao Wu, Yonghao
  Zhuang, Zi Lin, Zhuohan Li, Dacheng Li, Eric~P. Xing, Hao Zhang, Joseph~E.
  Gonzalez, and Ion Stoica.
\newblock Judging {LLM}-as-a-judge with {MT-Bench} and {Chatbot Arena}.
\newblock In \emph{Advances in Neural Information Processing Systems (NeurIPS),
  Datasets and Benchmarks Track}, 2023.

\bibitem[{Zhipu AI}(2026{\natexlab{a}})]{zhipu2026glm53flash}
{Zhipu AI}.
\newblock {GLM-5}: From vibe coding to agentic engineering.
\newblock \url{https://github.com/zai-org/GLM-5}, 2026{\natexlab{a}}.

\bibitem[{Zhipu AI}(2026{\natexlab{b}})]{zhipu2026zcode}
{Zhipu AI}.
\newblock {ZCode}.
\newblock \url{https://zcode.z.ai/en}, 2026{\natexlab{b}}.

\bibitem[Zhu et~al.(2025)Zhu, Lin, and Shou]{zhu2025paper2video}
Zeyu Zhu, Kevin~Qinghong Lin, and Mike~Zheng Shou.
\newblock {Paper2Video}: Automatic video generation from scientific papers.
\newblock \emph{arXiv preprint arXiv:2510.05096}, 2025.

\end{thebibliography}
}
\end{document}